\documentclass[11pt]{article}
\usepackage{booktabs}
\usepackage{caption}
\usepackage{adjustbox} 
\usepackage{booktabs}  
\usepackage{makecell}  
\usepackage{graphicx}  
\usepackage{tabularx}
\usepackage{booktabs}
\usepackage{array} 
\usepackage{makecell}
\usepackage{array}
\usepackage{authblk}
\usepackage[preprint]{acl}
\usepackage{amssymb}
\usepackage{times}
\usepackage{latexsym}
 \usepackage{amsmath}
\usepackage[T1]{fontenc}
\usepackage{adjustbox} 
\usepackage[normalem]{ulem} 

\usepackage[utf8]{inputenc}

\usepackage{microtype}

\usepackage{inconsolata}

\usepackage{graphicx}
\usepackage{enumitem}
\usepackage{booktabs}
\usepackage{caption}
\usepackage{adjustbox} 
\usepackage[normalem]
{ulem} 
\usepackage{graphicx}
\usepackage[table]{xcolor}
\usepackage{multirow}
\usepackage{arydshln}
\usepackage{array}
\usepackage{calc}
\usepackage{cancel}
\usepackage{tikz}
\usepackage{colortbl}
\usepackage{pifont}    

\usepackage[most]{tcolorbox}
\usepackage{longtable}  
\usepackage{graphicx}  
\usepackage{multicol}  

\definecolor{diagnosticpurple}{HTML}{544D61}
\definecolor{deepblue}{HTML}{1F5AA6}
\definecolor{alizarin}{HTML}{E74C3C}

\newcommand{\tcancel}[1]{%
  \tikz[baseline=(X.base)]{
    \node[inner sep=0pt, outer sep=0pt] (X) {#1};
    \draw[line width=0.4pt] (X.south west) -- (X.north east);
  }%
}
\title{MentalSeek-Dx: Towards Progressive Hypothetico-Deductive Reasoning for Real-world Psychiatric Diagnosis
}

\author{
Xiao Sun\textsuperscript{1†},
Yuming Yang\textsuperscript{1†},
Junnan Zhu\textsuperscript{2},
Jiang Zhong\textsuperscript{1*},
Xinyu Zhou\textsuperscript{3},
Kaiwen Wei\textsuperscript{1*}
\\
\textsuperscript{1} School of Computer Science, Chongqing University, Chongqing, China \\
\textsuperscript{2} MAIS, Institute of Automation, Chinese Academy of Sciences, China \\
\textsuperscript{3} The First Affiliated Hospital of Chongqing Medical University, Chongqing, China \\
\text{\href{mailto:sunx@stu.cqu.edu.cn}{sunx@stu.cqu.edu.cn},
\href{mailto:jiangzhong@cqu.edu.cn}{jiangzhong@cqu.edu.cn},
\href{mailto:weikaiwen@cqu.edu.cn}{weikaiwen@cqu.edu.cn}}
}

\begin{document}
\maketitle
\begin{abstract}
Mental health disorders represent a burgeoning global public health challenge. While Large Language Models (LLMs) have demonstrated potential in psychiatric assessment, their clinical utility is severely constrained by benchmarks that lack ecological validity and fine-grained diagnostic supervision. To bridge this gap, we introduce \textbf{MentalDx Bench}, the first benchmark dedicated to disorder-level psychiatric diagnosis within real-world clinical settings. Comprising 712 de-identified electronic health records annotated by board-certified psychiatrists under ICD-11 guidelines, the benchmark covers 76 disorders across 16 diagnostic categories. 
Evaluation of 18 LLMs reveals a critical \textit{paradigm misalignment}: strong performance at coarse diagnostic categorization contrasts with systematic failure at disorder-level diagnosis, underscoring a gap between pattern-based modeling and clinical hypothetico-deductive reasoning.
In response, we propose \textbf{MentalSeek-Dx}, a medical-specialized LLM trained to internalize this clinical reasoning process through supervised trajectory construction and curriculum-based reinforcement learning. Experiments on MentalDx Bench demonstrate that MentalSeek-Dx achieves state-of-the-art (SOTA) performance with only 14B parameters, establishing a clinically grounded framework for reliable psychiatric diagnosis. The dataset and code are available at \href{https://github.com/sunxiaofupo/MentalSeek-Dx}{MentalSeek-Dx}.

\end{abstract}

\section{Introduction}

\footnotetext[1]{† These authors contributed equally to this work.}
\footnotetext[2]{* Corresponding author.}

Mental health disorders affect over one billion individuals worldwide, constituting a major contributor to the global burden of disease\footnote{\href{https://www.who.int/news/item/02-09-2025-who-releases-new-reports-and-estimates-highlighting-urgent-gaps-in-mental-health}{WHO
: World Mental Health Today}}. However, the acute shortage of mental health professionals and clinical infrastructure often leads to delayed diagnoses and inadequate treatment, underscoring the urgent need for scalable, automated diagnostic alternatives. In this context, LLMs have emerged as promising tools for psychiatric assessment~\citep{ge2025survey}. While early applications primarily focused on population-level surveillance or detecting surface-level psychological distress~\citep{mangalik2024robust,zhuang2024postgraduate}, recent efforts have advanced toward identifying specific clinical risks, such as suicidal ideation and depression symptoms~\citep{levkovich2024large,kim2025baseline,lho2025large,wang2025large}. Despite this progress, a significant leap remains from symptom detection to the rigor of \textit{disorder-level diagnosis} required in clinical decision support.

\begin{figure}[t]
  \centering
  \includegraphics[width=\linewidth]{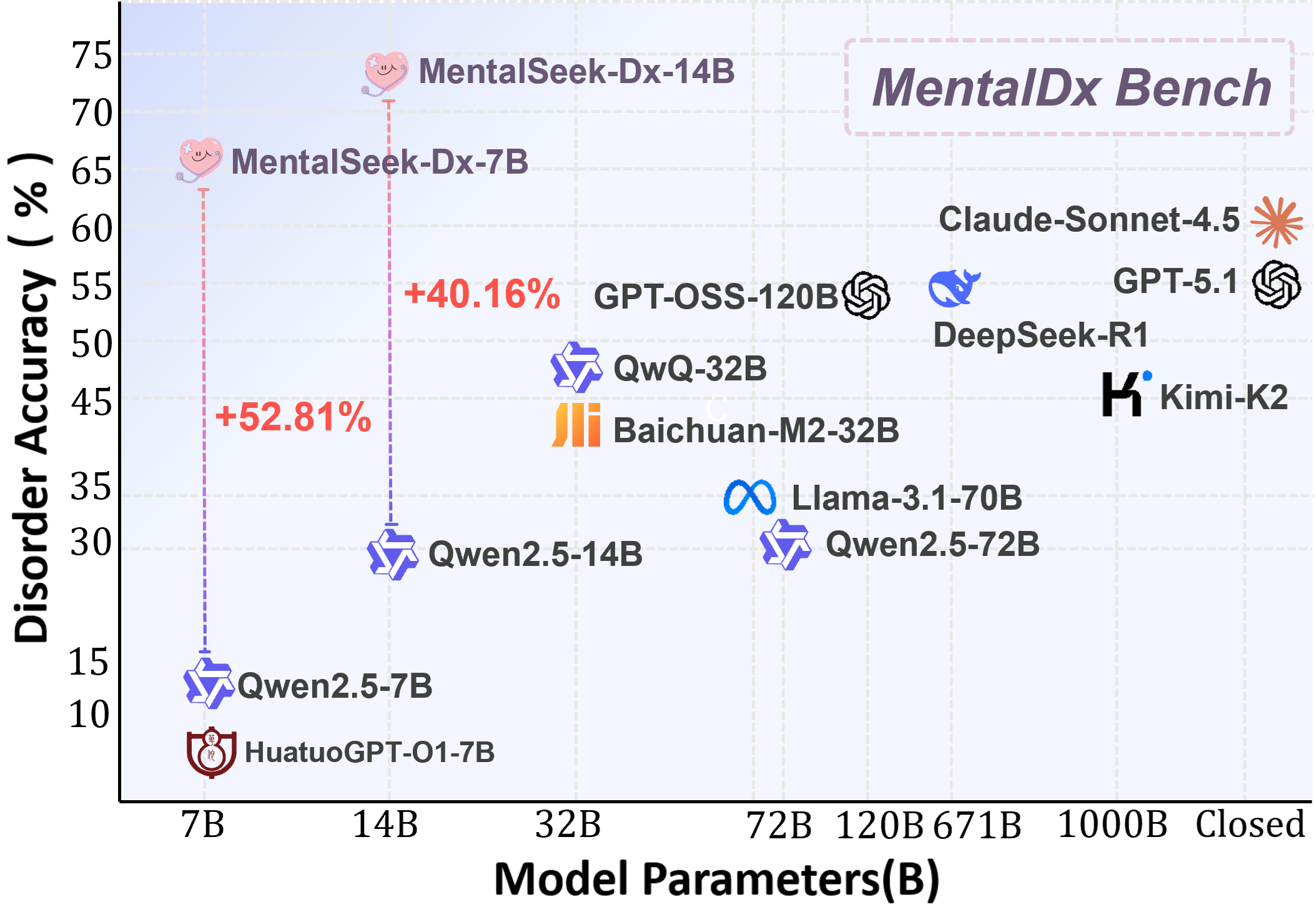}
\caption{Diagnostic accuracy results on the \textbf{MentalDx} benchmark, evaluated across mainstream LLMs. The proposed model \textsc{MentalSeek-Dx} achieves state-of-the-art performance while operating at smaller scales.}
  \label{fig_1_Introduction}
\end{figure}

A primary bottleneck impeding this advancement is the lack of clinically realistic benchmarks. Existing evaluation frameworks are limited in two fundamental respects. \textit{(1) Data Unrealism:} Most benchmarks rely on non-clinical proxies, such as social media posts~\citep{wu2025psychological} or synthetic narratives~\citep{hu2024psycollm}, which diverge substantially from the structured, jargon-rich, and information-dense documentation used in routine psychiatry. \textit{(2) Annotation Coarseness:} Prevailing datasets often capture only surface-level symptom mentions or assign broad disorder labels without grounding in formal diagnostic criteria (e.g., ICD-11)~\citep{hengle2024still}. Consequently, models optimized on these benchmarks may learn superficial pattern matching~\citep{liu2025psychbench, ignashina2025llm, lho2025large} rather than the nuanced differential diagnosis essential for real-world clinical application.

To address these limitations, we introduce \textbf{MentalDx Bench}, the first benchmark specifically designed to evaluate fine-grained, disorder-level psychiatric diagnosis in authentic clinical settings. Unlike prior datasets (see Table~\ref{tab:mh_benchmarks_compact}), MentalDx Bench comprises 712 de-identified medical records directly sourced from medical centers. Each case is rigorously annotated by board-certified psychiatrists following strict ICD-11 guidelines, providing a dual-layer label structure that spans 16 broad diagnostic categories and 76 specific disorders. This clinically grounded design allows for a high-fidelity assessment of an LLM's ability to navigate the complexity of real-world psychiatric cases.

We evaluate 18 mainstream and specialized LLMs on MentalDx Bench. The results reveal a striking phenomenon: while models achieve respectable accuracy in coarse diagnostic categorization (e.g., identifying a \textit{Mood Disorder}), their performance degrades sharply on fine-grained disorder-level diagnosis (e.g., distinguishing \textit{Bodily Distress Disorder}). Through psychiatrist-led diagnostic reviews, we identify the root cause as a fundamental \textbf{paradigm misalignment}. Mainstream LLMs predominantly rely on \textit{associative shortcuts}—linking symptom keywords directly to disease labels. In contrast, clinical psychiatry demands a structured, \textit{hypothetico-deductive reasoning} process: synthesizing symptoms, generating differential hypotheses, and rigorously excluding alternatives based on diagnostic criteria. This cognitive gap renders standard LLMs opaque and unreliable for precise clinical inference.


To resolve this misalignment, we propose \textbf{MentalSeek-Dx}, a domain-specialized LLM designed to align with the cognitive process of human psychiatrists. Inspired by clinical decision-making theories~\cite{collins2017clinical}, MentalSeek-Dx adopts a progressive reasoning framework trained in two stages. First, we employ \textit{Hypothetico-Deductive Trajectory Construction}, a supervised strategy decomposing diagnosis into explicit steps: symptom categorization, hypothesis generation, and differential diagnosis. Second, we introduce \textit{Reward-Based Curriculum Reinforcement Learning}, which progressively guides the model from mastering coarse categorization to refining fine-grained differential logic. As shown in Fig.~\ref{fig_1_Introduction}, experiments demonstrate this approach instills clinically coherent reasoning, enabling our 14B-parameter model to achieve SOTA diagnostic accuracy, surpassing significantly larger general-purpose LLMs. In summary, our contributions are as follows: 

(1) We introduce \textbf{MentalDx Bench}, the first benchmark dedicated to fine-grained, disorder-level diagnosis grounded in real-world clinical records. Expert-annotated under ICD-11 guidelines, it spans 76 disorders across 16 categories, bridging the gap between computational metrics and ecological clinical validity.

(2) Our systematic evaluation of 18 LLMs quantifies a critical \textbf{paradigm misalignment}: current models rely on superficial pattern associations rather than the structured clinical reasoning essential for psychiatry, resulting in significant performance collapse on fine-grained inference.

(3) We propose \textbf{MentalSeek-Dx}, a domain-specialized framework that aligns LLMs with clinical logic via progressive hypothetico-deductive reasoning and curriculum reinforcement learning. It achieves SOTA performance with only 14B parameters, demonstrating that reasoning alignment is the key to reliable psychiatric diagnosis.

\newcommand{\symScale}{0.90}

\newcommand{\Yes}{\raisebox{0.15ex}{\scalebox{\symScale}{\textcolor{green!70!black}{\ding{51}}}}}
\newcommand{\No}{\raisebox{0.10ex}{\scalebox{\symScale}{\textcolor{red!70!black}{\ding{55}}}}}

\newcommand{\HalfYes}{%
  \raisebox{0.10ex}{%
    \scalebox{\symScale}{%
      \tikz[baseline=-0.55ex, scale=0.70]{
        \draw[line width=0.85pt, line cap=round]
          (0,0) -- (0.25,-0.25) -- (0.75,0.35);
        \draw[line width=0.85pt, line cap=round]
          (0.45,0.15) -- (0.65,-0.05);
      }%
    }%
  }%
}

\begin{table}[t!]
\centering
{\small
\setlength{\tabcolsep}{3pt}
\renewcommand{\arraystretch}{1.15}

\resizebox{\columnwidth}{!}{%
\begin{tabular}{@{}
l
@{\hspace{6pt}}
c
@{\hspace{6pt}}
c
@{\hspace{6pt}}
c
@{\hspace{6pt}}
c
@{}}
\toprule
\textbf{Benchmark} & \textbf{Input} & \textbf{Scope} & \textbf{Hierarchy} & \textbf{Types} \\
\midrule

\makecell[l]{MindSET \citep{mankarious2025mindset}}
& Narrative  & Mental Health  & \cellcolor[HTML]{f2f7fc} Stratification & 7        \\

\makecell[l]{CRADLE Bench \citep{byun2025cradle}}
& Narrative  & Mental Health  & \cellcolor[HTML]{f2f7fc} Stratification & 7        \\

\makecell[l]{ANGST  \citep{hengle2024still}}
& Narrative  & Mood  & \cellcolor[HTML]{f2f7fc} Stratification  & 2 \\

\makecell[l]{IMHI Benchmark  \citep{yang2024mentallama}}
& Narrative  & Mood  & \cellcolor[HTML]{f2f7fc} Stratification  & 9 \\

\makecell[l]{CLPsych  \citep{kim2025baseline}}
& Narrative  & Suicide  & \cellcolor[HTML]{EDF3FB} Risk &  1 \\

\makecell[l]{D4  \citep{yao2022d4}}
& Narrative     & Depression  & \cellcolor[HTML]{e0e9f7} Symptom  & 2 \\

\makecell[l]{ReDSM5  \citep{bao2025redsm5}}
& Narrative  & Depression & \cellcolor[HTML]{e0e9f7} Symptom & 1 \\

\makecell[l]{MDPP  \citep{fu2025first}}
& Narrative  & Depression & \cellcolor[HTML]{e0e9f7} Symptom & 1 \\

\makecell[l]{EmoDep  \citep{oh2026clinically}}
& Clinical Note & Depression    & \cellcolor[HTML]{e0e9f7} Symptom    & 1 \\

\makecell[l]{SMDH  \citep{cohan2018smhd}}
& Narrative     & Psychiatry   & \cellcolor[HTML]{d7d4e9} Disorder& 9 \\

\makecell[l]{PsychiatryBench\citep{fouda2025psychiatrybench} }
& Book & Psychiatry & \cellcolor[HTML]{d7d4e9} Category+Disorder & 50\\

\midrule
\makecell[l]{\textbf{MentalDx Bench \textit{(Ours)}} }
& EHR & Psychiatry & \cellcolor[HTML]{d7d4e9} Category+Disorder    & 76\\

\bottomrule
\end{tabular}%
}
}
\caption{Comparison of representative benchmarks for mental health diagnosis, where our \textbf{MentalDx Bench} features realistic EHR inputs and dual-level diagnostic labels at the category and disorder levels.}
\label{tab:mh_benchmarks_compact}
\end{table}

\section{Related Work}
\textbf{Psychiatric Benchmarks.}
Psychiatric NLP benchmarks have progressed from disorder-level classification of social media texts \citep{cohan2018smhd} to symptom-level evaluation with clinically grounded labels \citep{yao2022d4,bao2025redsm5,fu2025first}. With LLMs, benchmarks increasingly adopt structured diagnostic setups—first in narrow domains \citep{hengle2024still,yang2024mentallama}, then across broader taxonomies and reasoning chains \citep{mankarious2025mindset,byun2025cradle,fouda2025psychiatrybench}—expanding into suicide risk detection \citep{kim2025baseline} and large-scale user-level corpora \cite{zheng2025rsd}. Some ground evaluation in real-world clinical decision-making \cite{liu2025psychbench}, or simulate multi-step psychiatric workflows \cite{johri2025evaluation}. Input modalities have shifted from narratives to clinical notes \citep{oh2026clinically}, enabling EHR-based modeling for outcome prediction~\cite{sao2024miroberta}, phenotype extraction~\cite{frydman2025large}, and clinical reasoning~\cite{zhu2025medtpe}. However, most benchmarks remain based on non-clinical data, limiting clinical representativeness.
 
\paragraph{Psychiatric Methods.} Large language models (LLMs) have emerged as promising tools to support psychiatric assessment \citep{ge2025survey}. Early applications focused on detecting subtle indicators of psychological distress \cite{lho2025large}, enabling risk screening \citep{shin2024using,chim2024overview}, subclinical symptom identification, and population-level stratification \citep{obradovich2024opportunities,orru2025large}. More recently, research has shifted toward diagnostic prediction, with two main approaches: one uses LLM-derived features combined with downstream classifiers \citep{radwan2024predictive,shewcraft2025algorithmic,lan2025depression}, and the other trains domain-specific LLMs in an end-to-end fashion \citep{xu2024mental,weber2025using,dai2025psyche}. This line of work reflects growing interest in deploying LLMs to assist Psychiatric clinical decision-making. However, disorder-level, general-purpose psychiatric diagnostic modeling remains underexplored.

\begin{table*}[t]

\definecolor{TopOne_2}{HTML}{ebd0dc}
\definecolor{TopTwo_2}{HTML}{EFE4E8}
\definecolor{Rest_2}{HTML}{f8f5fb}

\definecolor{TopOne_3}{HTML}{c8def7}
\definecolor{TopTwo_3}{HTML}{e0e9f7}
\definecolor{Rest_3}{HTML}{f2f7fc}

\definecolor{TopOne_1}{HTML}{c0bbd8}
\definecolor{TopTwo_1}{HTML}{d7d4e9}
\definecolor{Rest_1}{HTML}{f0eff8}

\centering
\newcolumntype{S}{>{\small}p{3.2cm}}
\centering
\renewcommand{\arraystretch}{1.25}
\resizebox{0.95 \textwidth}{!}{
\addtolength{\tabcolsep}{-1.5pt}
\begin{tabular}{S|ccccccccccccccccccc}
\hline
\textbf{Model}& \textit{ANX} & \textit{CATA} & \textit{SUD} & \textit{BOD} & \textit{STRESS}& \textit{DISR} & \textit{DISS} & \textit{ELIM} & \textit{EAT} & \textit{PREG}& \textit{MOOD} & \textit{NCOG} & \textit{NDEV} & \textit{OCD} & \textit{PERS} & \textit{SCHIZ}  & $\mathcal{CA}$ & $\mathcal{DA}$ & $\mathcal{JA}$ \\
\hline
\multicolumn{20}{l}{\textit{\textbf{Small Language Models}}} \\
\hline
Qwen2.5-7B  & \cellcolor{Rest_1}11.11 & \cellcolor{Rest_1}30.00 & \cellcolor{Rest_1}14.12 & \cellcolor{TopTwo_1}10.00 & \cellcolor{TopTwo_1}40.00 & \cellcolor{Rest_1}0.00 & \cellcolor{Rest_1}4.26 & \cellcolor{TopTwo_1}25.00 & \cellcolor{Rest_1}17.14 & \cellcolor{Rest_1}21.74 & \cellcolor{Rest_1}12.37 & \cellcolor{TopTwo_1}13.79 & \cellcolor{Rest_1}18.92 & \cellcolor{Rest_1}12.50 & \cellcolor{TopTwo_1}0.00 & \cellcolor{TopTwo_1}12.50 & \cellcolor{Rest_1}47.47 & \cellcolor{Rest_1}14.75 & \cellcolor{Rest_1}10.96\\

Qwen2.5-14B  & \cellcolor{Rest_1}18.06 & \cellcolor{TopTwo_1}63.33 & \cellcolor{TopTwo_1}29.41 & \cellcolor{Rest_1}0.00 & \cellcolor{TopOne_1}52.50 & \cellcolor{Rest_1}44.44 & \cellcolor{Rest_1}21.28 & \cellcolor{TopTwo_1}25.00 & \cellcolor{TopTwo_1}62.86 & \cellcolor{TopTwo_1}82.61 & \cellcolor{TopTwo_1}22.68 & \cellcolor{TopTwo_1}13.79 & \cellcolor{Rest_1}48.65 & \cellcolor{Rest_1}40.00 & \cellcolor{TopTwo_1}0.00 & \cellcolor{Rest_1}10.94  & \cellcolor{TopTwo_1}61.66 & \cellcolor{Rest_1}29.92 & \cellcolor{Rest_1}24.72\\

Qwen3-32B  & \cellcolor{TopTwo_1}30.56 & \cellcolor{Rest_1}20.00 & \cellcolor{Rest_1}22.35 & \cellcolor{TopOne_1}25.00 & \cellcolor{Rest_1}32.50 & \cellcolor{TopTwo_1}50.00 & \cellcolor{TopTwo_1}31.91 & \cellcolor{TopOne_1}75.00 & \cellcolor{Rest_1}48.57 & \cellcolor{Rest_1}52.17 & \cellcolor{TopTwo_1}22.68 & \cellcolor{TopOne_1}35.63 & \cellcolor{TopTwo_1}51.35 & \cellcolor{TopTwo_1}62.50 & \cellcolor{TopOne_1}15.38 & \cellcolor{Rest_1}1.56 & \cellcolor{Rest_1}58.71 & \cellcolor{TopTwo_1}31.04 & \cellcolor{TopTwo_1}26.26 \\
QWQ-32B & \cellcolor{TopOne_1}48.61 & \cellcolor{TopOne_1}70.00 & \cellcolor{TopOne_1}35.29 & \cellcolor{TopOne_1}25.00 & \cellcolor{Rest_1}35.00 & \cellcolor{TopOne_1}66.67 & \cellcolor{TopOne_1}42.55 & \cellcolor{TopOne_1}75.00 & \cellcolor{TopOne_1}91.43 & \cellcolor{TopOne_1}91.30 & \cellcolor{TopOne_1}41.24 & \cellcolor{TopOne_1}35.63 & \cellcolor{TopOne_1}75.68 & \cellcolor{TopOne_1}75.00 & \cellcolor{TopTwo_1}0.00 & \cellcolor{TopOne_1}29.69   & \cellcolor{TopOne_1}76.26 & \cellcolor{TopOne_1}47.89 & \cellcolor{TopOne_1}46.63 \\

\hline
\multicolumn{20}{l}{\textit{\textbf{Large Language Models}}} \\
\hline

LLaMA-3.1-70B   & \cellcolor{Rest_2}47.22 & \cellcolor{Rest_2}30.00 & \cellcolor{Rest_2}28.24 & \cellcolor{Rest_2}5.00 & \cellcolor{Rest_2}35.00 & \cellcolor{TopTwo_2}88.89 & \cellcolor{Rest_2}27.66 & \cellcolor{TopTwo_2}75.00 & \cellcolor{Rest_2}62.86 & \cellcolor{Rest_2}65.22 & \cellcolor{Rest_2}25.77 & \cellcolor{Rest_2}25.29 & \cellcolor{Rest_2}64.86 & \cellcolor{Rest_2}47.50 & \cellcolor{Rest_2}7.69 & \cellcolor{Rest_2}17.19  & \cellcolor{Rest_2}70.65 & \cellcolor{Rest_2}35.53 & \cellcolor{Rest_2}33.15\\
Qwen2.5-72B  & \cellcolor{Rest_2}34.72 & \cellcolor{Rest_2}40.00 & \cellcolor{Rest_2}32.94 & \cellcolor{Rest_2}0.00 & \cellcolor{Rest_2}35.00 & \cellcolor{Rest_2}38.89 & \cellcolor{Rest_2}34.04 & \cellcolor{TopTwo_2}75.00 & \cellcolor{Rest_2}74.29 & \cellcolor{Rest_2}78.26 & \cellcolor{Rest_2}23.71 & \cellcolor{Rest_2}16.09 & \cellcolor{Rest_2}62.16 & \cellcolor{Rest_2}55.00 & \cellcolor{TopOne_2}30.77 & \cellcolor{Rest_2}6.25 & \cellcolor{Rest_2}70.51 & \cellcolor{Rest_2}33.57 & \cellcolor{Rest_2}29.78 \\
Kimi-K2  & \cellcolor{Rest_2}55.56 & \cellcolor{Rest_2}33.33 & \cellcolor{Rest_2}44.71 & \cellcolor{Rest_2}0.00 & \cellcolor{TopTwo_2}57.50 & \cellcolor{Rest_2}72.22 & \cellcolor{Rest_2}48.94 & \cellcolor{TopOne_2}100.00 & \cellcolor{TopTwo_2}91.43 & \cellcolor{Rest_2}65.22 & \cellcolor{Rest_2}24.74 & \cellcolor{Rest_2}32.18 & \cellcolor{TopOne_2}89.19 & \cellcolor{Rest_2}70.00 & \cellcolor{TopTwo_2}23.08 & \cellcolor{Rest_2}10.94 & \cellcolor{Rest_2}75.70 & \cellcolor{Rest_2}45.08 & \cellcolor{Rest_2}43.40\\
Qwen3-Max  & \cellcolor{Rest_2}50.00 & \cellcolor{TopTwo_2}73.33 & \cellcolor{TopTwo_2}47.06 & \cellcolor{TopOne_2}20.00 & \cellcolor{Rest_2}52.50 & \cellcolor{TopOne_2}94.44 & \cellcolor{Rest_2}48.94 & \cellcolor{TopOne_2}100.00 & \cellcolor{TopTwo_2}91.43 & \cellcolor{TopTwo_2}86.96 & \cellcolor{Rest_2}32.99 & \cellcolor{TopTwo_2}48.28 & \cellcolor{TopOne_2}89.19 & \cellcolor{Rest_2}82.50 & \cellcolor{TopTwo_2}23.08 & \cellcolor{Rest_2}20.31 & \cellcolor{TopOne_2}83.15 & \cellcolor{Rest_2}52.67 & \cellcolor{Rest_2}52.11 \\
GPT-5.1  & \cellcolor{TopTwo_2}66.67 & \cellcolor{Rest_2}53.33 & \cellcolor{TopTwo_2}47.06 & \cellcolor{Rest_2}5.00 & \cellcolor{Rest_2}42.50 & \cellcolor{Rest_2}77.78 & \cellcolor{TopTwo_2}53.19 & \cellcolor{TopOne_2}100.00 & \cellcolor{TopTwo_2}91.43 & \cellcolor{Rest_2}82.61 & \cellcolor{Rest_2}38.14 & \cellcolor{TopTwo_2}48.28 & \cellcolor{TopTwo_2}78.38 & \cellcolor{TopTwo_2}85.00 & \cellcolor{TopOne_2}30.77 & \cellcolor{TopOne_2}40.62 & \cellcolor{TopTwo_2}81.18 & \cellcolor{Rest_2}54.49 & \cellcolor{Rest_2}53.79 \\
DeepSeek-R1  & \cellcolor{Rest_2}50.00 & \cellcolor{Rest_2}63.33 & \cellcolor{TopOne_2}48.24 & \cellcolor{TopTwo_2}15.00 & \cellcolor{TopOne_2}60.00 & \cellcolor{Rest_2}77.78 & \cellcolor{TopOne_2}61.70 & \cellcolor{TopOne_2}100.00 & \cellcolor{TopOne_2}94.29 & \cellcolor{TopOne_2}95.65 & \cellcolor{TopOne_2}45.36 & \cellcolor{Rest_2}40.23 & \cellcolor{TopTwo_2}78.38 & \cellcolor{TopTwo_2}85.00 & \cellcolor{Rest_2}15.38 & \cellcolor{TopTwo_2}37.50& \cellcolor{Rest_2}78.37 & \cellcolor{TopTwo_2}55.20 & \cellcolor{TopTwo_2}54.49 \\
Claude-Sonnet-4.5  & \cellcolor{TopOne_2}88.89 & \cellcolor{TopOne_2}83.33 & \cellcolor{TopTwo_2}47.06 & \cellcolor{TopTwo_2}15.00 & \cellcolor{Rest_2}52.50 & \cellcolor{TopTwo_2}88.89 & \cellcolor{TopOne_2}61.70 & \cellcolor{TopOne_2}100.00 & \cellcolor{Rest_2}85.71 & \cellcolor{TopOne_2}95.65 & \cellcolor{TopTwo_2}40.21 & \cellcolor{TopOne_2}49.43 & \cellcolor{TopTwo_2}78.38 & \cellcolor{TopOne_2}87.50 & \cellcolor{TopTwo_2}23.08 & \cellcolor{Rest_2}34.38 & \cellcolor{Rest_2}80.76 & \cellcolor{TopOne_2}59.69 & \cellcolor{TopOne_2}58.15 \\

\hline
\multicolumn{20}{l}{\textit{\textbf{Medical-specific Large Language Models}}} \\
\hline

ClinicalGPT-R1  & \cellcolor{Rest_3}9.72 & \cellcolor{Rest_3}23.33 & \cellcolor{Rest_3}10.59 & \cellcolor{Rest_3}5.00 & \cellcolor{Rest_3}27.50 & \cellcolor{Rest_3}5.56 & \cellcolor{Rest_3}2.13 & \cellcolor{Rest_3}0.00 & \cellcolor{Rest_3}14.29 & \cellcolor{Rest_3}17.39 & \cellcolor{Rest_3}16.49 & \cellcolor{Rest_3}5.75 & \cellcolor{Rest_3}13.51 & \cellcolor{Rest_3}5.00 & \cellcolor{Rest_3}0.00 & \cellcolor{Rest_3}1.56 & \cellcolor{Rest_3}36.10 & \cellcolor{Rest_3}10.53 & \cellcolor{Rest_3}6.46\\
HuatuoGPT-o1-7B  & \cellcolor{Rest_3}9.72 & \cellcolor{Rest_3}20.00 & \cellcolor{Rest_3}12.94 & \cellcolor{Rest_3}5.00 & \cellcolor{Rest_3}25.00 & \cellcolor{Rest_3}5.56 & \cellcolor{Rest_3}0.00 & \cellcolor{Rest_3}50.00 & \cellcolor{Rest_3}5.71 & \cellcolor{Rest_3}8.70 & \cellcolor{Rest_3}15.46 & \cellcolor{Rest_3}6.90 & \cellcolor{Rest_3}5.41 & \cellcolor{Rest_3}10.00 & \cellcolor{Rest_3}0.00 & \cellcolor{Rest_3}17.19 & \cellcolor{Rest_3}43.26 & \cellcolor{Rest_3}11.24 & \cellcolor{Rest_3}8.01\\
GPT-OSS-20B  & \cellcolor{TopTwo_3}52.78 & \cellcolor{TopTwo_3}63.33 & \cellcolor{Rest_3}23.53 & \cellcolor{Rest_3}15.00 & \cellcolor{Rest_3}35.00 & \cellcolor{TopTwo_3}83.33 & \cellcolor{TopTwo_3}44.68 & \cellcolor{TopOne_3}100.00 & \cellcolor{Rest_3}77.14 & \cellcolor{TopTwo_3}82.61 & \cellcolor{TopTwo_3}34.02 & \cellcolor{Rest_3}32.18 & \cellcolor{TopTwo_3}70.27 & \cellcolor{TopTwo_3}72.50 & \cellcolor{Rest_3}0.00 & \cellcolor{TopOne_3}45.31 & \cellcolor{TopTwo_3}74.30 & \cellcolor{TopTwo_3}45.65 & \cellcolor{TopTwo_3}43.96\\
MedGemma-27B  & \cellcolor{TopOne_3}54.17 & \cellcolor{Rest_3}50.00 & \cellcolor{Rest_3}28.24 & \cellcolor{TopOne_3}25.00 & \cellcolor{TopTwo_3}42.50 & \cellcolor{Rest_3}55.56 & \cellcolor{Rest_3}29.79 & \cellcolor{TopTwo_3}75.00 & \cellcolor{Rest_3}74.29 & \cellcolor{Rest_3}65.22 & \cellcolor{Rest_3}25.77 & \cellcolor{Rest_3}32.18 & \cellcolor{TopOne_3}72.97 & \cellcolor{Rest_3}70.00 & \cellcolor{Rest_3}23.08 & \cellcolor{TopTwo_3}40.62 & \cellcolor{Rest_3}66.99 & \cellcolor{Rest_3}42.84 & \cellcolor{Rest_3}39.89\\
Baichuan-M2  & \cellcolor{Rest_3}38.89 & \cellcolor{TopOne_3}66.67 & \cellcolor{TopTwo_3}35.29 & \cellcolor{Rest_3}5.00 & \cellcolor{Rest_3}37.50 & \cellcolor{Rest_3}72.22 & \cellcolor{Rest_3}42.55 & \cellcolor{TopTwo_3}75.00 & \cellcolor{TopTwo_3}88.57 & \cellcolor{TopOne_3}91.30 & \cellcolor{Rest_3}28.87 & \cellcolor{TopTwo_3}37.93 & \cellcolor{Rest_3}62.16 & \cellcolor{Rest_3}67.50 & \cellcolor{Rest_3}0.00 & \cellcolor{Rest_3}35.94 & \cellcolor{Rest_3}64.61 & \cellcolor{Rest_3}44.38 & \cellcolor{Rest_3}41.71\\
HuatuoGPT-o1-72B  & \cellcolor{Rest_3}45.83 & \cellcolor{Rest_3}50.00 & \cellcolor{TopTwo_3}35.29 & \cellcolor{Rest_3}10.00 & \cellcolor{Rest_3}35.00 & \cellcolor{Rest_3}66.67 & \cellcolor{Rest_3}40.43 & \cellcolor{TopTwo_3}75.00 & \cellcolor{Rest_3}80.00 & \cellcolor{Rest_3}69.57 & \cellcolor{Rest_3}26.80 & \cellcolor{Rest_3}19.54 & \cellcolor{TopTwo_3}70.27 & \cellcolor{Rest_3}65.00 & \cellcolor{TopTwo_3}30.77 & \cellcolor{Rest_3}18.75 & \cellcolor{Rest_3}74.02 & \cellcolor{Rest_3}39.75 & \cellcolor{Rest_3}36.94\\
GPT-OSS-120B  & \cellcolor{TopOne_3}54.17 & \cellcolor{TopOne_3}66.67 & \cellcolor{TopOne_3}38.82 & \cellcolor{TopTwo_3}20.00 & \cellcolor{TopOne_3}50.00 & \cellcolor{TopOne_3}88.89 & \cellcolor{TopOne_3}51.06 & \cellcolor{Rest_3}50.00 & \cellcolor{TopOne_3}94.29 & \cellcolor{Rest_3}65.22 & \cellcolor{TopOne_3}41.24 & \cellcolor{TopOne_3}43.68 & \cellcolor{TopTwo_3}70.27 & \cellcolor{TopOne_3}75.00 & \cellcolor{TopOne_3}38.46 & \cellcolor{TopOne_3}45.31 & \cellcolor{TopOne_3}75.84 & \cellcolor{TopOne_3}52.53 & \cellcolor{TopOne_3}50.00\\

\hline

\end{tabular}
}
\caption{Performance of mainstream LLMs on MentalDx Bench (\%). Models are grouped by parameter scale to examine how size affects performance across diagnostic categories. Each cell reports a model’s score on the corresponding task. For clarity, the best result in each category is shown in dark, the second-best in light.}
\label{tab:llm_performance}
\end{table*}

\section{MentalDx Bench}
We introduce MentalDx Bench, the first benchmark specifically designed for disorder-level psychiatric diagnosis across the full spectrum of psychiatric conditions, featuring real-world clinical data and expert annotations strictly aligned with guidelines:

\begin{figure}[t]
  \includegraphics[width=\linewidth]{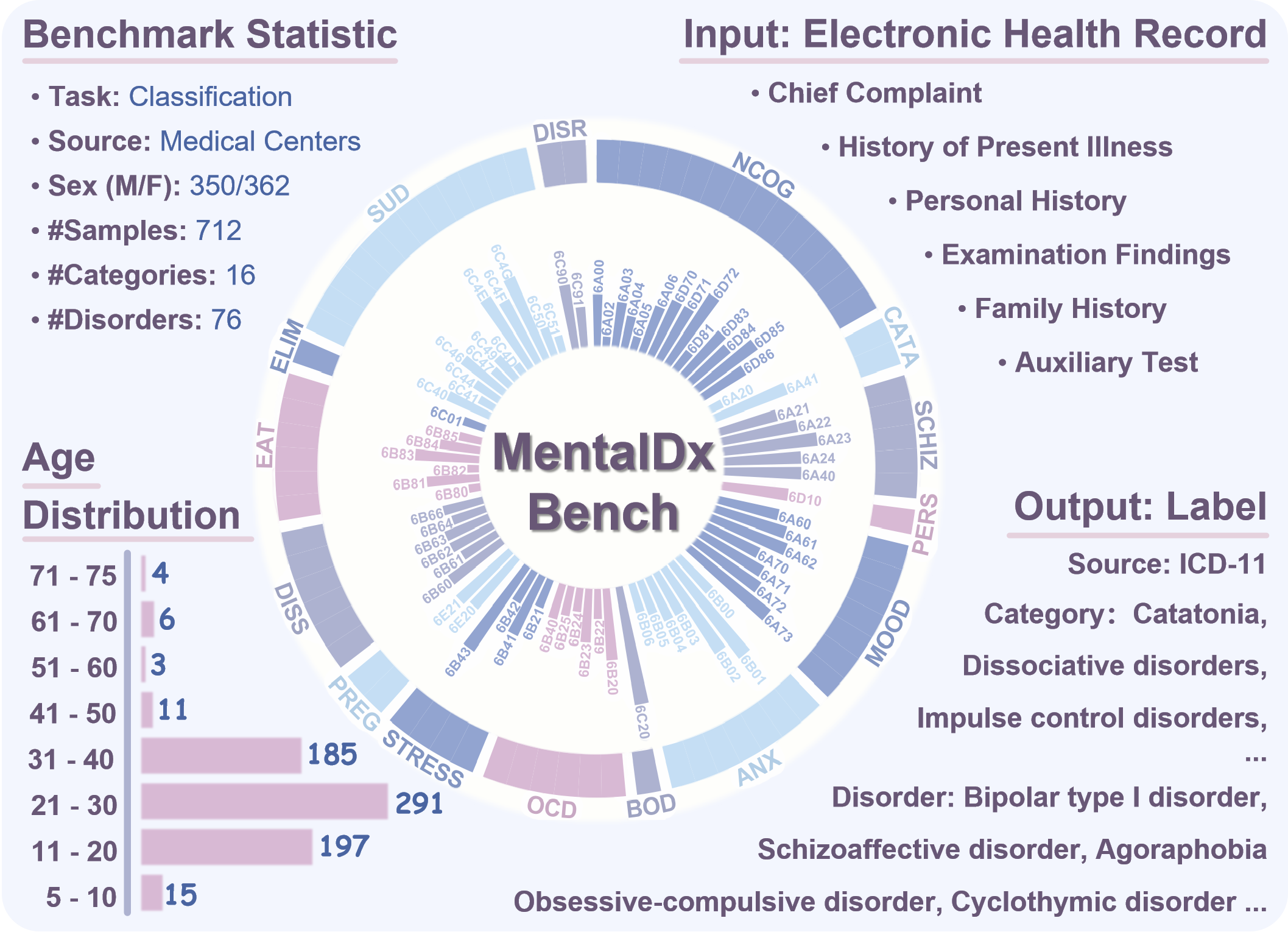}
\caption{Overview of the \textit{MentalDx Benchmark}, based on real-world medical centers, showing benchmark statistics, inputs, outputs, and data distributions.}
  \label{fig_2_statistic.png}
\end{figure}

\textbf{(1) Data Source.} The \textit{MentalDx Bench} consists of 712 de-identified psychiatric diagnostic records sourced from partner medical centers. To ensure data quality and diagnostic integrity, we exclude records with incomplete workups, ambiguous conclusions, comorbidities, or sensitive personal content. All data processing is conducted entirely within the hospital’s secure infrastructure.

\textbf{(2) Ethical Compliance.} The study protocol and annotation procedures were reviewed and approved by the institutional ethics boards of all participating clinical centers, ensuring the legal and ethical integrity. All data collection and labeling were conducted in accordance with the ethical principles of the \textit{Declaration of Helsinki}. As all records were fully de-identified prior to processing, the requirement for individual informed consent was waived.

\textbf{(3) Annotation Schema.} To ensure standardized labeling, all cases in \textit{MentalDx Bench} are annotated according to the ICD-11 taxonomy for \textit{Mental, Behavioural, and Neurodevelopmental Disorders} (Chapter 6A), using both official diagnostic names and ICD codes. A team of 12 board-certified psychiatrists conducted the labeling process following ICD-11 guidelines and internal consensus protocols. Each case receives two levels of diagnostic labels: a Category-level label denoting the broader group (e.g., Mood Disorders) and a Disorder-level label specifying the exact condition (e.g., Schizophrenia). All annotations are independently assigned by two senior psychiatrists, with inter-rater agreement measured by Cohen’s $\kappa$ coefficient ($\kappa > 0.8$). Disagreements are resolved through expert consensus meetings. Cases involving comorbidity are excluded to ensure label clarity.

\textbf{(4) Dataset Characteristics.} \textit{MentalDx Bench} is designed to support structured, fine-grained diagnostic evaluation while capturing real-world psychiatric diversity across age groups and genders. To ensure balanced label distribution and mitigate long-tail effects, we sample a uniform number of cases, as shown in Fig~\ref{fig_2_statistic.png}. Each case contains a structured clinical narrative composed of multiple sections—chief complaint, history of present illness, personal and family history, examination findings, and auxiliary test results—providing rich contextual input for diagnostic modeling. Each case is labeled with a single disorder and its parent category, forming a two-level labeling structure used in a single-label classification setup. Details and the ethics statement can be found in Appendix~\ref{ethics_statement}.

\begin{figure}[t!]
  \centering
  \includegraphics[width=\linewidth]{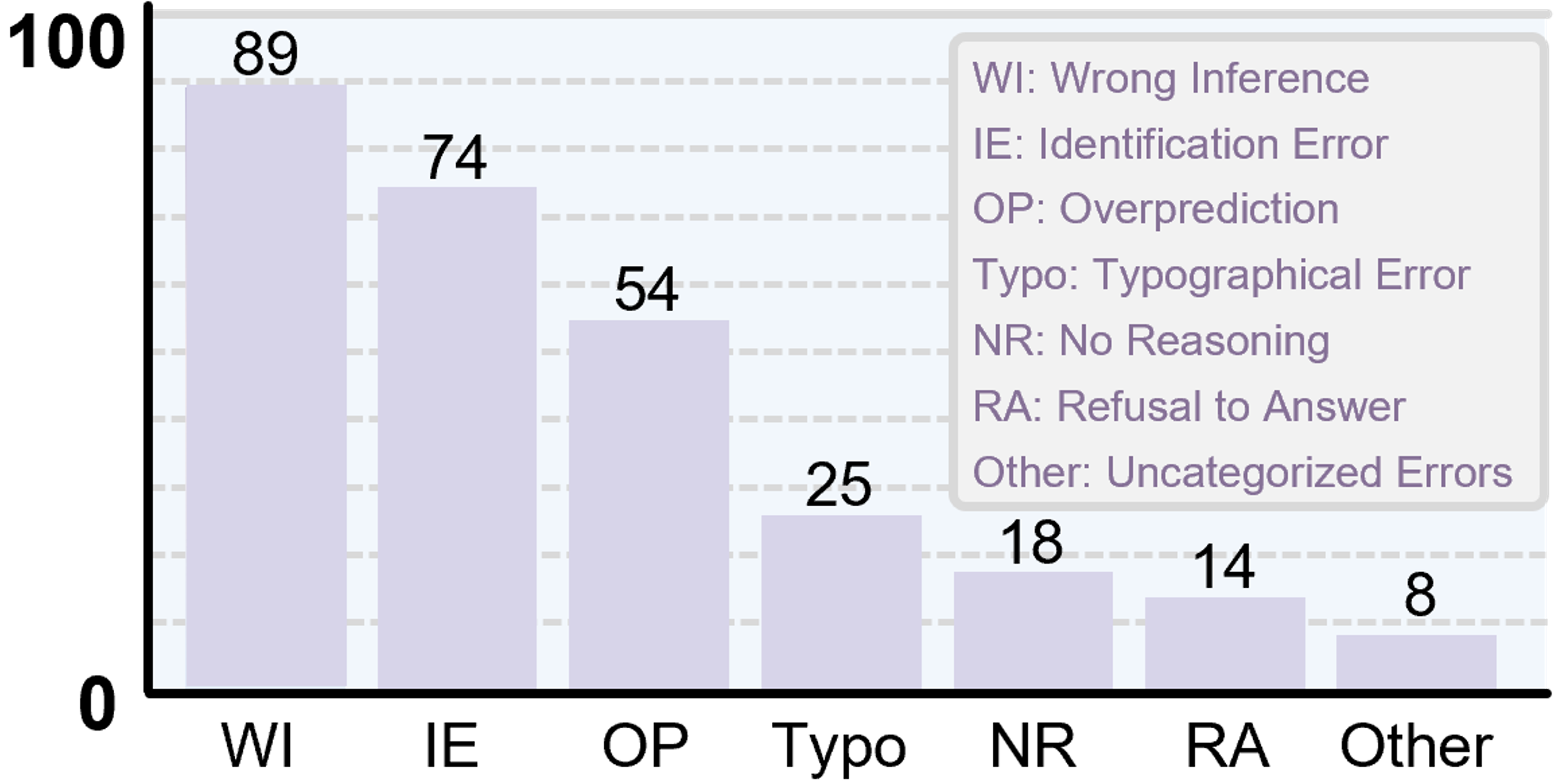}
\caption{Statistics of expert-annotated errors across 360 cases. Results highlight a paradigm misalignment between LLM reasoning and clinical diagnostic logic.}
  \label{fig_annotation.png}
\end{figure}

\section{Experiments on MentalDx Bench}

\subsection{Experiment Settings}
We present the task definition, evaluation metrics, and evaluated models for MentalDx Bench.

\noindent\textbf{Task Definition.}
After removing comorbid cases during preprocessing, the task is formulated as a multi-stage, single-label classification problem under a constrained candidate set.
Given a clinical case $x$ and a predefined list of ICD-11 categories and disorders, the model $M$ is required to predict the most likely category $c$ and the corresponding disorder $d$ based on the case description.

\noindent\textbf{Evaluation Metrics.}
We evaluate model performance using three metrics: \textit{Category Accuracy (CA)}, \textit{Disorder Accuracy (DA)}, and \textit{Joint Accuracy (JA)}.
\textit{CA} and \textit{DA} measure whether the predicted category or disorder matches the ground truth, respectively, while \textit{JA} requires both to be correct simultaneously. Let $(c_i, d_i)$ denote the \textit{gt} category and disorder for the $i$-th sample, and $(\hat{c}_i, \hat{d}_i)$ be the predictions. The metrics are defined as:
\begin{equation}
\small
\begin{aligned}
\mathcal{CA}, \mathcal{DA} &= \frac{1}{N} \sum_{i=1}^{N} \mathbb{1}[\hat{y}_i = y_i], \\
\mathcal{JA} &= \frac{1}{N} \sum_{i=1}^{N} \mathbb{1}[\hat{c}_i = c_i \land \hat{d}_i = d_i].
\end{aligned}
\end{equation}
where $N$ is the total number of test samples and $\mathbb{1}[\cdot]$ denotes the indicator function. With candidates provided in the input, we extract predictions through exact string matching.


\noindent\textbf{Baselines.}
We evaluate 18 language models spanning 3 categories as follows: 1) Small Language Models, including:
Qwen2.5-7B and 14B\citep{team2024qwen2}
Qwen3-32B\citep{yang2025qwen3}, and QWQ-32B\citep{team2024qwen2}. 2) {LLMs}, including:
LLaMA-3.1-70B\citep{patterson2022carbon},
Qwen2.5-72B\citep{team2024qwen2},
Qwen3-Max (2025-10-30)\citep{yang2025qwen3},
Kimi-K2 (2025-09-05)\citep{team2025kimi},
DeepSeek-R1(2025-05-28)\citep{guo2025deepseek},GPT-5.1(2025-11-13)\citep{bubeck2025early}, Claude-Sonnet-4.5(2025-09-29)\citep{appel2025anthropic}. 3) Medical-specific LLMs, including:
ClinicalGPT-R1\citep{lan2025clinicalgpt},
HuatuoGPT-O1-7B and 72B\citep{chen2024huatuogpt},GPT-OSS-20B and 120B\citep{agarwal2025gpt},
Medgemma-27B\citep{sellergren2025medgemma}, Baichuan-M2\citep{dou2025baichuan}.

\subsection{Experiment Results}

We aim to address the following research questions:

\vspace{-4pt}
\begin{itemize}[leftmargin=*]
  \setlength\itemsep{0pt}
  \setlength\parskip{0pt}
  \setlength\topsep{0pt}
  \item \textbf{RQ1:} How well do existing models perform on psychiatric diagnosis?
\item \textbf{RQ2:} What are the underlying causes of model limitations?
\vspace{-2pt}
\end{itemize}

\subsubsection{RQ1: How well do existing models perform on psychiatric diagnosis?}

We evaluate 18 LLMs on MentalDx Bench as shown in Table~\ref{tab:llm_performance} (The full category list and disorder explanations are provided in Appendix~\ref{catogory_and_disorder_list}.), and find that most perform well on category prediction—for example, QWQ-32B achieves 76.26\%, Qwen3-Max 83.15\%, and GPT-OSS-120B 75.84\%. However, some medical-specific models lag behind their general counterparts of similar scale; ClinicalGPT-R1 reaches only 36.10\%, which is 11.37 percentage points lower than its base model Qwen2.5-7B (47.47\%). Even with medical pretraining, gaps in coverage and distributional mismatch hinder generalization to psychiatric tasks.

While existing LLMs perform well on category-level diagnosis, their performance drops markedly at the disorder level. Qwen3-Max, despite achieving the highest \textit{CA}, reaches only 52.67\% on disorder prediction; ClinicalGPT-R1 performs even worse, with a \textit{JA} of just 6.46\%. Coarse-grained evaluation can obscure fine-grained diagnostic failures, motivating the need for fine-grained benchmarks. See Appendix~\ref{sec:Example} for examples.

\subsubsection{RQ2: What are the underlying causes of model limitations?}

We conducted expert validation to investigate diagnostic errors. For each of the 18 models, we sampled 20 incorrect predictions (360 total) and assigned 12 psychiatrists to independently assess a random subset of 30 cases. As shown in Figure~\ref{fig_annotation.png}, most errors fall into three categories—\textit{wrong inference} (25\%), \textit{identification failure} (22\%), and \textit{symptom omission} (21\%)—whereas superficial issues like typos (7\%) or refusals (4\%) were far less common. This distribution suggests that the dominant failure modes stem not from surface-level mistakes, but from deeper deficits in reasoning and symptom understanding. In particular, many models exhibited fragmented diagnostic logic, including unclear symptom delineation, incomplete analysis of candidate conditions, and a lack of differential diagnostic ability. Error definitions and procedures are described in Appendix~\ref{error_types}.

\begin{figure}[t!]
  \includegraphics[width=\linewidth]{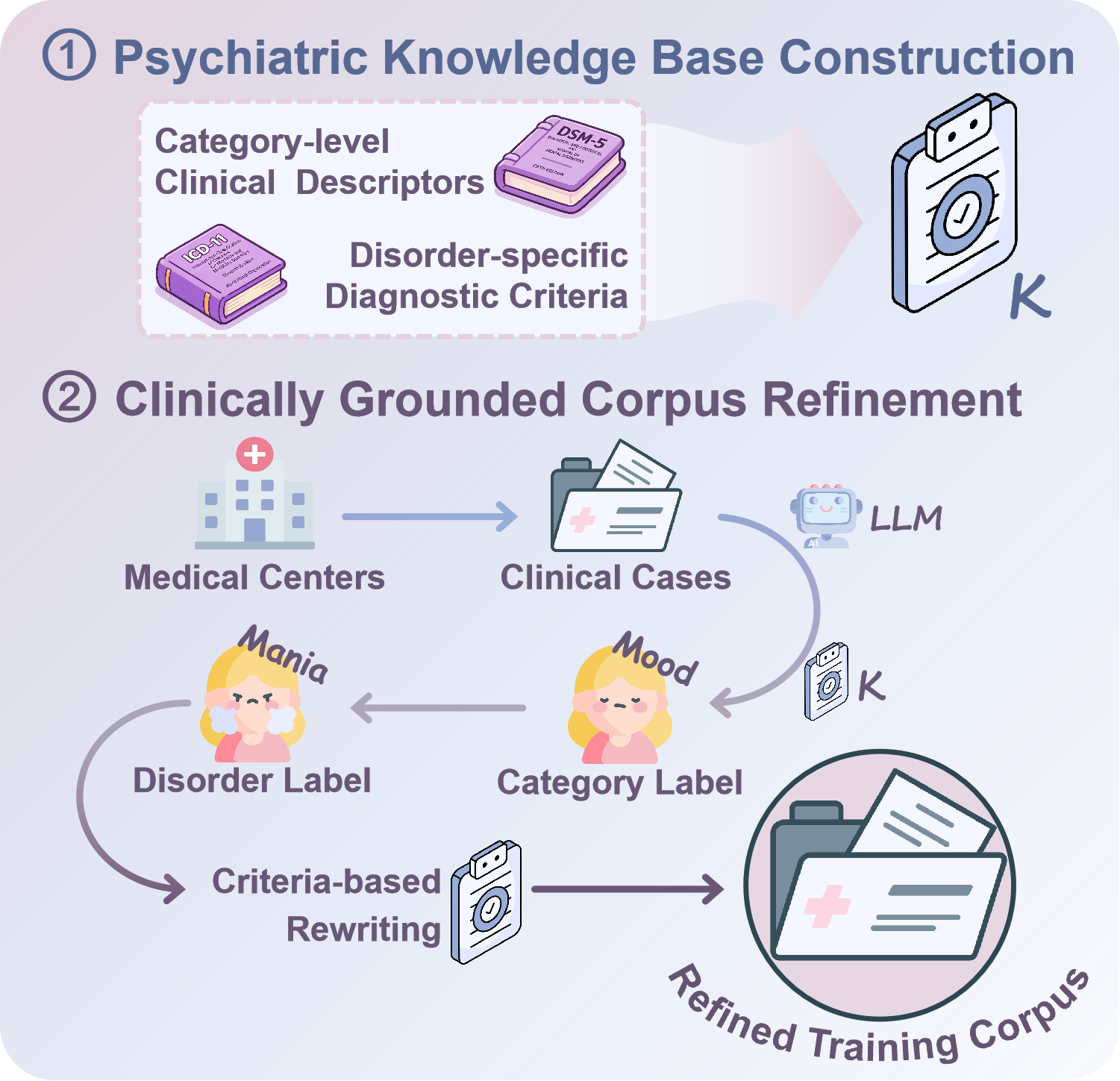}
\caption{Overview of psychiatric knowledge base construction and clinically grounded corpus refinement.}
  \label{fig_3_train_corpus.png}
\end{figure}

\section{MentalSeek-Dx}

Motivated by the above findings, we introduce \textsc{MentalSeek-Dx}, a medical-specific LLM for psychiatric reasoning diagnosis. 

\subsection{Pretraining Setup}
\noindent\textbf{Structured Psychiatric Knowledge Base.}  
We construct a structured knowledge base $\mathcal{K}$ in collaboration with psychiatrists to support diagnostic reasoning and model training. $\mathcal{K}$ follows a two-level hierarchy of diagnostic categories $c$ and fine-grained disorders $d$, with each entry comprising clinical definitions, symptom clusters, functional impairments, diagnostic criteria, and differential considerations. Its design follows ICD-11, clinical guidelines, and expert consensus, ensuring coverage and clinical validity. Given a $c$ or $d$, $\mathcal{K}$ enables automatic retrieval of relevant knowledge to support both training and reasoning trajectory building. A detailed flow is shown in Figure~\ref{fig_3_train_corpus.png}. Excerpts from the Knowledge Base are provided in Appendix~\ref{knowledge_base}.


\begin{figure*}[t!]
  \includegraphics[width=\textwidth]{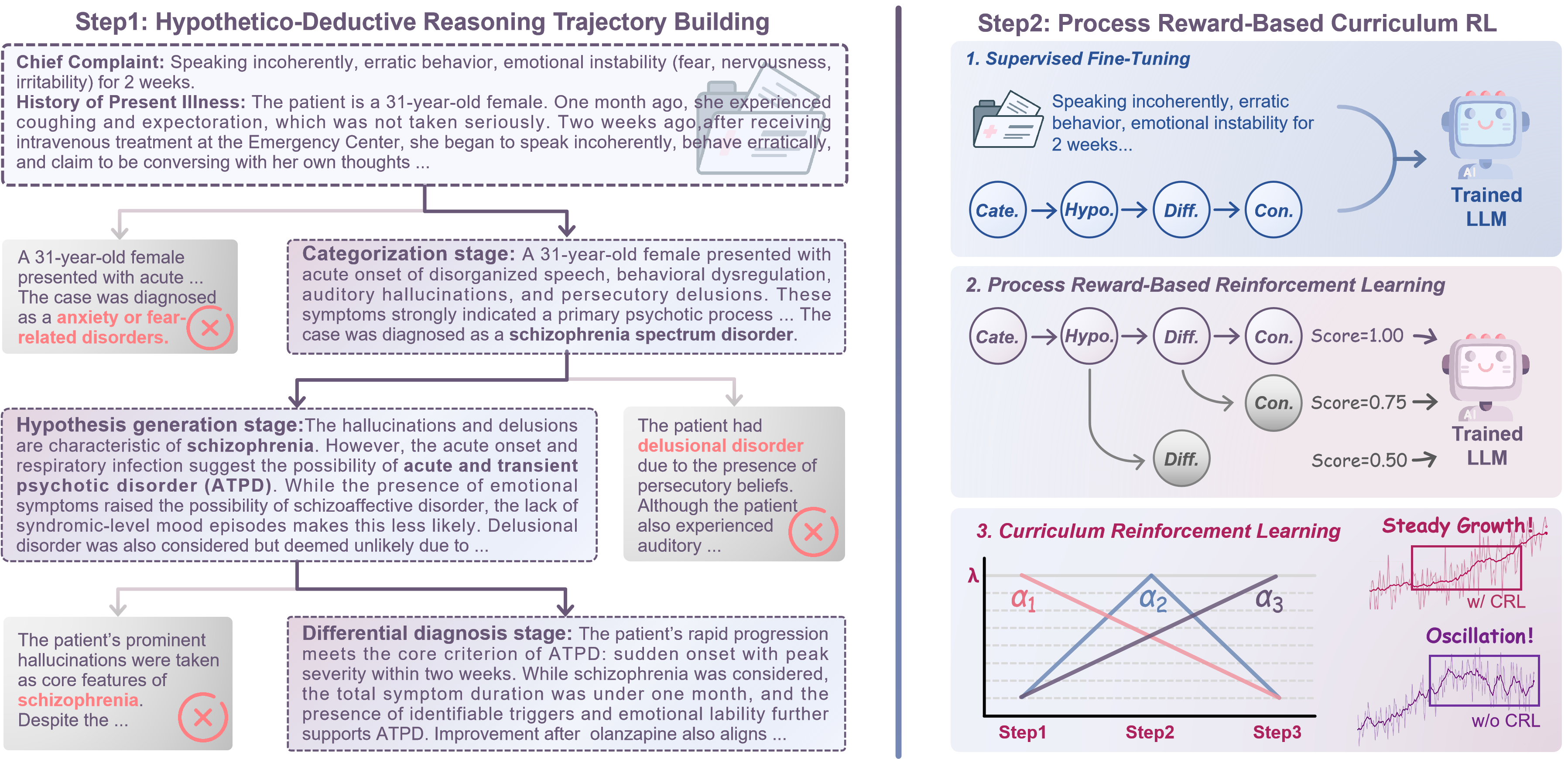}
\caption{Training pipeline of \textsc{MentalSeek-Dx} to address paradigm misalignment by modeling clinically grounded diagnostic reasoning, consisting of two stages: 1) Hypothetico-Deductive Reasoning Trajectory Building using $\mathcal{K}$ as SFT corpus; 2) Reward-Based Curriculum Reinforcement Learning to guide the diagnostic process.}
  \label{fig_main.png}
\end{figure*}

\textbf{Clinically Grounded Corpus.}  
We extract 15,000 clinical cases from a medical center, independently sourced and strictly disjoint from the MentalDx Bench. These cases are processed through a three-stage refinement pipeline, executed by {DeepSeek-R1}~\citep{guo2025deepseek},which performs labeling and rewriting under the guidance of $\mathcal{K}$. 

\textbf{Step 1.} Given a case $x$, we first retrieve all category definitions $\{d_i\}_{i=1}^{16}$ from $\mathcal{K}$ and prompt the LLM to select the most compatible category:
\begin{equation}
\small
\begin{aligned}
c = \texttt{LLM}(x, \{d_i\}) \quad \text{where } d_i \in \mathcal{K}_{\text{category}}.
\end{aligned}
\end{equation}

\textbf{Step 2.} Conditioned on $x$ and the predicted category $c$, we then extract all candidate disorders and their diagnostic criteria from $\mathcal{K}(c)$, and prompt the LLM to identify the most plausible disorder:
\begin{equation}
\small
\begin{aligned}
d = \texttt{LLM}(x, \mathcal{K}(c)).
\end{aligned}
\end{equation}

\textbf{Step 3.} To improve alignment with $(c, d)$, we prompt the LLM to rewrite the original case $x$ by referencing the diagnostic definitions and criteria of $c$ and $d$, ensuring clinical consistency as:
\begin{equation}
\small
\begin{aligned}
x' = \texttt{LLM\_rewrite}(x; \mathcal{K}(c, d)).
\end{aligned}
\end{equation}

The entire automatic procedure is executed on-site within the medical center using an on-premise deployment of {DeepSeek-R1}, ensuring that all data remains local and under institutional governance. Prompt details for Step 1, Step 2, and Step 3 are provided in Appendix~\ref{refinement_pipline_details}.


\subsection{Hypothetico-Deductive Reasoning Trajectory Building} 

\paragraph{Hypothetico‑deductive Reasoning.} 

A foundational model of clinical decision-making, it characterizes how clinicians systematically generate, evaluate, and iteratively refine diagnostic hypotheses based on observed evidence \cite{jefford2011decision,langridge2016role,collins2017clinical}. Inspired by this paradigm, we formulate psychiatric diagnosis as a structured three-step reasoning process: \textit{symptom categorization}, \textit{hypothesis generation}, and \textit{differential diagnosis}. Together, these steps trace a closed diagnostic trajectory—identifying dysfunctions, proposing candidate disorders, and progressively refining hypotheses toward a coherent and well-supported final diagnosis.


\vspace{-8pt}
\paragraph{Reasoning Trajectory Building.} Given an input case $x$, we construct a reasoning trajectory $(s, h, d)$ (i.e., symptom, hypothesis, and differential) by decomposing the diagnostic process into the three stages above. At each stage, the model generates the corresponding reasoning component conditioned on the input case, previous outputs, and structured knowledge retrieved from $\mathcal{K}$. 
Specifically, in the categorization stage, the model identifies relevant diagnostic categories $c$ by matching features in $x$ with characteristic symptoms in $\mathcal{K}(c)$, yielding $s$. 
In the hypothesis generation stage, the model produces a candidate set $h = \{ h_1, h_2, \dots \}$ based on $(x, s)$ and the disorder-specific features under category $c$. In the differential diagnosis stage, the model compares each $d \in h$ using their full feature sets from $\mathcal{K}(d)$, and selects the most plausible diagnosis. We retain only trajectories that successfully complete all trajectory as SFT corpus. For detailed procedures, see Appendix~\ref{trajectory_building_details}.

\subsection{Process Reward-Based Curriculum Reinforcement Learning}

\noindent\textbf{Reinforcement Learning.} 
After supervised training on HDR, we further optimize \textsc{MentalSeek-Dx} using \textit{Process Reward-Based Curriculum Reinforcement Learning (PRCRL)}, which builds upon Group Relative Policy Optimization (GRPO) \citep{guo2025deepseek}. For each case $x$, we sample candidate trajectories ${y_1, \dots, y_G}$ from the current policy $\pi_{\theta_{\text{old}}}$ and assign scalar rewards ${R_i}$: 
\begin{equation}
\small
\begin{aligned}
\hat{A}_i = \frac{R_i - \mu_R}{\sigma_R + \epsilon},
\end{aligned}
\end{equation}
where $\mu_R$ and $\sigma_R$ denote the mean and standard deviation of rewards within the group. The policy is then updated using a PPO-style clipped objective: 
\begin{equation}
\small
\begin{aligned}
\mathcal{J}(\theta)
= \mathbb{E}_{x, y_i}
\Big[
\min \big(
r_i \hat{A}_i,\;
\operatorname{clip}(r_i, 1 - \epsilon, 1 + \epsilon)\hat{A}_i
\big)
\Big].
\end{aligned}
\end{equation}
where $r_i$ is the importance sampling ratio.

\textbf{Reward Composition.}
The effectiveness of \textit{PRCRL} hinges on reward design. Instead of a single outcome-level reward, we define stage-aware, process-level rewards that reflect structured clinical reasoning. Each reward $R_i$ is computed as a weighted combination of multiple components:
\begin{equation}
\small
\begin{aligned}
R_{\text{diagnosis}} = \alpha_1 R_{\text{cat.}} + \alpha_2 R_{\text{hypo.}} + \alpha_3 R_{\text{diff.}},
\end{aligned}
\end{equation}
where $R_{\text{cat.}}$, $R_{\text{hypo.}}$, and $R_{\text{diff.}}$ correspond to category identification, hypothesis generation, and differential diagnosis, respectively, and $\alpha_k$ are fixed coefficients controlling their relative contribution.

\textit{1) Symptom categorization reward.}
At the first stage, \textsc{MentalSeek-Dx} is required to identify the correct high-level diagnostic category based on presenting symptoms. We formulate this stage as a coarse-grained classification task and define:
\begin{equation}
\small
\begin{aligned}
R_{\text{cat.}} = \mathbb{I}\left[C_{\text{pred}} = C_{\text{gt}}\right],
\end{aligned}
\end{equation}
where $C_{\text{pred}}$ and $C_{\text{gt}}$ denote the predicted and ground-truth diagnostic categories, and $\mathbb{I}[\cdot]$ is the indicator function. This reward enforces category fidelity and limits diagnostic error propagation.

\textit{2) Hypothesis generation reward.}
The second stage evaluates the quality of hypothesis construction, focusing on whether the correct disorder is included and appropriately prioritized among candidate diagnoses. Let $\mathcal{H}_{\text{pred}}$ denote the predicted hypothesis list and $i$ be the rank position of the ground-truth disorder $d_{\text{gt}}$ if present. We define:
\begin{equation}
\small
\begin{aligned}
R_{\text{hypo.}} = \mathbb{I}[i \leq 4] \cdot \left(1 - \frac{i - 1}{4}\right).
\end{aligned}
\end{equation}

This formulation rewards both coverage and ranking quality, encouraging concise yet focused hypotheses and penalizing overinclusive reasoning.

\textit{3) Differential diagnosis reward.}
At the final stage, the model is expected to converge to a single, clinically correct diagnosis. We define:
\begin{equation}
\small
\begin{aligned}
R_{\text{diff.}} = \mathbb{I}\left[d_{\text{pred}} = d_{\text{gt}}\right],
\end{aligned}
\end{equation}
where $d_{\text{pred}}$ and $d_{\text{gt}}$ denote the predicted and ground-truth diagnoses. This term directly measures diagnostic correctness and anchors the entire reasoning process toward clinically valid outcomes.


\textbf{Curriculum Learning.}
To guide the model through the progressive structure of clinical reasoning, inspired by curriculum learning~\citep{wang2021survey,liu2021competence}, we adopt a stage-aware reward scheduling strategy. Specifically, we divide training into five epochs and adjust the stage-wise reward weights $(\alpha_1, \alpha_2, \alpha_3)$ over time. Early epochs prioritize category-level accuracy by setting higher $\alpha_1$, while later epochs gradually increase $\alpha_2$ or $\alpha_3$ to emphasize hypothesis generation or differential diagnosis. This training schedule helps the model internalize diagnostic reasoning in a structured manner, supporting stable optimization and improved fine-grained diagnostic performance. For a detailed overview of the training flow, see Fig~\ref{fig_main.png}.


\begin{table}[t!]
\small
\centering
\definecolor{TopTwo}{HTML}{d7d4e9}
\resizebox{0.8 \columnwidth}{!}{%
\begin{tabular}{lccc}
\Xhline{1pt}
\textbf{Model} & \textbf{\(\mathcal{CA}\)} & \textbf{\(\mathcal{DA}\)} & \textbf{\(\mathcal{JA}\)} \\
\hline
Qwen3-32B         & 58.71 & 31.04 & 26.26 \\
QWQ-32B           & 76.26 & 47.89 & 46.63 \\
Baichuan-M2       & 64.61 & 44.38 & 41.71 \\
GPT-OSS-120B      & 75.84 & 52.53 & 50.00 \\
DeepSeek-R1       & 78.37 & 55.20 & 54.49 \\
Claude-Sonnet-4.5 & 80.76 & 59.69 & 58.15 \\
\hline\hline
\rowcolor{TopTwo} \textsc{MentalSeek-Dx-7B}  & \textbf{82.58} & \textbf{67.56} & \textbf{67.13} \\
\rowcolor{TopTwo} \textsc{MentalSeek-Dx-14B} & \textbf{83.99} & \textbf{70.08} & \textbf{69.38} \\
\Xhline{1pt}
\end{tabular}}
\caption{Comparison of \textsc{MentalSeek-Dx} with other models, achieving state-of-the-art on all metrics.}
\label{tab:llm_performance}
\end{table}

\section{\textsc{MentalSeek-Dx} Evaluation}
We train \textsc{MentalSeek-Dx} based on Qwen2.5-7B and 14B~\citep{team2024qwen2}; training details and hyperparameter settings can see in Appendix~\ref{sec:hyperparameters}.


\begin{table}[t!]

\definecolor{SizeSevenB}{HTML}{f0eff8}      
\definecolor{SizeFourteenB}{HTML}{e0e9f7}   

\centering
\begin{adjustbox}{max width=0.95\columnwidth}
\small
\resizebox{0.78\textwidth}{!}{
\begin{tabular}{lccc}
\toprule
& $\mathcal{CA}$
& $\mathcal{DA}$
& $\mathcal{JA}$ \\

\midrule

\multicolumn{4}{l}{\itshape Baseline LLMs} \\
\rowcolor{SizeSevenB}
{\scriptsize Qwen2.5-7B-Instruct}
& 47.47 & 14.75 & 10.96 \\
\rowcolor{SizeFourteenB}
{\scriptsize Qwen2.5-14B-Instruct}
& 61.66 & 29.92 & 24.72 \\

\midrule \midrule

\multicolumn{4}{l}{\itshape Hypothetico-Heductive Reasoning (HDR)} \\
\rowcolor{SizeSevenB}
{\scriptsize {\textcolor{deepblue}{7B-SFT}} w/o \tcancel{HDR}}
& 68.14 {\color{alizarin}(+20.67)} & 42.56 {\color{alizarin}(+27.81)} & 40.14 {\color{alizarin}(+29.18)} \\
\rowcolor{SizeSevenB}
{\scriptsize {\textcolor{deepblue}{7B-SFT}} w/ HDR}
& 82.02 {\color{alizarin}(+34.55)} & 62.78 {\color{alizarin}(+48.03)} & 62.50 {\color{alizarin}(+51.54)} \\
\rowcolor{SizeFourteenB}
{\scriptsize {\textcolor{deepblue}{14B-SFT}} w/o \tcancel{HDR}}
& 69.21 {\color{alizarin}(+7.55)} & 47.44 {\color{alizarin}(+17.52)} & 46.88 {\color{alizarin}(+22.16)} \\
\rowcolor{SizeFourteenB}
{\scriptsize {\textcolor{deepblue}{14B-SFT}} w/ HDR}
& 83.71 {\color{alizarin}(+22.05)} & 64.47 {\color{alizarin}(+34.55)} & 64.47 {\color{alizarin}(+39.75)} \\

\midrule \midrule

\multicolumn{4}{l}{\itshape Reinforcement Learning (RL)} \\
\rowcolor{SizeSevenB}
{\scriptsize {\textcolor{deepblue}{7B-HDR} w/o \tcancel{RL}} }
& 82.02 {\color{alizarin}(+34.55)} & 62.78 {\color{alizarin}(+48.03)} & 62.50 {\color{alizarin}(+51.54)} \\

\rowcolor{SizeSevenB}
{\scriptsize {\textcolor{deepblue}{7B-HDR}}  w/ RL}
& 82.16 {\color{alizarin}(+34.69)} & 66.01 {\color{alizarin}(+51.26)} & 65.87 {\color{alizarin}(+54.91)} \\

\rowcolor{SizeSevenB}
{\scriptsize {\textcolor{deepblue}{7B-HDR}}  w/ CRL}
& 82.58 {\color{alizarin}(+35.11)} & 67.56 {\color{alizarin}(+52.81)} & 67.13 {\color{alizarin}(+56.17)} \\

\rowcolor{SizeFourteenB}
{\scriptsize {\textcolor{deepblue}{14B-HDR}} w/o \tcancel{RL}}
& 81.59 {\color{alizarin}(+19.93)} & 64.47 {\color{alizarin}(+34.55)} & 64.47 {\color{alizarin}(+39.75)} \\

\rowcolor{SizeFourteenB}
{\scriptsize {\textcolor{deepblue}{14B-HDR}}  w/ RL}
& 83.71 {\color{alizarin}(+22.05)} & 67.28 {\color{alizarin}(+37.36)} & 67.28 {\color{alizarin}(+42.56)} \\

\rowcolor{SizeFourteenB}
{\scriptsize {\textcolor{deepblue}{14B-HDR}} w/ CRL}
& 83.99 {\color{alizarin}(+22.33)} & 70.08 {\color{alizarin}(+40.16)} & 69.38 {\color{alizarin}(+44.66)} \\





\bottomrule
\end{tabular}
}
\end{adjustbox}
\caption{The ablation results on MentalDx Bench.}
\label{tab:ablation_study}
\end{table}

\begin{figure}[t!]
  \centering
  \includegraphics[width=0.9\linewidth]{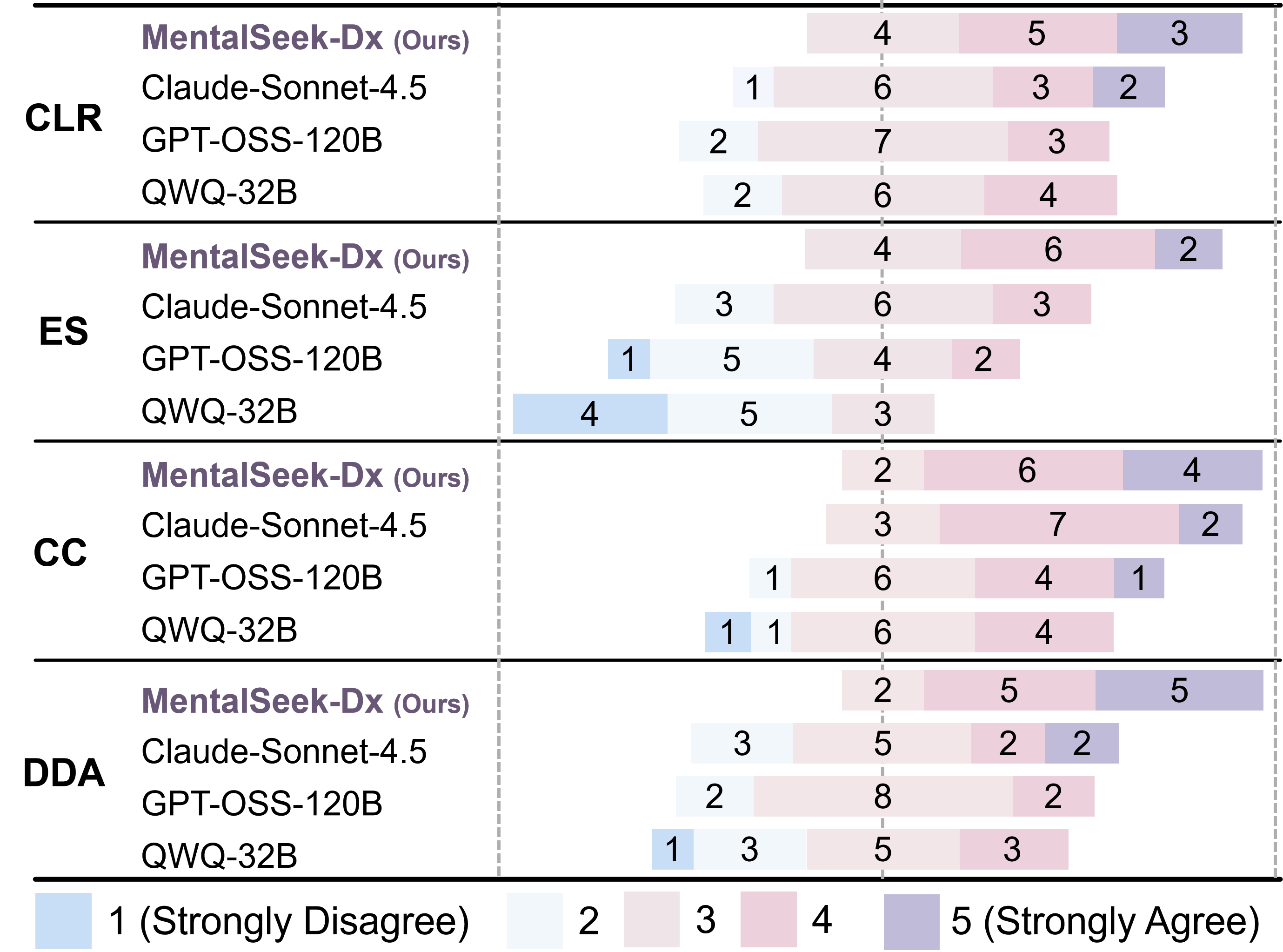}
\caption{Expert evaluation of diagnostic reasoning, showing superior performance of \textsc{MentalSeek-Dx}.}
  \label{reasoning_comparation.png}
\end{figure}

\textbf{Evaluation Results.} We compare MentalSeek-Dx with other LLMs on MentalDx Bench. As shown in Table~\ref{tab:llm_performance}, our model achieves {state-of-the-art} performance on all metrics among mainstream models, despite having only 7B and 14B parameters. Detailed results are provided in Appendix~\ref{bench_result}. Complementary to these quantitative results, we conduct a blinded evaluation (Figure~\ref{reasoning_comparation.png}) with 12 medical experts using a Likert scale to assess the reasoning process along four dimensions: Clinical Logical Rigor, Evidence Strength, Critical Completeness, and Differential Diagnosis Adequacy. Results indicate that MentalSeek-Dx, guided by progressive hypothetico-deductive reasoning, not only improves diagnostic accuracy but also demonstrates structured clinical reasoning toward transparent, accurate, and clinically trustworthy diagnosis. Annotation guidelines can see in Appendix~\ref{sec:cot}.

\begin{figure}[t!]
\centering
\includegraphics[width=\linewidth]{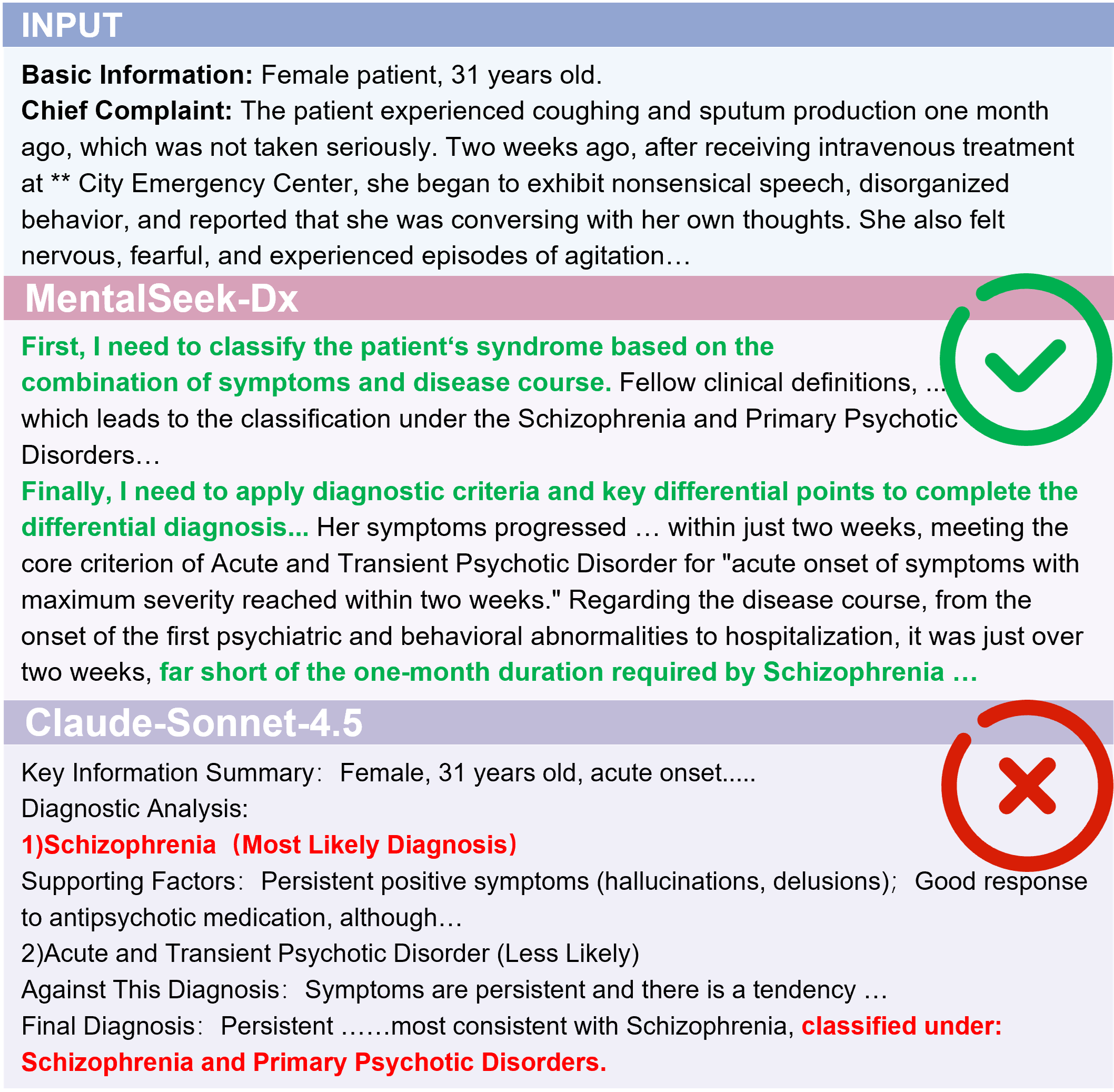}
\caption{Comparison between Claude-Sonnet-4.5 and \textsc{MentalSeek-Dx}, showing a paradigm misalignment.}

\label{case_study}
\end{figure}

\textbf{Ablation Study.} We evaluate the contribution of each training stage in \textsc{MentalSeek-Dx} on Qwen2.5-7B and 14B (Table~\ref{tab:ablation_study}). All components consistently improve diagnostic performance. Incorporating hypothetico-deductive reasoning yields substantial gains over direct training, increasing \textit{CA}, \textit{DA}, and \textit{JA} by 13.88\%, 20.22\%, and 22.36\% on Qwen2.5-14B, respectively, underscoring the importance of clinical reasoning in resolving the paradigm misalignment between pattern-based associations and diagnostic reasoning. Beyond SFT, reinforcement learning further improves diagnostic precision. On Qwen2.5-7B, RL adds 3.23\% in \textit{DA} and 3.37\% in \textit{JA}, indicating that RL helps refine model decisions in challenging cases. Notably, while CRL brings modest gains, it effectively stabilizes the RL training process (see Appendix~\ref{CRL}).


\textbf{Case Study.} In Fig.~\ref{case_study}, we present a case comparison. Claude-Sonnet-4.5 shows paradigm misalignment by mapping symptoms directly to a disorder and then re-inferring its category. This reverse process reflects unstructured knowledge and leads to diagnostic error. In contrast, MentalSeek-Dx follows a clinically grounded path: it first constrains category-level reasoning, then formulates disorder hypotheses, and finally applies diagnostic criteria. More details are provided in Appendix~\ref{case_study_2}.

\section{Conclusion}
This work bridges the critical gap between computational psychiatry and clinical reality. By introducing \textbf{MentalDx Bench}, we established the first ecologically valid framework for evaluating fine-grained disorder diagnosis under strict ICD-11 standards. Our investigation uncovers a fundamental \textit{paradigm misalignment}: while general-purpose LLMs rely on associative shortcuts, they fail to replicate the rigorous \textit{hypothetico-deductive reasoning} essential for clinical practice. We address this by proposing \textbf{MentalSeek-Dx}, a specialized framework that aligns model cognition with psychiatrist-level logic through trajectory-based supervision and curriculum reinforcement learning. Achieving state-of-the-art results with only 14B parameters, our work demonstrates that \textit{cognitive alignment}, rather than mere parameter scale, is the decisive factor for reliable automated diagnosis.

\section*{Limitations}
Despite the promising results, this study has limitations inherent to the complexity of computational psychiatry.
First, while \textbf{MentalDx Bench} offers high ecological validity, the dataset primarily comprises textual Electronic Health Records. In actual clinical practice, diagnosis is multimodal, incorporating patient demeanor, speech prosody, and non-verbal cues. Our current focus on textual modalities serves as a controlled foundational step, but it may not fully capture the complete spectrum of diagnostic signals available in face-to-face consultations.
Second, although \textbf{MentalSeek-Dx} demonstrates advanced reasoning capabilities, it is not immune to the broader challenges of interpretability and safety in Large Language Models. In high-stakes medical settings, our model is designed strictly as a clinical decision support tool. It cannot, and should not, replace the final judgment of qualified human psychiatrists. The human-in-the-loop paradigm remains essential to mitigate potential risks associated with algorithmic hallucinations or biases in long-tail rare disorders.

\section*{Ethics Statement}
\label{ethics_statement}
This study was conducted in strict adherence to ethical guidelines, ensuring the protection of participants’ rights and the integrity of the research process. The study protocol, including all data collection, labeling, and annotation procedures, was thoroughly reviewed and approved by the institutional ethics boards of all participating clinical centers. This approval process ensured that all aspects of the study met the highest standards for legal and ethical conduct in clinical research.

Data collection and annotation activities followed the ethical principles outlined in the \textit{Declaration of Helsinki}, which emphasizes respect for individuals, the protection of privacy, and the commitment to research integrity. In particular, all processes were designed to minimize any potential risks to participants and to uphold their dignity throughout the study.

In order to protect participant confidentiality and privacy, all records used in the study were fully de-identified prior to any processing or analysis. This de-identification process ensured that no personally identifiable information was retained, further safeguarding participants’ anonymity. As a result of this de-identification, the requirement for individual informed consent was waived by the institutional ethics boards. This decision was based on the fact that the study utilized fully anonymized data, with no identifiable link to individual participants, and posed no risk to personal privacy.

Additionally, the ethical standards for handling and analyzing sensitive data were rigorously followed throughout the study. The research was conducted solely for scientific purposes, with all data being used in a manner that was consistent with the ethical guidelines governing clinical research.

For transparency, we note that the dataset used in this study was reviewed and approved by the \textbf{relevant institutional ethics review committees under formal protocols}. Due to anonymization requirements, detailed ethics approval identifiers and data governance specifics are not disclosed at this stage. These details will be updated in the final published version of the paper where permitted.

\bibliography{arxiv}

@inproceedings{zheng2025rsd,
  title={RSD-15K: A Large-Scale User-Level Annotated Dataset for Suicide Risk Detection on Social Media},
  author={Zheng, Shouwen and Tao, Yingzhi and Zhou, Taiqi},
  booktitle={2025 IEEE 41st International Conference on Data Engineering Workshops (ICDEW)},
  pages={190--196},
  year={2025},
  organization={IEEE}
}

@article{johri2025evaluation,
  title={An evaluation framework for clinical use of large language models in patient interaction tasks},
  author={Johri, Shreya and Jeong, Jaehwan and Tran, Benjamin A and Schlessinger, Daniel I and Wongvibulsin, Shannon and Barnes, Leandra A and Zhou, Hong-Yu and Cai, Zhuo Ran and Van Allen, Eliezer M and Kim, David and others},
  journal={Nature medicine},
  volume={31},
  number={1},
  pages={77--86},
  year={2025},
  publisher={Nature Publishing Group US New York}
}

@article{sao2024miroberta,
  title={MIRoBERTa: Mental Illness Text Classification with Transfer Learning on Subreddits},
  author={Sao, Mavin and Lim, Hoi-Jeong},
  journal={IEEE Access},
  year={2024},
  publisher={IEEE}
}

@article{frydman2025large,
  title={Large Language Models for Psychiatric Phenotype Extraction from Electronic Health Records},
  author={Frydman-Gani, Clara and Arias, Alejandro and Vallejo, Maria Perez and Mart{\'\i}nez, John Daniel Londo{\~n}o and Valencia-Echeverry, Johanna and Casta{\~n}o, Mauricio and Bui, Alex AT and Freimer, Nelson B and Lopez-Jaramillo, Carlos and Loohuis, Loes M Olde},
  journal={medRxiv},
  year={2025}
}

@inproceedings{zhu2025medtpe,
  title={Medtpe: Compressing long ehr sequence for llm-based clinical prediction with token-pair encoding},
  author={Zhu, Mingcheng and Luo, Zhiyao and Liu, Yu and Zhu, Tingting},
  booktitle={Proceedings of the 11th Mining and Learning from Time Series Workshop@ KDD},
  volume={2025},
  year={2025}
}

@article{yao2022d4,
  title={D4: a chinese dialogue dataset for depression-diagnosis-oriented chat},
  author={Yao, Binwei and Shi, Chao and Zou, Likai and Dai, Lingfeng and Wu, Mengyue and Chen, Lu and Wang, Zhen and Yu, Kai},
  journal={arXiv preprint arXiv:2205.11764},
  year={2022}
}

@article{shin2024using,
  title={Using large language models to detect depression from user-generated diary text data as a novel approach in digital mental health screening: instrument validation study},
  author={Shin, Daun and Kim, Hyoseung and Lee, Seunghwan and Cho, Younhee and Jung, Whanbo},
  journal={Journal of Medical Internet Research},
  volume={26},
  pages={e54617},
  year={2024},
  publisher={JMIR Publications Toronto, Canada}
}

@article{obradovich2024opportunities,
  title={Opportunities and risks of large language models in psychiatry},
  author={Obradovich, Nick and Khalsa, Sahib S and Khan, Waqas U and Suh, Jina and Perlis, Roy H and Ajilore, Olusola and Paulus, Martin P},
  journal={NPP—Digital Psychiatry and Neuroscience},
  volume={2},
  number={1},
  pages={8},
  year={2024},
  publisher={Springer International Publishing Cham}
}

@article{radwan2024predictive,
  title={Predictive analytics in mental health leveraging LLM embeddings and machine learning models for social media analysis},
  author={Radwan, Ahmad and Amarneh, Mohannad and Alawneh, Hussam and Ashqar, Huthaifa I and AlSobeh, Anas and Magableh, Aws Abed Al Raheem},
  journal={International Journal of Web Services Research (IJWSR)},
  volume={21},
  number={1},
  pages={1--22},
  year={2024},
  publisher={IGI Global Scientific Publishing}
}

@article{xu2024mental,
  title={Mental-llm: Leveraging large language models for mental health prediction via online text data},
  author={Xu, Xuhai and Yao, Bingsheng and Dong, Yuanzhe and Gabriel, Saadia and Yu, Hong and Hendler, James and Ghassemi, Marzyeh and Dey, Anind K and Wang, Dakuo},
  journal={Proceedings of the ACM on Interactive, Mobile, Wearable and Ubiquitous Technologies},
  volume={8},
  number={1},
  pages={1--32},
  year={2024},
  publisher={ACM New York, NY, USA}
}

@article{dai2025psyche,
  title={Psyche-r1: Towards reliable psychological llms through unified empathy, expertise, and reasoning},
  author={Dai, Chongyuan and Hu, Jinpeng and Shi, Hongchang and Li, Zhuo and Yang, Xun and Wang, Meng},
  journal={arXiv preprint arXiv:2508.10848},
  year={2025}
}

@article{hu2024psycollm,
  title={Psycollm: Enhancing llm for psychological understanding and evaluation},
  author={Hu, Jinpeng and Dong, Tengteng and Luo, Gang and Ma, Hui and Zou, Peng and Sun, Xiao and Guo, Dan and Yang, Xun and Wang, Meng},
  journal={IEEE Transactions on Computational Social Systems},
  volume={12},
  number={2},
  pages={539--551},
  year={2024},
  publisher={IEEE}
}

@inproceedings{bao2025redsm5,
  title={ReDSM5: A Reddit Dataset for DSM-5 Depression Detection},
  author={Bao, Eliseo and P{\'e}rez, Anxo and Parapar, Javier},
  booktitle={Proceedings of the 34th ACM International Conference on Information and Knowledge Management},
  pages={6323--6327},
  year={2025}
}

@article{ignashina2025llm,
  title={Llm assistance for pediatric depression},
  author={Ignashina, Mariia and Bondaronek, Paulina and Santel, Dan and Pestian, John and Ive, Julia},
  journal={arXiv preprint arXiv:2501.17510},
  year={2025}
}

@article{mankarious2025mindset,
  title={MindSET: Advancing Mental Health Benchmarking through Large-Scale Social Media Data},
  author={Mankarious, Saad and Zirikly, Ayah and Wiechmann, Daniel and Kerz, Elma and Kempa, Edward and Qiao, Yu},
  journal={arXiv preprint arXiv:2511.20672},
  year={2025}
}

@article{kim2025baseline,
  title={A baseline for self-state identification and classification in mental health data: CLPsych 2025 Task},
  author={Kim, Laerdon},
  journal={arXiv preprint arXiv:2504.14066},
  year={2025}
}

@inproceedings{yang2024mentallama,
  title={MentaLLaMA: interpretable mental health analysis on social media with large language models},
  author={Yang, Kailai and Zhang, Tianlin and Kuang, Ziyan and Xie, Qianqian and Huang, Jimin and Ananiadou, Sophia},
  booktitle={Proceedings of the ACM Web Conference 2024},
  pages={4489--4500},
  year={2024}
}

@article{cohan2018smhd,
  title={SMHD: a large-scale resource for exploring online language usage for multiple mental health conditions},
  author={Cohan, Arman and Desmet, Bart and Yates, Andrew and Soldaini, Luca and MacAvaney, Sean and Goharian, Nazli},
  journal={arXiv preprint arXiv:1806.05258},
  year={2018}
}

@article{fu2025first,
  title={The First MPDD Challenge: Multimodal Personality-aware Depression Detection},
  author={Fu, Changzeng and Fu, Zelin and Zhang, Qi and Kuang, Xinhe and Dong, Jiacheng and Su, Kaifeng and Su, Yikai and Shi, Wenbo and Yao, Junfeng and Zhao, Yuliang and others},
  journal={arXiv preprint arXiv:2505.10034},
  year={2025}
}

@article{fouda2025psychiatrybench,
  title={PsychiatryBench: A Multi-Task Benchmark for LLMs in Psychiatry},
  author={Fouda, Aya E and Hassan, Abdelrahamn A and Hanafy, Radwa J and Fouda, Mohammed E},
  journal={arXiv preprint arXiv:2509.09711},
  year={2025}
}

@article{team2024qwen2,
  title={Qwen2 technical report},
  author={Team, Qwen and others},
  journal={arXiv preprint arXiv:2407.10671},
  volume={2},
  number={3},
  year={2024}
}

@article{yang2025qwen3,
  title={Qwen3 technical report},
  author={Yang, An and Li, Anfeng and Yang, Baosong and Zhang, Beichen and Hui, Binyuan and Zheng, Bo and Yu, Bowen and Gao, Chang and Huang, Chengen and Lv, Chenxu and others},
  journal={arXiv preprint arXiv:2505.09388},
  year={2025}
}

@article{patterson2022carbon,
  title={The carbon footprint of machine learning training will plateau, then shrink},
  author={Patterson, David and Gonzalez, Joseph and H{\"o}lzle, Urs and Le, Quoc and Liang, Chen and Munguia, Lluis-Miquel and Rothchild, Daniel and So, David R and Texier, Maud and Dean, Jeff},
  journal={Computer},
  volume={55},
  number={7},
  pages={18--28},
  year={2022},
  publisher={IEEE}
}

@article{team2025kimi,
  title={Kimi k2: Open agentic intelligence},
  author={Team, Kimi and Bai, Yifan and Bao, Yiping and Chen, Guanduo and Chen, Jiahao and Chen, Ningxin and Chen, Ruijue and Chen, Yanru and Chen, Yuankun and Chen, Yutian and others},
  journal={arXiv preprint arXiv:2507.20534},
  year={2025}
}

@article{lan2025clinicalgpt,
  title={Clinicalgpt-r1: Pushing reasoning capability of generalist disease diagnosis with large language model},
  author={Lan, Wuyang and Wang, Wenzheng and Ji, Changwei and Yang, Guoxing and Zhang, Yongbo and Liu, Xiaohong and Wu, Song and Wang, Guangyu},
  journal={arXiv preprint arXiv:2504.09421},
  year={2025}
}

@misc{chen2024huatuogpt,
      title={HuatuoGPT-o1, Towards Medical Complex Reasoning with LLMs}, 
      author={Junying Chen and Zhenyang Cai and Ke Ji and Xidong Wang and Wanlong Liu and Rongsheng Wang and Jianye Hou and Benyou Wang},
      year={2024},
      eprint={2412.18925},
      archivePrefix={arXiv}, 
}

@article{dou2025baichuan,
  title={Baichuan-m2: Scaling medical capability with large verifier system},
  author={Dou, Chengfeng and Liu, Chong and Yang, Fan and Li, Fei and Jia, Jiyuan and Chen, Mingyang and Ju, Qiang and Wang, Shuai and Dang, Shunya and Li, Tianpeng and others},
  journal={arXiv preprint arXiv:2509.02208},
  year={2025}
}

@article{sellergren2025medgemma,
  title={Medgemma technical report},
  author={Sellergren, Andrew and Kazemzadeh, Sahar and Jaroensri, Tiam and Kiraly, Atilla and Traverse, Madeleine and Kohlberger, Timo and Xu, Shawn and Jamil, Fayaz and Hughes, C{\'\i}an and Lau, Charles and others},
  journal={arXiv preprint arXiv:2507.05201},
  year={2025}
}

@article{agarwal2025gpt,
  title={gpt-oss-120b \& gpt-oss-20b model card},
  author={Agarwal, Sandhini and Ahmad, Lama and Ai, Jason and Altman, Sam and Applebaum, Andy and Arbus, Edwin and Arora, Rahul K and Bai, Yu and Baker, Bowen and Bao, Haiming and others},
  journal={arXiv preprint arXiv:2508.10925},
  year={2025}
}

@article{mangalik2024robust,
  title={Robust language-based mental health assessments in time and space through social media},
  author={Mangalik, Siddharth and Eichstaedt, Johannes C and Giorgi, Salvatore and Mun, Jihu and Ahmed, Farhan and Gill, Gilvir and V. Ganesan, Adithya and Subrahmanya, Shashanka and Soni, Nikita and Clouston, Sean AP and others},
  journal={NPJ Digital Medicine},
  volume={7},
  number={1},
  pages={109},
  year={2024},
  publisher={Nature Publishing Group UK London}
}

@article{zhuang2024postgraduate,
  title={Postgraduate psychological stress detection from social media using BERT-Fused model},
  author={Zhuang, Muni and Cheng, Dongsheng and Lu, Xin and Tan, Xu},
  journal={PloS one},
  volume={19},
  number={10},
  pages={e0312264},
  year={2024},
  publisher={Public Library of Science San Francisco, CA USA}
}

@article{wang2025large,
  title={Large language models in clinical psychiatry: Applications and optimization strategies},
  author={Wang, Yi-Fan and Li, Ming-Da and Wang, Su-Hong and Fang, Yin and Sun, Jie and Lu, Lin and Yan, Wei},
  journal={World Journal of Psychiatry},
  volume={15},
  number={11},
  pages={108199},
  year={2025}
}

@article{levkovich2024large,
  title={Large language models outperform general practitioners in identifying complex cases of childhood anxiety},
  author={Levkovich, Inbar and Rabin, Eyal and Brann, Michal and Elyoseph, Zohar},
  journal={Digital Health},
  volume={10},
  pages={20552076241294182},
  year={2024},
  publisher={SAGE Publications Sage UK: London, England}
}

@article{guo2025deepseek,
  title={Deepseek-r1: Incentivizing reasoning capability in llms via reinforcement learning},
  author={Guo, Daya and Yang, Dejian and Zhang, Haowei and Song, Junxiao and Zhang, Ruoyu and Xu, Runxin and Zhu, Qihao and Ma, Shirong and Wang, Peiyi and Bi, Xiao and others},
  journal={arXiv preprint arXiv:2501.12948},
  year={2025}
}

@book{collins2017clinical,
  title={Clinical reasoning in image guided radiotherapy: a multimethod study},
  author={Collins, Mark},
  year={2017},
  publisher={Sheffield Hallam University (United Kingdom)}
}

@article{langridge2016role,
  title={The role of clinician emotion in clinical reasoning: Balancing the analytical process},
  author={Langridge, Neil and Roberts, Lisa and Pope, Catherine},
  journal={Manual therapy},
  volume={21},
  pages={277--281},
  year={2016},
  publisher={Elsevier}
}

@article{jefford2011decision,
  title={Decision-making theories and their usefulness to the midwifery profession both in terms of midwifery practice and the education of midwives},
  author={Jefford, Elaine and Fahy, Kathleen and Sundin, Deborah},
  journal={International Journal of Nursing Practice},
  volume={17},
  number={3},
  pages={246--253},
  year={2011},
  publisher={Wiley Online Library}
}

@inproceedings{liu2021competence,
  title={Competence-based multimodal curriculum learning for medical report generation},
  author={Liu, Fenglin and Ge, Shen and Wu, Xian},
  booktitle={Proceedings of the 59th annual meeting of the association for computational linguistics and the 11th international joint conference on natural language processing (volume 1: long papers)},
  pages={3001--3012},
  year={2021}
}

@article{wang2021survey,
  title={A survey on curriculum learning},
  author={Wang, Xin and Chen, Yudong and Zhu, Wenwu},
  journal={IEEE transactions on pattern analysis and machine intelligence},
  volume={44},
  number={9},
  pages={4555--4576},
  year={2021},
  publisher={IEEE}
}

@article{bubeck2025early,
  title={Early science acceleration experiments with GPT-5},
  author={Bubeck, S{\'e}bastien and Coester, Christian and Eldan, Ronen and Gowers, Timothy and Lee, Yin Tat and Lupsasca, Alexandru and Sawhney, Mehtaab and Scherrer, Robert and Sellke, Mark and Spears, Brian K and others},
  journal={arXiv preprint arXiv:2511.16072},
  year={2025}
}

@article{appel2025anthropic,
  title={Anthropic economic index report: Uneven geographic and enterprise ai adoption},
  author={Appel, Ruth and McCrory, Peter and Tamkin, Alex and McCain, Miles and Neylon, Tyler and Stern, Michael},
  journal={arXiv preprint arXiv:2511.15080},
  year={2025}
}

@article{byun2025cradle,
  title={CRADLE Bench: A Clinician-Annotated Benchmark for Multi-Faceted Mental Health Crisis and Safety Risk Detection},
  author={Byun, Grace and Lipschutz, Rebecca and Minton, Sean T and Lott, Abigail and Choi, Jinho D},
  journal={arXiv preprint arXiv:2510.23845},
  year={2025}
}

@inproceedings{ge2025survey,
  title={A survey of large language models in mental health disorder detection on social media},
  author={Ge, Zhuohan and Hu, Nicole and Li, Darian and Wang, Yubo and Qi, Shihao and Xu, Yuming and Shi, Han and Zhang, Jason},
  booktitle={2025 IEEE 41st International Conference on Data Engineering Workshops (ICDEW)},
  pages={164--176},
  year={2025},
  organization={IEEE}
}

@inproceedings{chim2024overview,
  title={Overview of the clpsych 2024 shared task: Leveraging large language models to identify evidence of suicidality risk in online posts},
  author={Chim, Jenny and Tsakalidis, Adam and Gkoumas, Dimitris and Atzil-Slonim, Dana and Ophir, Yaakov and Zirikly, Ayah and Resnik, Philip and Liakata, Maria},
  booktitle={Proceedings of the 9th Workshop on Computational Linguistics and Clinical Psychology (CLPsych 2024)},
  pages={177--190},
  year={2024}
}

@article{orru2025large,
  title={Large language models and psychiatry},
  author={Orr{\`u}, Graziella and Melis, Giulia and Sartori, Giuseppe},
  journal={International Journal of Law and Psychiatry},
  volume={101},
  pages={102086},
  year={2025},
  publisher={Elsevier}
}

@article{shewcraft2025algorithmic,
  title={Algorithmic Classification of Psychiatric Disorder--Related Spontaneous Communication Using Large Language Model Embeddings: Algorithm Development and Validation},
  author={Shewcraft, Ryan Allen and Schwarz, John and Micsinai Balan, Mariann},
  journal={JMIR AI},
  volume={4},
  pages={e67369},
  year={2025},
  publisher={JMIR Publications Toronto, Canada}
}

@inproceedings{lan2025depression,
  title={Depression detection on social media with large language models},
  author={Lan, Xiaochong and Han, Zhiguang and Cheng, Yiming and Sheng, Li and Feng, Jie and Gao, Chen and Li, Yong},
  booktitle={Proceedings of the 2025 Conference on Empirical Methods in Natural Language Processing: Industry Track},
  pages={2155--2171},
  year={2025}
}

@article{weber2025using,
  title={Using a fine-tuned large language model for symptom-based depression evaluation},
  author={Weber, Samantha and Deperrois, Nicolas and Heun, Robert and Fr{\"u}hsch{\"u}tz, Laura and Monn, Anna and Homan, Stephanie and H{\"a}fliger, Andrea and Seifritz, Erich and Kowatsch, Tobias and MULTICAST consortium J{\"a}ger Lena 8 Schultebraucks Katharina 9 Gershov Sapir 9 Mocellin Jacopo 1 4 and others},
  journal={npj Digital Medicine},
  volume={8},
  number={1},
  pages={598},
  year={2025},
  publisher={Nature Publishing Group UK London}
}

@article{liu2025psychbench,
  title={PsychBench: A comprehensive and professional benchmark for evaluating the performance of LLM-assisted psychiatric clinical practice},
  author={Liu, Shuyu and Wang, Ruoxi and Zhang, Ling and Zhu, Xuequan and Yang, Rui and Zhou, Xinzhu and Wu, Fei and Yang, Zhi and Jin, Cheng and Wang, Gang},
  journal={arXiv preprint arXiv:2503.01903},
  year={2025}
}

@inproceedings{wu2025psychological,
  title={Psychological health knowledge-enhanced LLM-based social network crisis intervention text transfer recognition method},
  author={Wu, Shurui and Huang, Xinyi and Lu, Dingxin},
  booktitle={Proceedings of the 2025 International Conference on Health Big Data},
  pages={156--161},
  year={2025}
}

@inproceedings{hengle2024still,
  title={Still not quite there! evaluating large language models for comorbid mental health diagnosis},
  author={Hengle, Amey and Kulkarni, Atharva and Patankar, Shantanu Deepak and Chandrasekaran, Madhumitha and D’silva, Sneha and Jacob, Jemima S and Gupta, Rashmi},
  booktitle={Proceedings of the 2024 Conference on Empirical Methods in Natural Language Processing},
  pages={16698--16721},
  year={2024}
}

@article{oh2026clinically,
  title={Clinically validated depression dataset aligned with DSM-5 criteria for major depressive disorder (MDD)},
  author={Oh, Jaedong and Lim, Jooyoung and Oh, Hayoung},
  journal={Expert Systems with Applications},
  volume={296},
  year={2026},
  publisher={Elsevier Ltd}
}

@article{lho2025large,
  title={Large Language Models and Text Embeddings for Detecting Depression and Suicide in Patient Narratives},
  author={Lho, Silvia Kyungjin and Park, Sang-Cheol and Lee, Hahyun and Oh, Da Young and Kim, Hyeonjin and Jang, Soomin and Jung, Hee Yeon and Yoo, So Young and Park, Su Mi and Lee, Jun-Young},
  journal={JAMA Network Open},
  volume={8},
  number={5},
  pages={e2511922--e2511922},
  year={2025},
  publisher={American Medical Association}
}

\section{Appendix}
\label{sec:appendix}


\subsection{Category and Disorder List}
\label{catogory_and_disorder_list}
In this section, we provide the full forms and brief explanations of the category abbreviations used in the main text. Fig~\ref{Disorders_List} presents the names and corresponding codes for the 76 specific disorders included in the study.

The full form of \textbf{ANX} is Anxiety or fear-related disorders. Anxiety or fear-related disorders are a group of mental health conditions characterized by excessive fear and anxiety, along with associated behavioral disturbances. These symptoms are severe enough to cause significant distress or impairment in an individual's personal, social, educational, occupational, or other important areas of functioning. According to the ICD-11, specific types of disorders under this category include generalized anxiety disorder, panic disorder, agoraphobia, specific phobias, social anxiety disorder, and separation anxiety disorder. Patients may exhibit specific cognitive features, such as excessive worry about particular situations or objects, which help distinguish between different types of anxiety or fear-related disorders. 

The full form of \textbf{CATA} is Catatonia. Catatonia is a clinical syndrome characterized by psychomotor disturbances, which manifest as a combination of reduced, increased, or abnormal psychomotor activity. Typical symptoms include stupor (maintaining a fixed posture for extended periods), rigidity (abnormal increase in muscle tension), waxy flexibility (limbs can be manipulated and maintained in a posture), mutism, negativism (active or passive resistance to instructions), echolalia (repetition of others' speech), or echopraxia (repetition of others' actions). This syndrome may arise secondary to psychiatric disorders such as schizophrenia, bipolar disorder, and depressive disorders, and can also be triggered by drug reactions, neurological diseases, or physical illnesses.

The full form of \textbf{SUD} is Disorders due to substance use or addictive behaviors. These disorders refer to functional impairment or distress caused by the repeated use of psychoactive substances (including drugs) or specific repetitive rewarding behaviors, manifesting in cognitive, behavioral, and physiological symptoms. The core features include a persistent craving for the substance or behavior, impaired control, prioritizing use despite harm, and neuroadaptive changes. Key characteristics also include difficulty controlling the frequency, duration, and intensity of use, withdrawal symptoms after reducing or stopping use (such as tremors, anxiety, insomnia), tolerance (requiring increased doses or intensity for the same effect), functional impairment in personal, family, occupational, or social roles, and continued use despite knowing the harm to physical and mental health.

The full form of \textbf{BOD} is Disorders of bodily distress or bodily experience. These disorders are mental health conditions primarily characterized by bodily symptoms. Patients often show excessive worry about their health, accompanied by persistent physical symptoms such as pain, fatigue, and digestive issues, among others. These symptoms typically cannot be fully explained by routine medical examinations.

The full form of \textbf{STRESS} is Disorders specifically associated with stress. These disorders are a group of mental health conditions triggered by clearly identifiable stressors or traumatic events. Patients typically present with acute stress disorder, which occurs within days of exposure to extreme stressors and is characterized by intense fear, helplessness, and flashbacks. Post-traumatic stress disorder (PTSD) lasts longer and is marked by the recurrent re-experiencing of traumatic events, avoidance behaviors, and heightened arousal. Adjustment disorder is another form, where emotional or behavioral problems arise from identifiable life events that exceed normal coping mechanisms, typically occurring within one month of the event and affecting daily functioning.

The full form of \textbf{DISR} is Disruptive behavior or dissocial disorders. Impulse control disorders are a group of mental health conditions characterized by the difficulty of patients to resist strong, inappropriate desires or impulses, leading to the performance of certain specific behaviors. These behaviors often cause long-term harm to the individual or others, or result in significant impairment in important functional areas such as social, family, or occupational domains. Although patients typically recognize the negative impact of these behaviors, they are unable to resist the impulse in the short term. Common ICDs include pyromania, kleptomania, intermittent explosive disorder, and compulsive sexual behavior disorder.

The full form of \textbf{DISS} is Dissociative disorders. Dissociative disorders are complex mental health conditions primarily characterized by the involuntary partial or complete disintegration of psychological functions such as identity, memory, consciousness, perception, or behavior. These symptoms are not directly caused by drugs, substance abuse, or other psychiatric, behavioral, or neurodevelopmental disorders, and must be distinguished from normal manifestations in cultural or religious practices. The main types include, but are not limited to, dissociative amnesia (selective memory loss), depersonalization-derealization disorder (feeling detached from reality), dissociative fugue (manifesting as a confused state), and dissociative identity disorder (formerly known as multiple personality disorder).

The full form of \textbf{ELIM} is Elimination disorders. Elimination disorders refer to the persistent or recurrent inability to control urination or defecation after reaching the normal developmental age. The main manifestations include enuresis, which is the involuntary urination during the night or day, typically in bed or clothing, and encopresis, which is the inappropriate defecation in places such as clothing. This diagnosis is applicable when children fail to achieve typical bladder and bowel control by the expected developmental age (enuresis by age 5 and encopresis by age 4). If the incontinence symptoms can be fully attributed to other health conditions, congenital or acquired abnormalities in the urinary or gastrointestinal systems, or medication side effects, it should not be diagnosed as an elimination disorder.

The full form of \textbf{EAT} is Feeding or eating disorders. Feeding or eating disorders are a group of mental health conditions characterized by abnormal eating behaviors that cannot be explained by other health conditions and do not conform to normal developmental patterns in the cultural context. These include, but are not limited to, rumination/regurgitation, avoidant/restrictive food intake, anorexia nervosa, bulimia, binge eating, and pica. These disorders often involve issues with controlling food intake, concerns about weight and body shape, or inappropriate dietary choices.

The full form of \textbf{PREG} is Mental or behavioral disorders associated with pregnancy, childbirth, or the puerperium. These disorders refer to mental health issues that occur during pregnancy, during childbirth, and within the first six weeks postpartum. Such disorders include, but are not limited to, depressive disorders, anxiety disorders, and psychotic disorders. Based on their severity and specific manifestations, they can be further categorized into non-psychotic disorders (e.g., postpartum depression) and psychotic disorders (e.g., postpartum psychosis). These mental health conditions significantly affect a woman's psychological state and social functioning, requiring professional medical intervention.

The full form of \textbf{MOOD} is Mood disorders. Mood disorders are a group of mental health conditions characterized by significant and persistent changes in mood or emotional states. These disorders include two main types: bipolar disorder and depressive disorder. They are typically manifested by several major types of mood episodes: depressive episodes, which involve persistent low mood, loss of interest, and reduced energy; manic episodes, which involve abnormally elevated mood, increased activity, and reduced need for sleep; mixed episodes, where both depressive and manic symptoms are present simultaneously; and hypomanic episodes, which are similar to manic episodes but less severe, with minimal impairment of social functioning. These mood episodes are not independent diagnostic entities but are the primary components of mood disorders.

The full form of \textbf{NCOG} is Neurocognitive disorders. Neurocognitive disorders refer to major clinical impairments in cognitive function that are acquired, rather than developmental or congenital. Specifically, these disorders do not include cognitive issues that are present from birth, nor do they cover problems that typically emerge during the developmental stage (which are classified as neurodevelopmental disorders). Instead, neurocognitive disorders refer to a decline in cognitive function from a previously attained level.

The full form of \textbf{NDEV} is Neurodevelopmental disorders. Neurodevelopmental disorders are a group of mental or behavioral disorders that appear during an individual's developmental stage, typically in early childhood, especially before school age, and are characterized by developmental deficits that cause impairment in social, academic, or occupational functioning. The core features of these disorders include significant difficulties in acquiring and performing specific intellectual, motor, language, or social functions. While many mental and behavioral disorders can emerge during the developmental period, only those disorders that are primarily characterized by neurodevelopmental impairments are classified under this group.

The full form of \textbf{OCD} is Obsessive-compulsive or related disorders. Obsessive-compulsive or related disorders are a group of mental health conditions characterized by repetitive thoughts and behaviors. The main symptoms include: obsessive thoughts, which are unwanted, recurring ideas, doubts, or impulses, and compulsive behaviors, which are repetitive actions or mental rituals. These symptoms persist despite the patient's awareness of their meaninglessness, leading to significant personal distress and impairment in social functioning.

The full form of \textbf{PERS} is Personality disorders and related traits. Personality disorders refer to a persistent and stable pattern of emotional experiences, cognitive patterns, and behaviors that significantly deviate from the cultural expectations of the individual's background, leading to widespread functional impairment or subjective distress. This pattern typically emerges during adolescence or early adulthood, with its stability confirmed through observation in adulthood. Personality disorders and related personality traits cover two main aspects: prominent personality features or patterns, where individuals exhibit significant deviations in personality traits, but the severity or extent may not fully meet the diagnostic criteria for a personality disorder, and personality disorders, which include various subtypes (such as borderline, antisocial, schizoid, etc.), characterized by persistent functional impairment, and must exclude direct consequences of other psychiatric disorders or substance use.

The full form of \textbf{SCHIZ} is Schizophrenia or other primary psychotic disorders. Schizophrenia and other primary psychotic disorders are a group of mental health conditions characterized by significant impairment in reality testing. The core symptoms include: positive symptoms such as delusions, hallucinations, and disorganized speech; negative symptoms such as flat affect, reduced volition, and social withdrawal; and cognitive dysfunctions such as impairments in attention, memory, and executive function. These symptoms cause the individual's thoughts, emotions, and behaviors to deviate markedly from the cultural norms, and cannot be attributed to other psychiatric disorders, such as bipolar disorder or substance use disorders.

\begin{figure*}[tb]
\centering
\begin{tcolorbox}[
    colframe=black!60,       
    colback=black!5,         
    coltitle=white,
    fonttitle=\bfseries,
    title=\small Specific Disorders List,
    sharp corners,
    boxrule=0.4mm,
    boxsep=2pt,
    left=2pt, right=2pt,
    top=2pt, bottom=2pt,
    before upper={\setlength{\parskip}{0pt}} 
]
\small
["6A20 Schizophrenia","6B00 Generalised anxiety disorder","6A72 Dysthymic disorder","6E20 Mental or behavioural disorders associated with pregnancy, childbirth or the puerperium, without psychotic symptoms","6A61 Bipolar type II disorder","6A00 Disorders of intellectual development","6D70 Delirium","6E21 Mental or behavioural disorders associated with pregnancy, childbirth or the puerperium, with psychotic symptoms","6B20 Obsessive-compulsive disorder","6A40 Catatonia associated with another mental disorder","6D72 Amnestic disorder","6A62 Cyclothymic disorder","6A23 Acute and transient psychotic disorder","6B43 Adjustment disorder","6A21 Schizoaffective disorder","6A70 Single episode depressive disorder","6A73 Mixed depressive and anxiety disorder","6A71 Recurrent depressive disorder","6A24 Delusional disorder","6A60 Bipolar type I disorder","6C20 Bodily distress disorder","6A22 Schizotypal disorder","6A41 Catatonia induced by substances or medications","6D71 Mild neurocognitive disorder","6D85 Dementia due to diseases classified elsewhere","6C4G Disorders due to use of unknown or unspecified psychoactive substances","6C4E Disorders due to use of other specified psychoactive substances, including medications","6C46 Disorders due to use of stimulants including amphetamines, methamphetamine or methcathinone","6B04 Social anxiety disorder","6C40 Disorders due to use of alcohol","6D10 Personality disorder","6A02 Autism spectrum disorder","6B02 Agoraphobia","6B60 Dissociative neurological symptom disorder","6C91 Conduct-dissocial disorder","6C90 Oppositional defiant disorder","6B83 Avoidant-restrictive food intake disorder","6B40 Post traumatic stress disorder","6C4F Disorders due to use of multiple specified psychoactive substances, including medications","6D84 Dementia due to psychoactive substances including medications","6B63 Possession trance disorder","6B03 Specific phobia","6B41 Complex post traumatic stress disorder","6A03 Developmental learning disorder","6C51 Gaming disorder","6D86 Behavioural or psychological disturbances in dementia","6A06 Stereotyped movement disorder","6D83 Frontotemporal dementia","6B80 Anorexia Nervosa","6B23 Hypochondriasis","6C49 Disorders due to use of hallucinogens","6B81 Bulimia Nervosa","6A05 Attention deficit hyperactivity disorder","6B62 Trance disorder","6B84 Pica","6B21 Body dysmorphic disorder","6B05 Separation anxiety disorder","6B64 Dissociative identity disorder","6B61 Dissociative amnesia","6A04 Developmental motor coordination disorder","6C4D Disorders due to use of dissociative drugs including ketamine and phencyclidine [PCP]","6B66 Depersonalization-derealization disorder","6B25 Body-focused repetitive behaviour disorders","6B42 Prolonged grief disorder","6B82 Binge eating disorder","6C44 Disorders due to use of sedatives, hypnotics or anxiolytics","6B22 Olfactory reference disorder","6C01 Encopresis","6D81 Dementia due to cerebrovascular disease","6B06 Selective mutism","6C50 Gambling disorder","6B24 Hoarding disorder","6C41 Disorders due to use of cannabis","6B85 Rumination-regurgitation disorder","6C47 Disorders due to use of synthetic cathinones"]
\normalsize
\end{tcolorbox}

\captionsetup{aboveskip=4pt, belowskip=0pt, width=\linewidth}
\caption{\textbf{Specific Disorders List.} A list of 76 disorders mentioned in the paper, including their corresponding codes and specific names.}
\label{Disorders_List}
\end{figure*}

\subsection{Knowledge Base}
\label{knowledge_base}
 In our work, we have developed a comprehensive and structured psychiatric knowledge base, denoted as $\mathcal{K}$, to facilitate diagnostic reasoning and improve model training. This knowledge base was designed through extensive collaboration with expert psychiatrists, ensuring it is grounded in clinical relevance and real-world psychiatric practice. The structure of $\mathcal{K}$ is organized into a two-tiered hierarchy: the top level consists of broad diagnostic categories, such as mood disorders or psychotic disorders, while the lower level consists of more specific disorders, such as schizophrenia or bipolar disorder. Each entry in $\mathcal{K}$ contains both general and specific clinical information. For each category, the knowledge base includes formal definitions, prototypical symptom clusters, common functional impairments, and the range of conditions that fall under the category. For example, in mood disorders, descriptors include definitions of typical mood disturbances like depressive or manic episodes and their impact on cognitive and social functioning, as well as the differential diagnostic boundaries with other conditions.

For each specific disorder, $\mathcal{K}$ provides disorder-specific diagnostic criteria aligned with international standards, such as the ICD-11. These criteria include core symptoms, auxiliary features, and typical course. Additionally, the knowledge base outlines physical signs, relevant examination findings, and differential considerations that help distinguish it from other similar disorders. In the case of schizophrenia, for instance, the criteria include the required symptoms, the necessary duration for diagnosis, and the signs needed to differentiate it from other psychotic disorders.

The creation of $\mathcal{K}$ was informed by a combination of authoritative sources, including the ICD-11, clinical guidelines from psychiatric associations, and expert consensus. This approach ensures that the knowledge base remains comprehensive, clinically valid, and directly relevant to psychiatric practice.

Once developed, $\mathcal{K}$ serves as an essential resource for both model training and diagnostic reasoning. It allows for the automatic retrieval of relevant knowledge for a given diagnostic category or disorder, facilitating tasks such as training machine learning models and supporting hypothetico-deductive reasoning in psychiatric diagnosis. This ensures that both model-driven and human-driven diagnostic processes are informed by a reliable and expert-curated knowledge base.

\begin{figure}[t]
  \centering
  \includegraphics[width=\linewidth]{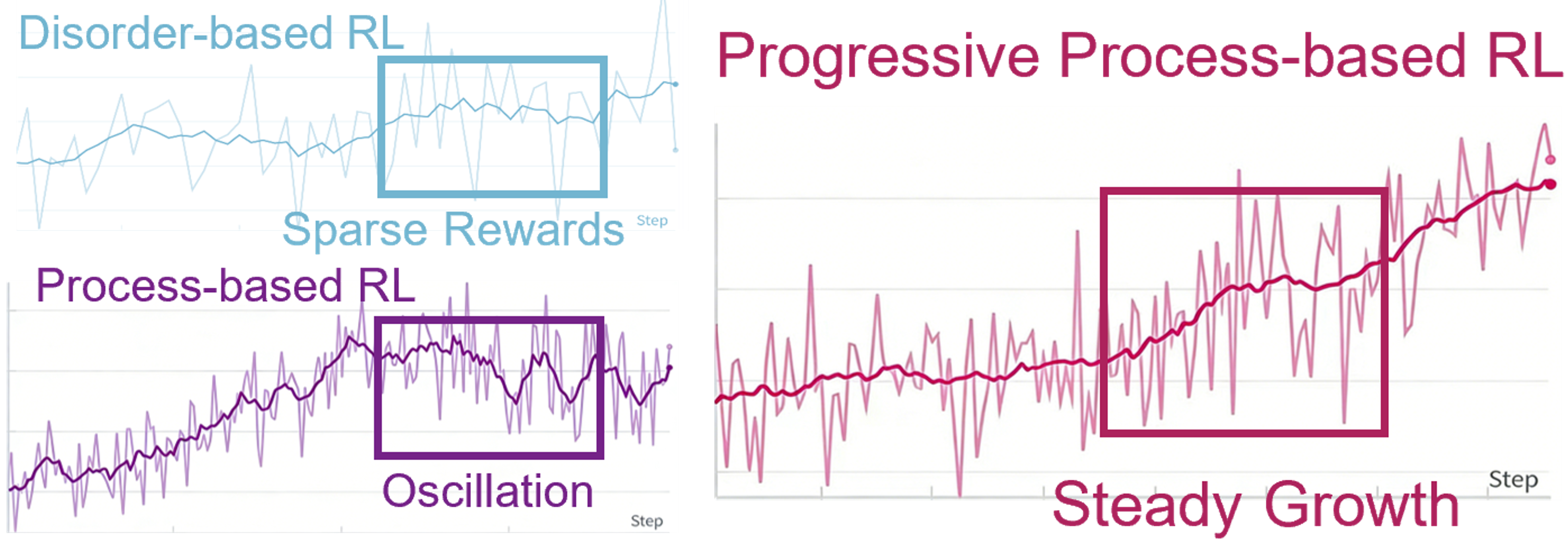}
\caption{Comparison of three reward strategies during reinforcement learning. The progressive process-based reward yields a smoother and more stable training trajectory compared to disorder-based and fixed process-based rewards.}
  \label{RL_CRL.png}
\end{figure}

\begin{figure*}[tb]
\centering
\begin{tcolorbox}[colframe=black!60, colback=black!5, coltitle=white, fonttitle=\bfseries, 
    title=\small Diagnostic Category Knowledge Example, sharp corners, boxrule=0.4mm, 
    boxsep=2pt, left=2pt, right=2pt, top=2pt, bottom=2pt, 
    before upper={\setlength{\parskip}{0pt}}] 
\small
\textbf{Diagnostic Category: Schizophrenia or Other Primary Psychotic Disorders}

\textbf{Code List:}

\begin{itemize}
  \item 6A20
  \item 6A21
  \item 6A22
  \item 6A23
  \item 6A24
\end{itemize}

\textbf{Definition:}
Schizophrenia and other primary psychotic disorders are a group of mental disorders characterized by significant impairment in reality testing. These disorders are marked by:

\begin{itemize}
  \item \textbf{Positive Symptoms:}
  \begin{itemize}
    \item Delusions
    \item Hallucinations
    \item Disorganized speech or behavior
  \end{itemize}
  
  \item \textbf{Negative Symptoms:}
  \begin{itemize}
    \item Emotional blunting
    \item Reduced willpower or motivation
    \item Social withdrawal or isolation
  \end{itemize}
  
  \item \textbf{Cognitive Dysfunction:}
  \begin{itemize}
    \item Impaired attention
    \item Memory deficits
    \item Executive function impairments
  \end{itemize}
\end{itemize}

These symptoms cause the patient's thoughts, emotions, and behaviors to deviate significantly from cultural norms and cannot be attributed to other mental disorders, such as bipolar disorder or substance use disorders.

\normalsize
\end{tcolorbox}

\captionsetup{aboveskip=4pt, belowskip=0pt, width=\linewidth}
\caption{Diagnostic Category Knowledge Example for Schizophrenia and Other Primary Psychotic Disorders, detailing the ICD-11 codes, definition, and key symptoms, including positive, negative, and cognitive dysfunction.}
\label{category}
\end{figure*}
Fig\ref{category} illustrates the broader category-level knowledge for schizophrenia. This category encompasses a range of disorders characterized by significant impairments in reality testing. The figure highlights the diagnostic category itself, with specific ICD-11 codes (6A20 to 6A24), offering a detailed view of the various positive, negative, and cognitive dysfunction symptoms associated with schizophrenia. It presents an overview of the disorder’s general features, helping contextualize the condition in a larger diagnostic framework. This knowledge is essential for understanding the overarching classification of schizophrenia and its primary manifestations, providing the basis for accurate model training and reasoning during diagnostic processes.

\begin{figure*}[tb]
\centering
\begin{tcolorbox}[colframe=black!60, colback=black!5, coltitle=white, fonttitle=\bfseries, 
    title=\small Diagnostic Disorder Knowledge Example, sharp corners, boxrule=0.4mm, 
    boxsep=2pt, left=2pt, right=2pt, top=2pt, bottom=2pt, 
    before upper={\setlength{\parskip}{0pt}}] 
\small

\textbf{Diagnostic Category}: Feeding or Eating Disorders \\

\textbf{ICD-11 Code}: 6B82 \\

\textbf{Specific Disorder}: Binge Eating Disorder \\

\textbf{Definition}: \\
Binge Eating Disorder (BED) is an eating behavior disorder characterized by frequent and repeated binge eating episodes (e.g., at least once a week over a few months). During each binge episode, the individual feels a lack of control over eating and consumes an amount of food significantly greater than usual or in a short period. This eating behavior is often associated with intense distress and negative emotions such as shame, guilt, or disgust. Unlike bulimia nervosa, individuals with BED do not regularly engage in compensatory behaviors such as self-induced vomiting, excessive exercise, or laxative abuse to prevent weight gain.

\textbf{Clinical Manifestations}:

\textbf{Symptoms}: 
\begin{itemize}
    \item \textbf{Uncontrollable Binge Eating}: Consuming large amounts of food in a short time, often accompanied by a sense of loss of control over eating (high: 70\%-90\%).
    \item \textbf{Post-Binge Distress}: Emotional distress after binge eating episodes, including feelings of guilt, shame, or self-loathing (high: 70\%-90\%).
    \item \textbf{Emotional Eating}: Using food to cope with negative emotions such as anxiety or loneliness (moderate: 50\%-70\%).
    \item \textbf{Night Eating}: Frequent eating at night, especially during emotional fluctuations (moderate: 30\%-50\%).
    \item \textbf{Increased Appetite}: Persistent hunger or desire to eat even when not physically hungry (moderate: 30\%-50\%).
    \item \textbf{Social Avoidance}: Avoiding social situations due to embarrassment or fear of eating in public (low: 10\%-30\%).
\end{itemize}

\textbf{Diagnostic Criteria}:

\textbf{Mandatory Criteria (Diagnostic Basis)}: 
\begin{enumerate}
    \item Repeated binge episodes: Occurring at least once a week for at least 3 months. During each binge, the individual feels a lack of control and consumes a larger quantity of food than usual, typically within a short period.
    \item Emotional distress: Significant negative emotions such as shame, guilt, or disgust follow the binge eating episode.
    \item Lack of compensatory behaviors: Unlike bulimia nervosa, individuals with BED do not engage in regular compensatory behaviors (e.g., self-induced vomiting, excessive exercise, or laxative abuse) to prevent weight gain.
\end{enumerate}

\textbf{Supportive Criteria (Clinical and Epidemiological Basis)}:
\begin{itemize}
    \item Frequent night eating: Eating frequently at night, especially when bored or under stress.
    \item Increased appetite: Persistent craving for food even when not hungry.
    \item Social avoidance: Avoiding social situations due to fear of eating in public.
    \item Weight concerns: Significant concern about body weight and shape, with a low self-evaluation.
\end{itemize}

\textbf{Threshold Criteria}: 
\begin{enumerate}
    \item Fulfillment of all conditions in the \textbf{Mandatory Criteria} confirms the diagnosis.
    \item If only some of the \textbf{Mandatory Criteria} are met but multiple \textbf{Supportive Criteria} are present, further assessment of psychosocial factors and family history is necessary.
\end{enumerate}

\normalsize
\end{tcolorbox}

\captionsetup{aboveskip=4pt, belowskip=0pt, width=\linewidth}
\caption{Binge Eating Disorder. Diagram illustrating the diagnostic criteria and clinical manifestations of Binge Eating Disorder, along with its ICD-11 code and specific symptoms.}
\label{fig:binge_eating_disorder}
\end{figure*}

Fig~\ref{fig:binge_eating_disorder} delves deeper into a specific disorder within the category of Feeding or Eating Disorders (ICD-11 code: 6B82). It presents the detailed diagnostic criteria for Binge Eating Disorder (BED), which includes frequent binge episodes, the loss of control over eating, and the absence of compensatory behaviors seen in disorders like bulimia nervosa. The figure outlines both mandatory and supportive criteria for BED, with clinical manifestations such as emotional distress post-binge, emotional eating, and social avoidance. This knowledge representation zooms in on the specific disorder, providing precise diagnostic criteria and clinical features that are essential for accurate diagnosis and clinical decision-making.

\subsection{Benchmark Example}\label{sec:Example}
\begin{figure*}[th]
  \centering
  \includegraphics[width=\linewidth]{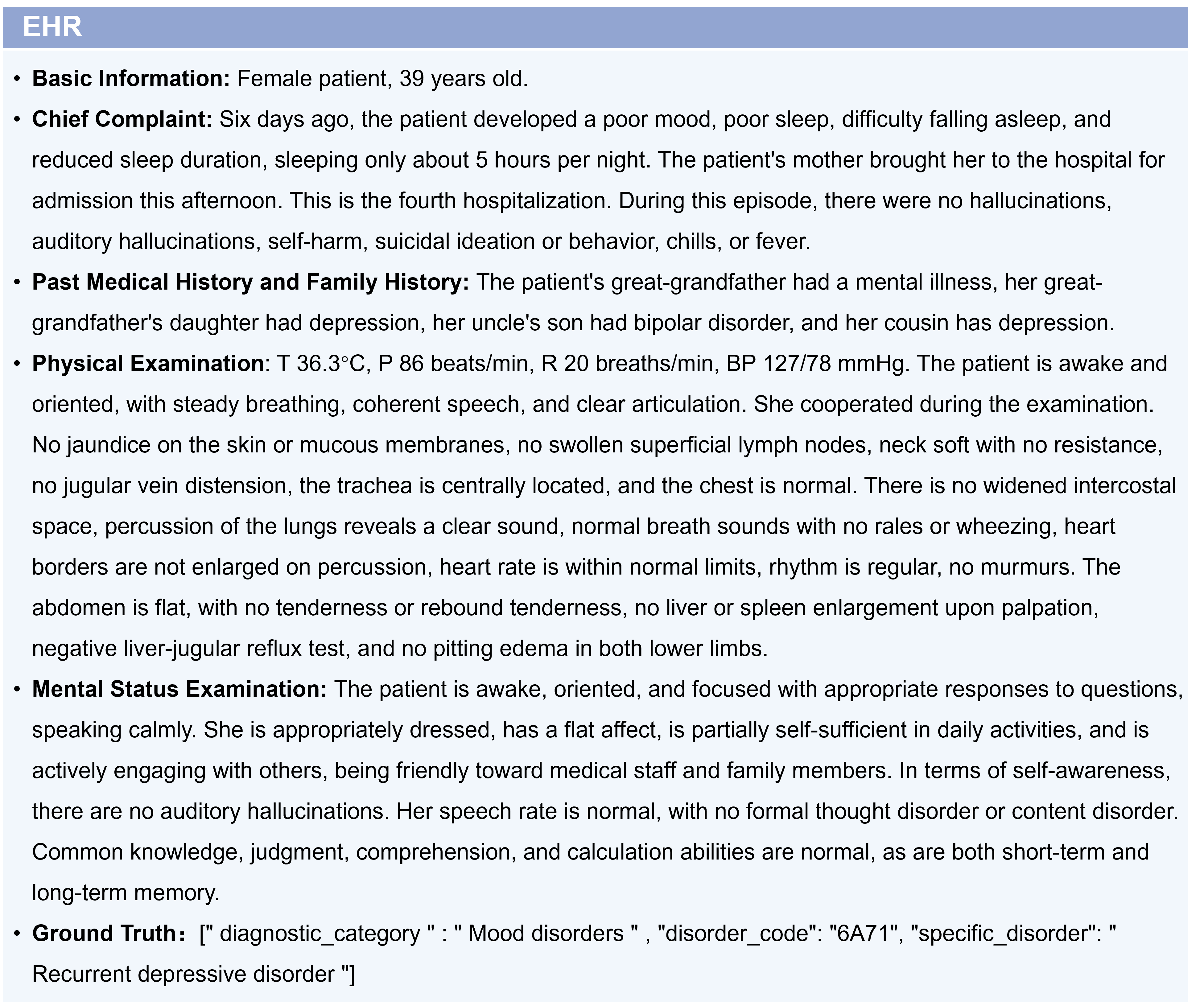}
\caption{\textbf{Electronic Health Record (EHR):} Includes patient details, chief complaint, past medical and family history, physical examination, mental status examination, and ground truth. }
  \label{appendix_case_bench}
\end{figure*}

Each example in the \textbf{MentalDx Bench} consists of two main components: the patient's Electronic Health Record (EHR) and the diagnostic conclusion provided by the doctor. The EHR includes a comprehensive set of details, such as the patient's Basic Information, Chief Complaint, Past Medical and Family History, Physical Examination results, and Mental Status Examination. The doctor's diagnostic conclusion includes a  diagnostic category, a specific disorder from that category, and the corresponding disorder code according to the classification system. A specific example illustrating this structure is shown in the fig \ref{appendix_case_bench}, which demonstrates how all these elements are integrated into a complete case representation. 

\begin{figure*}[h]
  \centering
  \includegraphics[width=\linewidth]{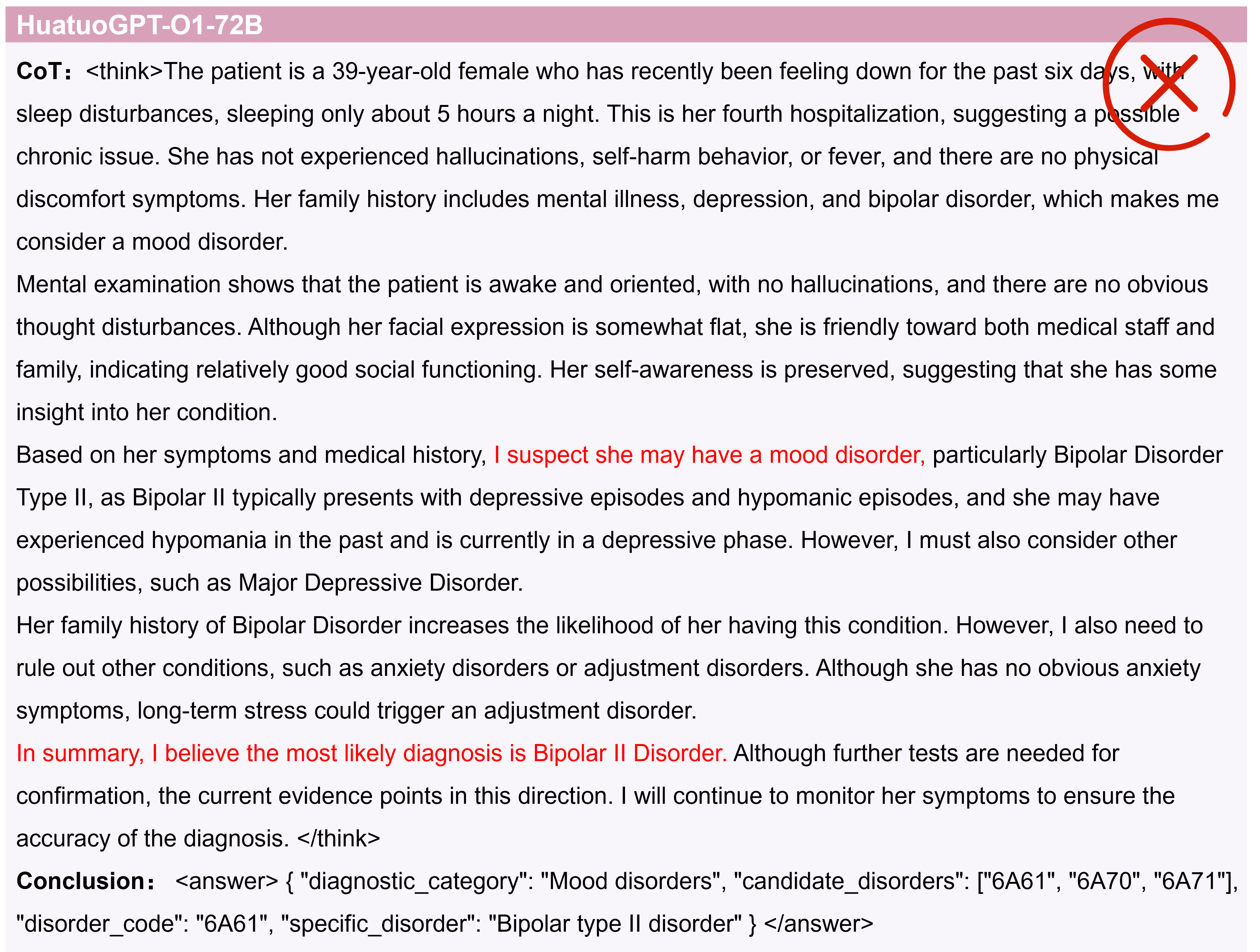}
\caption{HuatuoGPT-O1-72B: Chain of thinking and diagnostic conclusion.}
  \label{appendix_case_bench_huatuo}
\end{figure*}

\begin{figure*}[h]
  \centering
  \includegraphics[width=\linewidth]{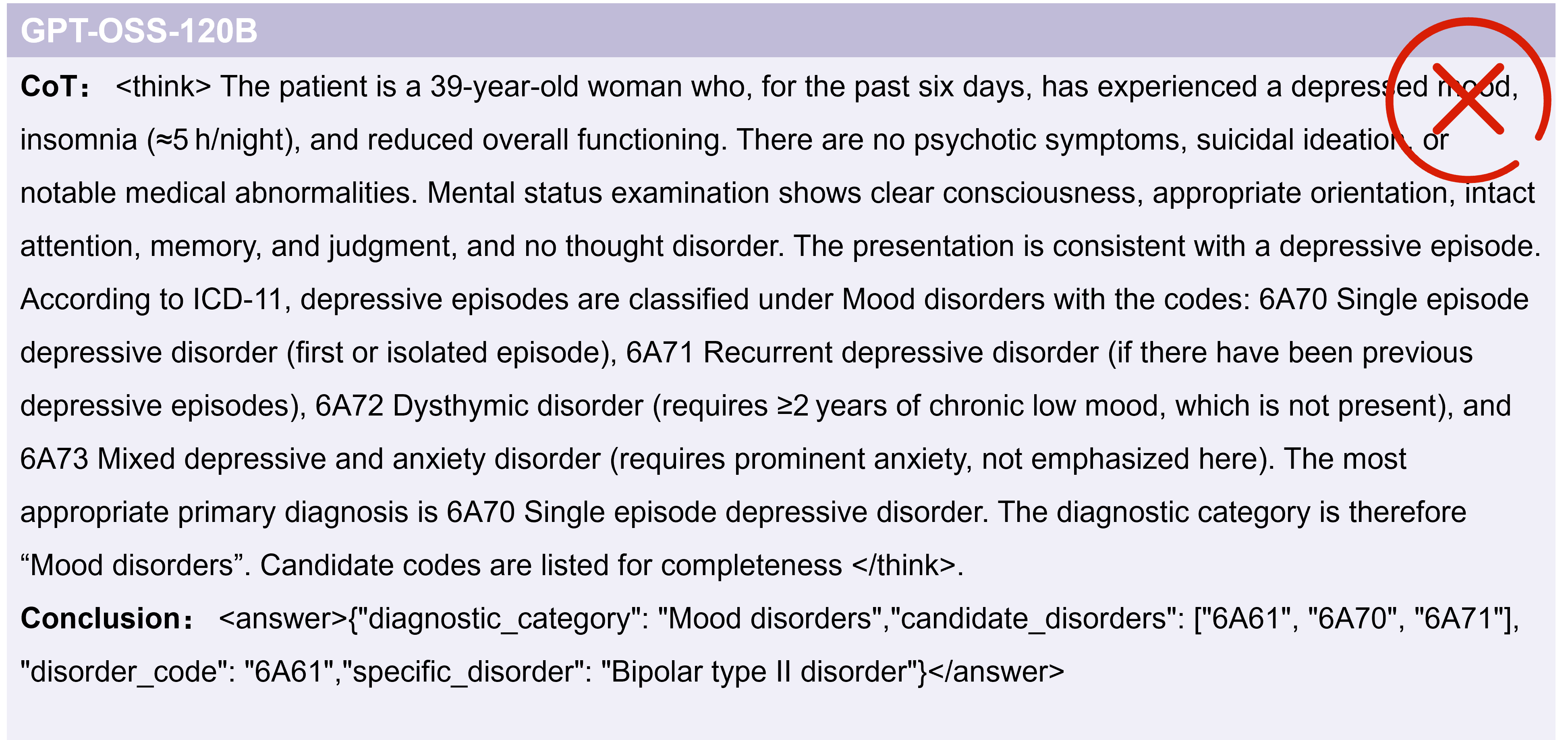}
\caption{GPT-OSS-120B: Chain of thinking and diagnostic conclusion.}
  \label{appendix_case_bench_GPTOSS}
\end{figure*}

Fig\ref{appendix_case_bench_huatuo} and \ref{appendix_case_bench_GPTOSS} display the chain of thinking and final diagnostic results of HuatuoGPT-O1-72B and GPT-OSS-120B, respectively, when diagnosing based on patient medical records. Both models correctly identified the diagnostic category, but there were discrepancies in the diagnosis of the specific disorder. Specifically, HuatuoGPT-O1-72B did not strictly adhere to the information in the medical records during its diagnostic process, instead making assumptions about potential conditions the patient might have, which led to an incorrect disorder diagnosis. Meanwhile, GPT-OSS-120B overlooked key information in the medical records, such as the patient's multiple hospitalizations due to depression, resulting in an inaccurate diagnosis. Although existing models perform reasonably well in certain areas, they still exhibit notable limitations during the diagnostic process. When handling complex cases, models often rely on hypothetical reasoning rather than comprehensively analyzing the patient's medical records or fail to consider important background information, such as hospitalization history, which can ultimately lead to incorrect diagnoses.

\subsection{Benchmark Results}
\label{bench_result}

\begin{table*}[t]

\definecolor{TopOne}{HTML}{c8def7}
\definecolor{TopTwo}{HTML}{e0e9f7}

\centering
\newcolumntype{S}{>{\small}p{3.2cm}}
\centering
\renewcommand{\arraystretch}{1.25}
\resizebox{\textwidth}{!}{
\addtolength{\tabcolsep}{-1.5pt}
\begin{tabular}{S|ccccccccccccccccccc}
\hline
\textbf{Model}& \textit{ANX} & \textit{CATA} & \textit{SUD} & \textit{BOD} & \textit{STRESS}& \textit{DISR} & \textit{DISS} & \textit{ELIM} & \textit{EAT} & \textit{PREG}& \textit{MOOD} & \textit{NCOG} & \textit{NDEV} & \textit{OCD} & \textit{PERS} & \textit{SCHIZ}  & $\mathcal{CA}$ & $\mathcal{DA}$ & $\mathcal{JA}$ \\
\hline
\multicolumn{20}{l}{\textit{\textbf{Small Language Models}}} \\
\hline
Qwen2.5-7B  & 11.11 & 30.00 & 14.12 &10.00 &40.00 & 0.00 & 4.26 &25.00 & 17.14 & 21.74 & 12.37 &13.79 & 18.92 & 12.50 &0.00 &12.50 & 47.47 & 14.75 & 10.96\\

Qwen2.5-14B  & 18.06 &63.33 &29.41 & 0.00 & 52.50 & 44.44 & 21.28 &25.00 &62.86 &82.61 &22.68 &13.79 & 48.65 & 40.00 &0.00 & 10.94  &61.66 & 29.92 & 24.72\\

Qwen3-32B  &30.56 & 20.00 & 22.35 & \cellcolor{TopOne}25.00 & 32.50 &50.00 &31.91 & \cellcolor{TopTwo}75.00 & 48.57 & 52.17 &22.68 & 35.63 &51.35 &62.50 & 15.38 & 1.56 & 58.71 &31.04 &26.26 \\
QWQ-32B & 48.61 & 70.00 & 35.29 & \cellcolor{TopOne}25.00 & 35.00 & 66.67 & 42.55 & \cellcolor{TopTwo}75.00 & 91.43 & \cellcolor{TopTwo}91.30 & 41.24 & 35.63 & 75.68 & 75.00 &0.00 & 29.69   & 76.26 & 47.89 & 46.63 \\

\hline
\multicolumn{20}{l}{\textit{\textbf{Large Language Models}}} \\
\hline

LLaMA-3.1-70B   & 47.22 & 30.00 & 28.24 & 5.00 & 35.00 & \cellcolor{TopTwo}88.89 & 27.66 & \cellcolor{TopTwo}75.00 & 62.86 & 65.22 & 25.77 & 25.29 & 64.86 & 47.50 & 7.69 & 17.19  & 70.65 & 35.53 & 33.15\\
Qwen2.5-72B  & 34.72 & 40.00 & 32.94 & 0.00 & 35.00 & 38.89 & 34.04 & \cellcolor{TopTwo}75.00 & 74.29 & 78.26 & 23.71 & 16.09 & 62.16 & 55.00 & 30.77 & 6.25 & 70.51 & 33.57 & 29.78 \\
Kimi-K2  & 55.56 & 33.33 & 44.71 & 0.00 & \cellcolor{TopTwo}57.50 & 72.22 & 48.94 & \cellcolor{TopOne}100.00 &91.43 & 65.22 & 24.74 & 32.18 & \cellcolor{TopTwo}89.19 & 70.00 &23.08 & 10.94 & 75.70 & 45.08 & 43.40\\
Qwen3-Max  & 50.00 &73.33 & \cellcolor{TopTwo}47.06 & \cellcolor{TopTwo}20.00 & 52.50 & \cellcolor{TopOne}94.44 & 48.94 & \cellcolor{TopOne}100.00 &91.43 &86.96 & 32.99 &48.28 & \cellcolor{TopTwo}89.19 & 82.50 &23.08 & 20.31 & \cellcolor{TopTwo}83.15 & 52.67 & 52.11 \\
GPT-5.1  &66.67 & 53.33 & \cellcolor{TopTwo}47.06 & 5.00 & 42.50 & 77.78 &53.19 & \cellcolor{TopOne}100.00 &91.43 & 82.61 & 38.14 &48.28 &78.38 &85.00 & 30.77 & 40.62 &81.18 & 54.49 & 53.79 \\
DeepSeek-R1  & 50.00 & 63.33 & \cellcolor{TopOne}48.24 &15.00 & \cellcolor{TopOne}60.00 & 77.78 & 61.70 & \cellcolor{TopOne}100.00 & \cellcolor{TopTwo}94.29 & \cellcolor{TopOne}95.65 & 45.36 & 40.23 &78.38 &85.00 & 15.38 &37.50& 78.37 &55.20 &54.49 \\
Claude-Sonnet-4.5  & \cellcolor{TopOne}88.89 & \cellcolor{TopTwo}83.33 & \cellcolor{TopTwo}47.06 &15.00 & 52.50 & \cellcolor{TopTwo}88.89 & 61.70 & \cellcolor{TopOne}100.00 & 85.71 & \cellcolor{TopOne}95.65 &40.21 & \cellcolor{TopTwo}49.43 &78.38 & \cellcolor{TopTwo}87.50 &23.08 & 34.38 & 80.76 & 59.69 & 58.15 \\

\hline
\multicolumn{20}{l}{\textit{\textbf{Medical-specific Large Language Models}}} \\
\hline

ClinicalGPT-R1  & 9.72 & 23.33 & 10.59 & 5.00 & 27.50 & 5.56 & 2.13 & 0.00 & 14.29 & 17.39 & 16.49 & 5.75 & 13.51 & 5.00 & 0.00 & 1.56 & 36.10 & 10.53 & 6.46\\
HuatuoGPT-o1-7B  & 9.72 & 20.00 & 12.94 & 5.00 & 25.00 & 5.56 & 0.00 & 50.00 & 5.71 & 8.70 & 15.46 & 6.90 & 5.41 & 10.00 & 0.00 & 17.19 & 43.26 & 11.24 & 8.01\\
GPT-OSS-20B  & 52.78 & 63.33 & 23.53 & 15.00 & 35.00 & 83.33 & 44.68 & \cellcolor{TopOne}100.00 & 77.14 & 82.61 & 34.02 & 32.18 & 70.27 & 72.50 & 0.00 & 45.31 & 74.30 & 45.65 & 43.96\\
MedGemma-27B  & 54.17 & 50.00 & 28.24 & \cellcolor{TopOne}25.00 & 42.50 & 55.56 & 29.79 & \cellcolor{TopTwo}75.00 & 74.29 & 65.22 & 25.77 & 32.18 & 72.97 & 70.00 & 23.08 & 40.62 & 66.99 & 42.84 & 39.89\\
Baichuan-M2  & 38.89 & 66.67 & 35.29 & 5.00 & 37.50 & 72.22 & 42.55 & \cellcolor{TopTwo}75.00 & 88.57 & \cellcolor{TopTwo}91.30 & 28.87 & 37.93 & 62.16 & 67.50 & 0.00 & 35.94 & 64.61 & 44.38 & 41.71\\
HuatuoGPT-o1-72B  & 45.83 & 50.00 & 35.29 & 10.00 & 35.00 & 66.67 & 40.43 & \cellcolor{TopTwo}75.00 & 80.00 & 69.57 & 26.80 & 19.54 & 70.27 & 65.00 & 30.77 & 18.75 & 74.02 & 39.75 & 36.94\\
GPT-OSS-120B  & 54.17 & 66.67 & 38.82 & \cellcolor{TopTwo}20.00 & 50.00 & \cellcolor{TopTwo}88.89 & 51.06 & 50.00 & \cellcolor{TopTwo}94.29 & 65.22 & 41.24 & 43.68 & 70.27 & 75.00 & 38.46 & 45.31 & 75.84 & 52.53 & 50.00\\

\hline
\multicolumn{20}{l}{\textit{\textbf{Ours}}} \\
\hline

MentalSeek-Dx-7B & 81.94 & 70.00 & \cellcolor{TopOne}48.24 & \cellcolor{TopOne}25.00 & 52.50 & \cellcolor{TopTwo}88.89 & \cellcolor{TopTwo}72.34 & \cellcolor{TopOne}100.00 & \cellcolor{TopOne}97.14 & 86.96 & \cellcolor{TopTwo}54.64 & \cellcolor{TopOne}66.67 & \cellcolor{TopOne}94.59 & \cellcolor{TopOne}90.00 & \cellcolor{TopOne}61.54 & \cellcolor{TopTwo}56.25 & 82.58 & \cellcolor{TopTwo}67.56 & \cellcolor{TopTwo}67.13  \\

MentalSeek-Dx-14B & \cellcolor{TopTwo}84.72 & \cellcolor{TopOne}90.00 & \cellcolor{TopOne}48.24 & \cellcolor{TopTwo}20.00 & 52.50 & \cellcolor{TopTwo}88.89 & \cellcolor{TopOne}82.98 & \cellcolor{TopOne}100.00 & \cellcolor{TopOne}97.14 & \cellcolor{TopTwo}91.30 & \cellcolor{TopOne}57.73 & \cellcolor{TopOne}66.67 & \cellcolor{TopTwo}89.19 & 85.00 & \cellcolor{TopTwo}53.85 & \cellcolor{TopOne}67.19 & \cellcolor{TopOne}83.99 & \cellcolor{TopOne}70.08 & \cellcolor{TopOne}69.38 \\

\hline

\end{tabular}
}
\caption{Performance of mainstream LLMs on MentalDx Bench (\%). Models are grouped by parameter scale to examine how size affects performance across diagnostic categories. Each cell reports a model’s score on the corresponding task. For clarity, the best result in each category is shown in dark, the second-best in light.}
\label{tab:all_llm_performance}
\end{table*}

Tab~\ref{tab:all_llm_performance} presents the performance of various language models on the \textbf{MentalDx Bench}, with results shown for individual diagnostic categories as well as for overall category accuracy (\textbf{CA}), disorder accuracy (\textbf{DA}), and joint accuracy (\textbf{JA}). Each cell represents the model's score on a specific diagnostic task, expressed as a percentage. \textbf{CA} refers to the overall category diagnostic accuracy, \textbf{DA} refers to the accuracy in diagnosing disorders, and \textbf{JA} is the accuracy when both category and disorder diagnoses are correct simultaneously. The best result in each category is highlighted in dark, with the second-best in light.

Our models, \textbf{MentalSeek-Dx-7B} and \textbf{MentalSeek-Dx-14B}, show outstanding performance across a number of key diagnostic tasks. Specifically, \textbf{MentalSeek-Dx-7B} achieves strong results in \textit{ANX} (81.94\%), \textit{CATA} (70.00\%), \textit{SUD} (48.24\%), and \textit{BOD} (25.00\%), demonstrating its capability in accurately diagnosing these critical mental health categories.

Furthermore, \textbf{MentalSeek-Dx-14B} shows further improvements, particularly in \textit{CATA} (90.00\%) and \textit{DISS} (82.98\%), surpassing other models in these areas. This demonstrates that scaling up the model enhances diagnostic accuracy, reinforcing the effectiveness of \textbf{MentalSeek-Dx} in handling complex mental health tasks.

In terms of overall category accuracy (\textbf{CA}), both \textbf{MentalSeek-Dx-7B} and \textbf{MentalSeek-Dx-14B} perform excellently, with \textbf{MentalSeek-Dx-14B} achieving the highest accuracy at 67.56\%, outpacing many other large language models. The disorder accuracy (\textbf{DA}) also shows strong results, particularly with \textbf{MentalSeek-Dx-14B} at 69.38\%, further highlighting the model's strength in diagnosing mental health disorders.

Notably, in terms of joint accuracy (\textbf{JA}), which considers both category and disorder diagnosis, \textbf{MentalSeek-Dx-7B} and \textbf{MentalSeek-Dx-14B} again perform impressively, with \textbf{MentalSeek-Dx-14B} reaching 69.38\%, showcasing its robust capability in simultaneously diagnosing both the overall category and the specific disorder.

Overall, the \textbf{MentalSeek-Dx} models excel across diagnostic categories, overall category accuracy, disorder accuracy, and joint accuracy, outperforming many other state-of-the-art and medical-specific models. This demonstrates the innovative contribution of \textbf{MentalSeek-Dx} in the field of psychiatric diagnosis, especially with the breakthroughs achieved by \textbf{MentalSeek-Dx-7B} and \textbf{MentalSeek-Dx-14B} in applying large-scale language models to mental health diagnosis.

\subsection{Error Types Explanation}
\label{error_types}
\begin{table*}[t]
\resizebox{\linewidth}{!}{
\begin{tabular}{cll}
\toprule
\textbf{Abbreviation} & \textbf{Full Form} & \textbf{Explanation} \\
\midrule
NR & No Reasoning & Directly outputs an incorrect conclusion without reasoning. \\
WI & Wrong Inference & Symptoms are identified but the reasoning leads to an incorrect diagnosis. \\
IE & Identification Error & A list of potential diagnoses is given but the correct diagnosis is not identified. \\
OP & Overprediction & No clear primary diagnosis, but multiple possible diagnoses are listed. \\
RA & Refusal to Answer & Refuses to answer citing safety concerns. \\
Typo & Typographical Error & Character errors that prevent matching the correct diagnosis. \\
Other & Uncategorized Errors & Other errors that don't fit into the above categories. \\
\bottomrule
\end{tabular}
}
\caption{Explanation of abbreviations for error types in diagnosis}
\label{tab:error_types}
\end{table*}

Tab~\ref{tab:error_types} presents a breakdown of different types of errors observed in the diagnostic processes, each represented by an abbreviation. NR (No Reasoning) indicates cases where the model outputs an incorrect conclusion without providing any reasoning. WI (Wrong Inference) occurs when symptoms are correctly identified, but the reasoning leads to an incorrect diagnosis. IE (Identification Error) happens when the model lists potential diagnoses but fails to identify the correct one. OP (Overprediction) refers to situations where the model provides multiple possible diagnoses without a clear primary diagnosis. RA (Refusal to Answer) happens when the model refuses to answer, often citing safety concerns. Typo (Typographical Error) represents errors in characters that prevent the correct diagnosis from being made. Lastly, Other (Uncategorized Errors) covers any errors that do not fit into the above categories. These categories help to identify common pitfalls and guide improvements in diagnostic models.

As shown in Figure~\ref{fig_annotation.png}, the most frequent error is \textbf{Wrong Inference (WI)}, with a score of 89, indicating that this error occurred the most often in the dataset. The least frequent error is \textbf{Refusal to Answer (RA)}, with a score of only 8, representing the fewest occurrences among the error types. The distribution of errors revealed that the majority of mistakes fell into three primary categories: \textit{wrong inference} (25\%), \textit{identification failure} (22\%), and \textit{symptom omission} (21\%). These categories reflect key areas where the models struggled, with \textit{wrong inference} representing errors in the diagnostic reasoning process, \textit{identification failure} reflecting difficulties in recognizing the correct diagnosis from the set of possible conditions, and \textit{symptom omission} pointing to the failure of the models to consider or properly incorporate key symptoms in their diagnostic conclusions.On the other hand, superficial errors, such as \textit{typos} (7\%) and \textit{refusal to answer} (4\%), were considerably less frequent. This finding highlights that the majority of diagnostic errors were not merely due to trivial issues, like typographical mistakes or the model's refusal to respond, but instead stemmed from more profound problems related to the reasoning and understanding of symptoms.

The error distribution further suggests that the dominant sources of failure in the models lie in deeper, more structural flaws in their diagnostic logic. Specifically, many models exhibited fragmented or incomplete diagnostic reasoning. This included unclear or vague delineation of symptoms, meaning the models struggled to differentiate between overlapping or similar conditions. Additionally, the models often failed to conduct a comprehensive analysis of the potential conditions, leading to incomplete or erroneous diagnostic conclusions. Moreover, the inability to effectively perform differential diagnosis, a critical component of psychiatric diagnosis, was a common limitation. These findings emphasize the need for further refinement in model training, particularly in improving the depth of reasoning, symptom analysis, and differential diagnostic capabilities, to reduce errors and enhance diagnostic accuracy.

\subsection{Prompt Details}\label{sec:prompt}

Given a detailed medical record and a list of 16 distinct diagnostic categories, each accompanied by a brief definition and explanation, the model is tasked with inferring the most relevant suspected diagnoses based on the provided medical information. The specific format for the prompt is shown in the fig~\ref{category_classification}. The model should perform a comprehensive analysis of various aspects of the medical record  to assess which diagnostic categories best match the features of the case.
Each diagnostic category has a clear and distinct definition with specific symptom characteristics. The model’s task is to identify which categories most closely align with the information provided in the medical record. The classification process must carefully consider all relevant details in the record, particularly the form of symptom presentation, onset and duration, severity, and factors such as the patient’s age and sex. Additionally, the model should not restrict itself to a single category; rather, it should consider multiple potentially relevant categories based on the medical history. Therefore, the model should be capable of identifying one or more plausible suspected categories, demonstrating a thorough understanding and accurate analysis of the medical record.

\begin{figure*}[tb]
\centering
\begin{tcolorbox}[
    colframe=black!60,       
    colback=black!5,         
    coltitle=white,
    fonttitle=\bfseries,
    title=\small Prompt for the Category classification,
    sharp corners,
    boxrule=0.4mm,
    boxsep=2pt,
    left=2pt, right=2pt,
    top=2pt, bottom=2pt,
    before upper={\setlength{\parskip}{0pt}} 
]
\small
\textbf{Task} \\
You are a psychiatric diagnostic expert. I will now provide the patient's medical history. Please analyze the provided medical information and give the corresponding diagnosis, i.e., the relevant category. I will also give you a list of all categories and their corresponding brief explanations.

\textbf{Reference for Diagnostic Categories and Explanations} \\
\{
"Neurodevelopmental disorders": "Neurodevelopmental disorders are a group of mental or behavioral disorders that appear during an individual's developmental stage, typically in early childhood, especially before school age, and are characterized by developmental deficits that cause impairment in social, academic, or occupational functioning. The core features of these disorders include significant difficulties in acquiring and performing specific intellectual, motor, language, or social functions. While many mental and behavioral disorders can emerge during the developmental period, only those disorders that are primarily characterized by neurodevelopmental impairments are classified under this group", \\
"Schizophrenia or other primary psychotic disorders": "Schizophrenia and other primary psychotic disorders are a group of mental health conditions characterized by significant impairment in reality testing. The core symptoms include: positive symptoms such as delusions, hallucinations, and disorganized speech; negative symptoms such as flat affect, reduced volition, and social withdrawal; and cognitive dysfunctions such as impairments in attention, memory, and executive function. These symptoms cause the individual's thoughts, emotions, and behaviors to deviate markedly from the cultural norms, and cannot be attributed to other psychiatric disorders, such as bipolar disorder or substance use disorders.", \\
\text{... other categories ...}
\}

\textbf{Output Format:} \\
The diagnostic results should be output in JSON format, including: \\
- \texttt{"diagnostic\_category"}: Choose the most likely category or categories from the above list of disorders.

\textbf{Note} \\
- Do not output \texttt{'''json} or similar content, strictly follow the output example format.

\textbf{Output Example:} \\
\texttt{-------------} \\
\texttt{\{} \\
\texttt{"id":"123456"} \\
\texttt{"diagnostic\_category":[ "Disorders specifically associated with stress"]} \\
\texttt{\}}

\textbf{Input:} \\
- Test medical record: \texttt{<MEDICAL\_RECORD>}

\normalsize
\end{tcolorbox}

\captionsetup{aboveskip=4pt, belowskip=0pt, width=\linewidth}
\caption{Prompt for the Category classification}
\label{category_classification}
\end{figure*}

Based on the suspected categories identified, the system will activate the classification tasks for all specific diseases associated with these categories. A total of 76 different diseases will need to be analyzed and classified in detail. The classification tasks for each disease will be handled according to the features of its respective category, ensuring that the diagnostic results for each disease accurately reflect its unique clinical manifestations and pathological characteristics.

In practice, these classification tasks will be processed in parallel, meaning that for each disease classification requirement, the system will perform multiple disease analyses simultaneously, rather than processing them sequentially one by one. This parallel processing approach not only improves efficiency but also ensures the independence and accuracy of the diagnosis for each disease. The specific format for the prompt is shown in the fig~\ref{category_classification}.

By activating the corresponding classification tasks and processing them in parallel, the system can quickly and comprehensively analyze the clinical characteristics of each disease, combining the existing suspected category information to make accurate inferences about each disease's diagnosis.
\begin{figure*}[tb]
\centering
\begin{tcolorbox}[
    colframe=black!60,       
    colback=black!5,         
    coltitle=white,
    fonttitle=\bfseries,
    title=\small Prompt for the Disorder diagnosis,
    sharp corners,
    boxrule=0.4mm,
    boxsep=2pt,
    left=2pt, right=2pt,
    top=2pt, bottom=2pt,
    before upper={\setlength{\parskip}{0pt}} 
]
\small
\textbf{Task} \\
You are a psychiatric diagnostic expert. I will provide the patient’s medical record, along with the specific disorder and its diagnostic criteria. Your task is to determine whether the patient meets the criteria for the given disorder.

\textbf{Disorder and Diagnostic Criteria} \\
\textbf{Disorder:} "6A20" \\
\textbf{Diagnostic Criteria:}
\begin{itemize}
  \item \textbf{Mandatory Criteria (for Diagnosis):}
  \begin{itemize}
    \item \textbf{Symptom Duration:} Core symptoms (positive symptoms, negative symptoms, or cognitive impairment) must persist for at least one month.
    \item \textbf{Functional Impairment:} Significant impairment in important areas such as work, education, and social interactions.
    \item \textbf{Exclusion of Other Causes:} Exclude similar symptoms caused by other medical conditions (such as brain tumors or epilepsy) or substance use (including drug abuse).
  \end{itemize}
  \item \textbf{Supporting Criteria (Clinical and Epidemiological Evidence):}
  \begin{itemize}
    \item \textbf{Typical Clinical Presentation:}
    \begin{itemize}
      \item \textbf{Positive Symptoms:} Hallucinations (especially auditory hallucinations), delusions (paranoid delusions, delusions of reference, etc.), thought disorder (thought fragmentation, poverty of thought).
      \item \textbf{Negative Symptoms:} Blunted affect, social withdrawal, lack of motivation, poverty of speech.
      \item \textbf{Cognitive Impairment:} Attention deficits, memory problems, executive function impairments.
      \item \textbf{Behavioral Abnormalities:} Stereotyped movements, bizarre behaviors, catatonia (e.g., waxy flexibility, echopraxia).
    \end{itemize}
    \item \textbf{Neuropsychological Testing:} Impairment in multiple cognitive domains.
  \end{itemize}
  \item \textbf{Threshold Criteria:}
  \begin{itemize}
    \item The patient must meet all conditions of the “Mandatory Criteria” and exhibit at least two core symptoms (either positive, negative, or cognitive impairment).
  \end{itemize}
\end{itemize}

The diagnosis process should also take into account the patient's detailed medical history, family history, and clinical presentation to ensure the exclusion of other potential causes.

\textbf{Output Format:}
\begin{itemize}
  \item Provide the diagnosis result in JSON format.
  \item Do not include any extra formatting like \texttt{'''json} or similar. Strictly follow the output sample format.
\end{itemize}

\textbf{Output Sample:}
\begin{verbatim}
{
    "id": "123456",
    "has_disorder_6A20": "no"
}
\end{verbatim}

\textbf{Input:}
\begin{itemize}
  \item Patient's medical record: \texttt{<MEDICAL\_RECORD>}
\end{itemize}

\normalsize
\end{tcolorbox}

\captionsetup{aboveskip=4pt, belowskip=0pt, width=\linewidth}
\caption{Prompt for the Category classification}
\label{Disorder_diagnosis}
\end{figure*}

\subsection{Refinement Pipeline Details}
\label{refinement_pipline_details}
Given a specified disorder, the original medical record, the diagnostic criteria for the specified disorder, clinical manifestations, and common scenarios, the model should perform a detailed refinement and modification of the original medical record based on this information, ensuring that the record aligns solely with the characteristics and criteria of the specified disorder. During this modification process, the model should thoroughly analyze and comprehend all the diagnostic criteria and typical clinical manifestations of the specified disorder, ensuring that every part of the medical record adheres to the relevant requirements of the disorder. Any symptoms, signs, or medical history unrelated to the specified disorder should be removed or adjusted to avoid misleading or irrelevant descriptions.

The refined medical record should emphasize and retain symptoms, signs, medical history, and other relevant elements that are highly associated with the specified disorder, ensuring that the record perfectly aligns with the diagnostic criteria for the disorder. Special attention should be given to the typical symptoms in the clinical presentation, symptom duration, the patient's medical history, and other factors that may influence the diagnosis. Furthermore, the model should ensure that the modified medical record meets the standards of medical writing in terms of semantics, logic, and structure, ensuring clarity and ease of understanding, while avoiding any ambiguous or contradictory descriptions that could lead to misdiagnosis or inaccurate conclusions. The specific format for the prompt is shown in the Fig~\ref{Selected_Disorder},\ref{Disorder Reasoning}and \ref{Category Reasoning}.

Ultimately, the modified medical record should fully comply with the diagnostic requirements for the specified disorder, making it a reliable clinical reference for subsequent diagnosis and treatment decisions.

\begin{figure*}[tb]
\centering
\begin{tcolorbox}[
    colframe=black!60,       
    colback=black!5,         
    coltitle=white,
    fonttitle=\bfseries,
    title=\small Prompt for Selected Disorder,
    sharp corners,
    boxrule=0.4mm,
    boxsep=2pt,
    left=2pt, right=2pt,
    top=2pt, bottom=2pt,
    before upper={\setlength{\parskip}{0pt}} 
]
\small

\textbf{Task Definition} \\
You are a professional psychiatrist. You are given an anonymized clinical case that needs to be revised and completed to strictly align with the diagnostic criteria of \texttt{<selected\_disorder>}. Any irrelevant or non-conforming symptoms should be removed.

\textbf{Input} \\
You will receive the following information for ICD-11 code 6A20: \\
\textbf{Diagnostic Criteria} \\
\texttt{<disorder\_category>} \\

\textbf{Clinical Manifestations} \\
\texttt{<clinical\_manifestation>} \\

\textbf{Record to Revise} \\
\texttt{<record\_input>} \\

\textbf{Core Requirements} 
\begin{itemize}
    \item The revised case must fully meet the diagnostic criteria of the target disorder. If any core criteria are missing, you must complete them through generation.
    \item Remove unrelated symptoms or information to avoid introducing comorbid or confounding conditions.
\end{itemize}

\textbf{Output Format} \\
Output in JSON format only, without any explanations or \texttt{```json} tags, as follows:

\begin{verbatim}
{
	"revised_record": "..."
}
\end{verbatim}

\normalsize
\end{tcolorbox}

\captionsetup{aboveskip=4pt, belowskip=0pt, width=\linewidth}
\caption{Prompt for Selected Disorder}
\label{Selected_Disorder}
\end{figure*}

\subsection{Trajectory Building Details}
\label{trajectory_building_details}

We construct a Hypothetico-Deductive Reasoning Trajectory that explicitly mirrors the clinical diagnostic process adopted by human psychiatrists. Specifically, the trajectory is decomposed into three sequential reasoning stages: category-level hypothesis formation, disorder-level hypothesis narrowing, and criteria-based differential confirmation. At the first stage, the model is guided to reason from the raw medical record toward a broad diagnostic category by grounding symptom patterns and illness course in formal clinical definitions. Conditioned on this category-level hypothesis, the second stage further refines the reasoning to identify a small set of plausible disorders within the category, leveraging detailed clinical manifestations to generate competing diagnostic hypotheses. Finally, the model performs strict differential reasoning by systematically aligning patient evidence with diagnostic criteria and key exclusion points, culminating in a single confirmed diagnosis. By structuring reasoning as a progressive hypothesis generation and elimination process, our trajectory enforces a transparent, stepwise diagnostic logic that reflects real-world psychiatric decision-making, rather than allowing conclusions to emerge implicitly or heuristically.

\begin{figure*}[tb]
\centering
\begin{tcolorbox}[
    colframe=black!60,       
    colback=black!5,         
    coltitle=white,
    fonttitle=\bfseries,
    title=\small Prompt for Category Reasoning,
    sharp corners,
    boxrule=0.4mm,
    boxsep=2pt,
    left=2pt, right=2pt,
    top=2pt, bottom=2pt,
    before upper={\setlength{\parskip}{0pt}} 
]
\small

You are an experienced psychiatric expert with extensive clinical diagnostic experience. You will first be provided with: the patient's medical record, diagnostic category, and clinical definition, and need to supplement the reasoning process between the two.

\textbf{Requirements:}
\begin{itemize}
    \item Act as a professional doctor, presenting your internal diagnostic monologue.
    \item Start with the patient's medical record, gradually combining the clinical definition standards to pinpoint the category.
    \item Evidence-based reasoning should naturally link the reasoning process with the definition.
    \item Begin with: "Alright, let me go through the medical reasoning step by step. First, I need to combine the symptom combinations and the course of the illness to frame the broad syndrome category this patient belongs to..."
\end{itemize}

\textbf{Input Information}
\begin{itemize}
    \item \textbf{Patient Medical Record:} \texttt{<INPUT>}
    \item \textbf{Diagnostic Category:} \texttt{<DIAGNOSTIC\_CATEGORY>}
    \item \textbf{Clinical Definition:} \texttt{<DEFINITION>}
\end{itemize}

\textbf{Output Example}
\begin{verbatim}
{
    "category_reasoning": "Alright, let me go through the medical reasoning step by step.
                            First, I need to combine the symptom combinations and the
                            course of the illness to frame the broad syndrome category
                            this patient belongs to...", // Here, place the reasoning
                            process you generate.
    "diagnostic_category": "<DIAGNOSTIC_CATEGORY>"
}
\end{verbatim}

\normalsize
\end{tcolorbox}

\captionsetup{aboveskip=4pt, belowskip=0pt, width=\linewidth}
\caption{Prompt for Category Reasoning}
\label{Category Reasoning}
\end{figure*}

\begin{figure*}[tb]
\centering
\begin{tcolorbox}[
    colframe=black!60,       
    colback=black!5,         
    coltitle=white,
    fonttitle=\bfseries,
    title=\small Prompt for Disorder Reasoning,
    sharp corners,
    boxrule=0.4mm,
    boxsep=2pt,
    left=2pt, right=2pt,
    top=2pt, bottom=2pt,
    before upper={\setlength{\parskip}{0pt}} 
]
\small

You are an experienced psychiatric expert with extensive clinical diagnostic experience. You will first be provided with: the patient's medical record, the corresponding list of disorders, and clinical manifestations. Your task is to list the possible corresponding disorders the patient may have based on the provided information.

\textbf{Requirements:}
\begin{itemize}
    \item Act as a professional doctor, presenting your internal diagnostic monologue.
    \item Start with the patient's medical record, gradually combining the clinical manifestations to pinpoint 2-3 suspected disorders.
    \item Use evidence-based reasoning to naturally and closely connect the reasoning process with the clinical manifestations.
    \item Begin with: "Next, I need to work within the syndrome range and, combining clinical symptoms, deduce the possible disorders..."
\end{itemize}

\textbf{Input Information:}
\begin{itemize}
    \item \textbf{Patient's Medical Record:} \texttt{<INPUT>}
    \item \textbf{Candidate Disorder List:} \texttt{<SPECIFIC\_DISORDER>}
\end{itemize}

\textbf{Output Example:}
\begin{verbatim}
{
    "disorder_reasoning": "Next, I need to work within the syndrome range and, combinin
                           clinical symptoms, deduce the possible disorders...", 
                           Here, place the reasoning process you generate
    "candidate_disorders": [...] // "code" list
}
\end{verbatim}

\normalsize
\end{tcolorbox}

\captionsetup{aboveskip=4pt, belowskip=0pt, width=\linewidth}
\caption{Prompt for Differential Reasoning}
\label{Disorder_diagnosis}
\end{figure*}

\begin{figure*}[tb]
\centering
\begin{tcolorbox}[
    colframe=black!60,       
    colback=black!5,         
    coltitle=white,
    fonttitle=\bfseries,
    title=\small Prompt for Differential Reasoning,
    sharp corners,
    boxrule=0.4mm,
    boxsep=2pt,
    left=2pt, right=2pt,
    top=2pt, bottom=2pt,
    before upper={\setlength{\parskip}{0pt}} 
]
\small

You are an experienced psychiatric expert with extensive clinical diagnostic experience. You will first be provided with: the patient's medical record, a list of suspected disorders, the corresponding diagnostic criteria, and the final confirmed diagnosis. Your task is to supplement the differential diagnostic reasoning process.

\textbf{Requirements:}
\begin{itemize}
    \item Act as a professional doctor, presenting your internal diagnostic monologue.
    \item Strictly follow the diagnostic criteria for differential diagnosis, and conclude the diagnosis as \texttt{<CODE>}.
    \item Use evidence-based reasoning to naturally and closely link the differential reasoning process with the diagnostic criteria.
    \item Begin with: "Finally, I need to combine the diagnostic criteria and key differential points to complete the differentiation and confirm the final diagnosis..."
\end{itemize}

\textbf{Input Information:}
\begin{itemize}
    \item \textbf{Patient's Medical Record:} \texttt{<INPUT>}
    \item \textbf{Candidate Disorder List:} \texttt{<SPECIFIC\_DISORDER>}
\end{itemize}

\textbf{Output Example:}
\begin{verbatim}
{
    "differential_reasoning": "Finally, I need to combine the diagnostic criteria and key
    differential points to complete the differentiation and confirm the final diagnosis...",
    Here, place the reasoning process you generate
    "confirmed_disorder": "<CODE>"
}
\end{verbatim}

\normalsize
\end{tcolorbox}

\captionsetup{aboveskip=4pt, belowskip=0pt, width=\linewidth}
\caption{Prompt for Differential Reasoning}
\label{Disorder Reasoning}
\end{figure*}

\subsection{Medical Reasoning Quality}
\label{sec:cot}

\begin{table*}[t]
\resizebox{\linewidth}{!}{
\begin{tabular}{cll}
\toprule
\textbf{Abbreviation} & \textbf{Full Form} & \textbf{Explanation} \\
\midrule
CLR & Clinical Logical Rigor & The CoT aligns with the clinical diagnostic process in psychiatry.\\
ES & Evidence Strength & The CoT is based on authoritative guidelines, consensus, or reliable research.\\
CC & Critical Completeness & The CoT adequately utilizes both positive and negative information.\\
DDA & Differential Diagnosis & The CoT's differential diagnoses are systematically listed and excluded.\\

\bottomrule
\end{tabular}
}
\caption{Explanation of abbreviations for CoT quality assessment in LLMs generation}
\label{tab:CoT_quality}
\end{table*}

Tab~\ref{tab:CoT_quality} presents a breakdown of the abbreviations used for assessing the quality of medical reasoning in large language models (LLMs) generation. The abbreviations include CLR (Clinical Logical Rigor), which measures the alignment of the CoT with the clinical diagnostic process in psychiatry; ES (Evidence Strength), which evaluates whether the CoT is based on authoritative guidelines, consensus, or reliable research; CC (Critical Completeness), which checks if the CoT adequately utilizes both positive and negative information; and DDA (Differential Diagnosis), which assesses whether the CoT systematically lists and excludes potential differential diagnoses. These abbreviations help to evaluate and ensure the quality of reasoning within LLMs' diagnostic processes.

To further validate our reasoning process, we conducted a blinded evaluation, inviting 12 experienced medical experts. The experts used a Likert scale to assess four core dimensions of the reasoning process: Clinical Logical Rigor (CLR), Evidence Strength (ES), Critical Completeness (CC), and Differential Diagnosis Adequacy (DDA). Each dimension evaluates the quality and reliability of the reasoning process from different perspectives. Specifically, CLR assesses whether the reasoning follows a standard clinical thought process and whether any logical inconsistencies are present; ES examines whether the evidence cited in the reasoning is authoritative and credible; CC evaluates whether all relevant information is considered and whether alternative diagnoses are reasonably excluded; and DDA focuses on whether all possible diagnoses have been reasonably ruled out. Each dimension has clear rating criteria, and the experts scored various aspects of the reasoning process based on these standards. The specific rating guidelines are shown in Fig~\ref{CLR},~\ref{ES},~\ref{CC} , and ~\ref{DDA}. The results provided valuable feedback, helping to optimize the diagnostic system and enhance its transparency, accuracy, and clinical reliability.

\begin{figure*}[tb]
\centering
\begin{tcolorbox}[
    colframe=black!60,       
    colback=black!5,         
    coltitle=white,
    fonttitle=\bfseries,
    title=\small Clinical Logical Rigor Rating Criteria,
    sharp corners,
    boxrule=0.4mm,
    boxsep=2pt,
    left=2pt, right=2pt,
    top=2pt, bottom=2pt,
    before upper={\setlength{\parskip}{0pt}} 
]
\small

\textbf{Dimension 1: Clinical Logical Rigor}

\textbf{Purpose:} \\
Evaluate whether the reasoning process aligns with standard clinical thinking. This dimension assesses the internal consistency and logical progression of the diagnostic reasoning path.

\textbf{Core Focus:} 
\begin{itemize}
    \item Does the reasoning follow a coherent clinical flow (e.g., from symptom presentation to diagnostic hypothesis to investigation or confirmation)?
    \item Are transitions between reasoning steps medically justified?
    \item Is there evidence of logical gaps or contradictions?
\end{itemize}

\textbf{Guidelines for Rating (1–5 Likert Scale):}

\begin{itemize}
    \item \textbf{Score 5 – Excellent:} \\
    The reasoning process is rigorous and mirrors professional diagnostic workflows. Each step logically builds upon the previous one, demonstrating clear clinical justification and no inconsistencies.
    
    \item \textbf{Score 4 – Good:} \\
    Mostly consistent reasoning with minor issues. Logical flow is preserved overall, but one or two steps may lack explicit justification or be slightly underdeveloped.
    
    \item \textbf{Score 3 – Fair:} \\
    The reasoning includes some coherent transitions but also contains noticeable logical gaps, unclear assumptions, or weak linkages between major steps.
    
    \item \textbf{Score 2 – Poor:} \\
    The reasoning shows significant inconsistencies or unjustified jumps between steps. Clinical logic is frequently broken or poorly structured.
    
    \item \textbf{Score 1 – Very Poor:} \\
    The reasoning path is fragmented or incoherent, with no clear structure resembling clinical logic. Diagnostic conclusions appear arbitrary or unsupported by prior steps.
\end{itemize}

\normalsize
\end{tcolorbox}

\captionsetup{aboveskip=4pt, belowskip=0pt, width=\linewidth}
\caption{Clinical Logical Rigor Rating Criteria}
\label{CLR}
\end{figure*}

\begin{figure*}[tb]
\centering
\begin{tcolorbox}[
    colframe=black!60,       
    colback=black!5,         
    coltitle=white,
    fonttitle=\bfseries,
    title=\small Evidence Strength Rating Criteria,
    sharp corners,
    boxrule=0.4mm,
    boxsep=2pt,
    left=2pt, right=2pt,
    top=2pt, bottom=2pt,
    before upper={\setlength{\parskip}{0pt}} 
]
\small

\textbf{Dimension 2: Evidence Strength }

\textbf{Purpose:} \\
Assess the strength, clarity, and credibility of the medical evidence used to support the reasoning. This dimension evaluates whether the knowledge cited is grounded in authoritative and up-to-date clinical resources.

\textbf{Core Focus:} 
\begin{itemize}
    \item Is the reasoning supported by evidence from recognized guidelines (e.g., ICD, DSM, WHO), expert consensus, or peer-reviewed clinical studies?
    \item Are the references to evidence explicit, precise, and contextually appropriate?
\end{itemize}

\textbf{Guidelines for Rating (1–5 Likert Scale):}

\begin{itemize}
    \item \textbf{Score 5 – Excellent:} \\
    The reasoning explicitly cites high-quality sources such as current clinical guidelines, landmark trials, or authoritative reviews. The evidence is specific, well-integrated, and directly supports the reasoning.

    \item \textbf{Score 4 – Good:} \\
    The reasoning is generally well-supported by clinical knowledge, with some explicit references. However, some sources may be less current or less clearly stated.

    \item \textbf{Score 3 – Fair:} \\
    The evidence is partially grounded in clinical understanding but lacks specificity. The reasoning may include general or vague references such as “studies show” or “it is commonly believed,” without citing concrete sources.

    \item \textbf{Score 2 – Poor:} \\
    The reasoning relies on outdated, unclear, or weak sources. It may reflect misconceptions or obsolete diagnostic categories, reducing its credibility.

    \item \textbf{Score 1 – Very Poor:} \\
    No discernible clinical evidence is used. The reasoning appears speculative, inaccurate, or contrary to established medical knowledge.
\end{itemize}

\normalsize
\end{tcolorbox}

\captionsetup{aboveskip=4pt, belowskip=0pt, width=\linewidth}
\caption{Evidence Strength Rating Criteria}
\label{ES}
\end{figure*}

\begin{figure*}[tb]
\centering
\begin{tcolorbox}[
    colframe=black!60,       
    colback=black!5,         
    coltitle=white,
    fonttitle=\bfseries,
    title=\small Critical Completeness Rating Criteria,
    sharp corners,
    boxrule=0.4mm,
    boxsep=2pt,
    left=2pt, right=2pt,
    top=2pt, bottom=2pt,
    before upper={\setlength{\parskip}{0pt}} 
]
\small
\textbf{Dimension 3: Critical Completeness}

\textbf{Purpose:} \\
Evaluate whether the reasoning thoroughly integrates all relevant clinical information, including both positive findings and critical negative evidence.

\textbf{Core Focus:} 
\begin{itemize}
    \item Does the reasoning account for all key clues in the case, including negative symptoms, lab findings, and contextual information?
    \item Are any important signs, contradictions, or alternative possibilities overlooked?
\end{itemize}

\textbf{Guidelines for Rating (1–5 Likert Scale):}

\begin{itemize}
    \item \textbf{Score 5 – Excellent:} \\
    The reasoning comprehensively incorporates all relevant case information, including less obvious symptoms or negative findings. Contradictory data is acknowledged and thoughtfully analyzed.

    \item \textbf{Score 4 – Good:} \\
    Most critical information is considered, though a few minor details or counterpoints may be overlooked. Overall, the analysis remains balanced and well-supported.

    \item \textbf{Score 3 – Fair:} \\
    The reasoning captures the main symptoms but partially overlooks important secondary findings or negative evidence. Some diagnostic bias may be present.

    \item \textbf{Score 2 – Poor:} \\
    Key details—especially disconfirming or contradictory evidence—are neglected. The analysis appears narrow or overly focused on a single hypothesis.

    \item \textbf{Score 1 – Very Poor:} \\
    The reasoning is based on incomplete information, ignoring significant findings that undermine the conclusion. The result is a superficial or misleading diagnostic interpretation.
\end{itemize}

\normalsize
\end{tcolorbox}

\captionsetup{aboveskip=4pt, belowskip=0pt, width=\linewidth}
\caption{Critical Completeness Rating Criteria}
\label{CC}
\end{figure*}

\begin{figure*}[tb]
\centering
\begin{tcolorbox}[
    colframe=black!60,       
    colback=black!5,         
    coltitle=white,
    fonttitle=\bfseries,
    title=\small Differential Diagnosis Adequacy Rating Criteria,
    sharp corners,
    boxrule=0.4mm,
    boxsep=2pt,
    left=2pt, right=2pt,
    top=2pt, bottom=2pt,
    before upper={\setlength{\parskip}{0pt}} 
]
\small
\textbf{Dimension 4: Differential Diagnosis Adequacy}

\textbf{Purpose:} \\
Assess whether the reasoning demonstrates a structured and comprehensive approach to differential diagnosis.

\textbf{Core Focus:} 
\begin{itemize}
    \item Has the response considered a broad but relevant range of possible diagnoses?
    \item Are common, high-probability, or life-threatening conditions addressed?
    \item Are exclusions logically and explicitly justified?
\end{itemize}

\textbf{Guidelines for Rating (1–5 Likert Scale):}

\begin{itemize}
    \item \textbf{Score 5 – Excellent:} \\
    A well-structured differential diagnosis is presented, including both common and critical conditions. Diagnoses are prioritized by likelihood, and exclusions are clearly justified using clinical evidence.

    \item \textbf{Score 4 – Good:} \\
    Several plausible alternatives are considered, and key conditions are ruled out with reasonable justification. Some minor omissions may exist, but the diagnostic reasoning remains strong.

    \item \textbf{Score 3 – Fair:} \\
    A limited differential is provided with partial justification. Some relevant conditions may be missing or excluded without clear rationale. The diagnostic scope is somewhat narrow.

    \item \textbf{Score 2 – Poor:} \\
    Only one diagnosis is offered, or the differential includes mostly irrelevant or unsupported options. Little to no effort is made to rule out other possibilities.

    \item \textbf{Score 1 – Very Poor:} \\
    No meaningful differential diagnosis is attempted. The reasoning is overly simplistic, with no recognition of alternative explanations.
\end{itemize}

\normalsize
\end{tcolorbox}

\captionsetup{aboveskip=4pt, belowskip=0pt, width=\linewidth}
\caption{Differential Diagnosis Adequacy Rating Criteria}
\label{DDA}
\end{figure*}

\subsection{Hyperparameter Settings}\label{sec:hyperparameters}

We trained \textbf{MentalSeek-Dx-7B} and \textbf{MentalSeek-Dx-14B} by building upon the pre-existing models Qwen2.5-7B-Instruct and Qwen2.5-14B-Instruct, respectively. The training process was carried out in two distinct stages, each designed to address different aspects of fine-tuning and optimization for the models. The experiments were conducted on 16 A800 GPUs.

In the first stage, we performed supervised fine-tuning using a dataset of 14,093 carefully refined medical records. These records were selected to ensure they represented a broad range of medical diagnoses, allowing the models to learn from high-quality data. This stage lasted for 3 epochs, with a learning rate of \(5 \times 10^{-5}\) and a batch size of 128. The primary goal of this stage was to expose the models to the specific language and terminology used in psychiatric and medical contexts, enhancing their ability to handle complex diagnostic tasks.

Following the supervised fine-tuning, the second stage focused on reinforcement learning using GPRO (Generalized Proximal Policy Optimization) applied to \textbf{MentalSeek-Dx-7B} and \textbf{MentalSeek-Dx-14B}. In this stage, we aimed to improve the model's diagnostic accuracy by addressing specific errors identified during Stage 1. The errors were analyzed and selected to form a training corpus, which was used to guide the learning process in areas where the models exhibited weaknesses. By leveraging model generations that showed diagnostic errors, we were able to focus the training on correcting those mistakes.

The reinforcement learning process utilized an Actor-Critic architecture, with separate learning rates for the Actor and Critic components set at \(3 \times 10^{-6}\) and \(1 \times 10^{-5}\), respectively. The training was performed with a batch size of 64 and 3 epochs, providing enough iterations for effective fine-tuning. Several additional hyperparameters were used to guide the learning process, such as a discount factor of 1.0, which ensures equal weighting between future and immediate rewards, a value loss coefficient of 0.5 to control the learning of state values, and a clipping range of 0.2 to avoid large, destabilizing updates. A KL divergence coefficient of 0.001 was applied to maintain a balance between the old and new policies, ensuring that the learned policy did not deviate too far from the baseline. For each diagnostic prompt, the model generated 8 candidate responses, which were then evaluated to select the most appropriate diagnostic output.

This two-stage training procedure, combining supervised fine-tuning and reinforcement learning, allowed \textbf{MentalSeek-Dx-7B} and \textbf{MentalSeek-Dx-14B} to effectively learn from both curated data and targeted error correction, enhancing their performance in the challenging task of psychiatric diagnosis.

\subsection{Reinforcement Learning Setup}
\label{CRL}

The results presented in Fig~\ref{RL_CRL.png} show that although the improvement with CRL is relatively small, it significantly facilitates a smoother transition during the reinforcement learning process. In our experiment, we compared three different reward strategies to assess their effectiveness in handling reinforcement learning tasks. The first strategy, Disorder-based reward, awards points only for the last predicted disorder. If the predicted disorder matches the actual disorder, points are awarded; otherwise, no points are given. The advantage of this strategy is its simplicity and clarity, but its major limitation is its heavy reliance on the final disorder prediction, which may hinder the optimization of earlier reasoning stages. The second strategy, Process-based reward, provides rewards for three key components—category, hypothesis list, and disorder prediction—with fixed scoring proportions of 1:1:1 for each component. This strategy takes a more holistic approach by considering each element’s contribution, but its limitation lies in the fixed ratio, which may not adapt flexibly to the learning needs at different stages, potentially leading to insufficient optimization in certain areas. The third strategy, \textbf{Progressive Process-based reward}, dynamically adjusts the reward distribution based on the training step. In different steps, the reward distribution changes as follows: in step 1, the ratio of category:list:disorder is 2:1:1; in step 2, it is 1:2:1; and in step 3, it is 1:1:2. This approach’s advantage is its ability to gradually emphasize disorder prediction as training progresses, allowing the model to focus more on precise diagnosis in the later stages while placing greater importance on category and hypothesis list in the earlier stages.

The reward updates for the process-based strategy are normalized to ensure comparability across different training phases. The experimental results show that while the Disorder-based reward strategy provides relatively balanced feedback, its sparse scoring does not significantly enhance the model's stability or accuracy. The Process-based reward strategy exhibits significant fluctuations during training, often interrupting the learning process and making stable improvement difficult. In contrast, the \textbf{Progressive Process-based reward} strategy demonstrates a more stable upward trend in the scores, indicating that this dynamic reward adjustment approach is more effective in gradually optimizing the model. In conclusion, the \textbf{Progressive Process-based reward} strategy clearly facilitates a smoother reinforcement learning process, particularly when handling complex diagnostic tasks, by reducing instability in reasoning and improving the accuracy of the final diagnosis.

\subsection{Case Study}

\begin{figure*}[t]
\centering
\includegraphics[width=\linewidth]{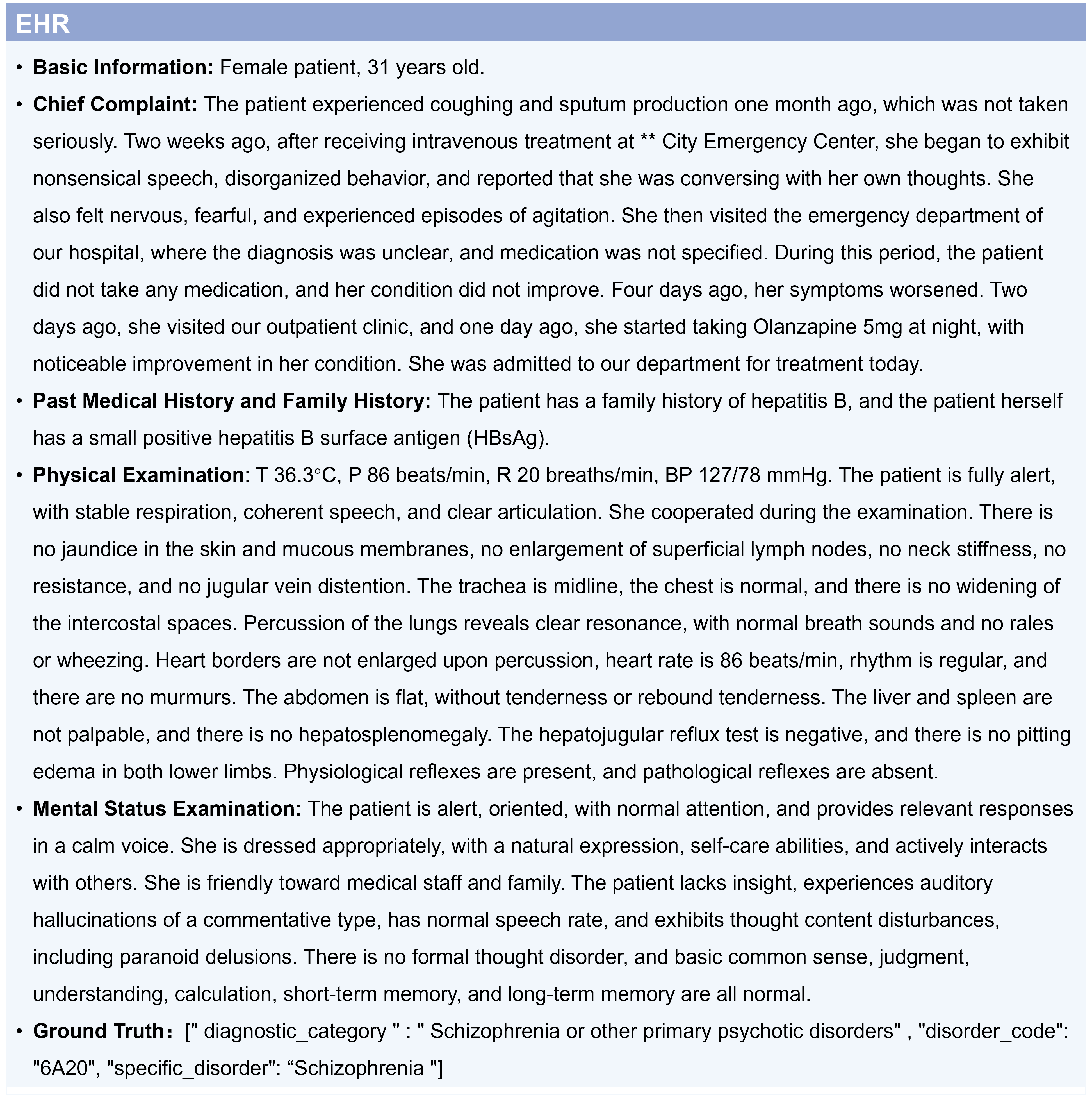}
\caption{EHR inputs.}
\label{appendix_case_information}
\end{figure*}

\begin{figure*}[th]
\centering
\includegraphics[width=\linewidth]{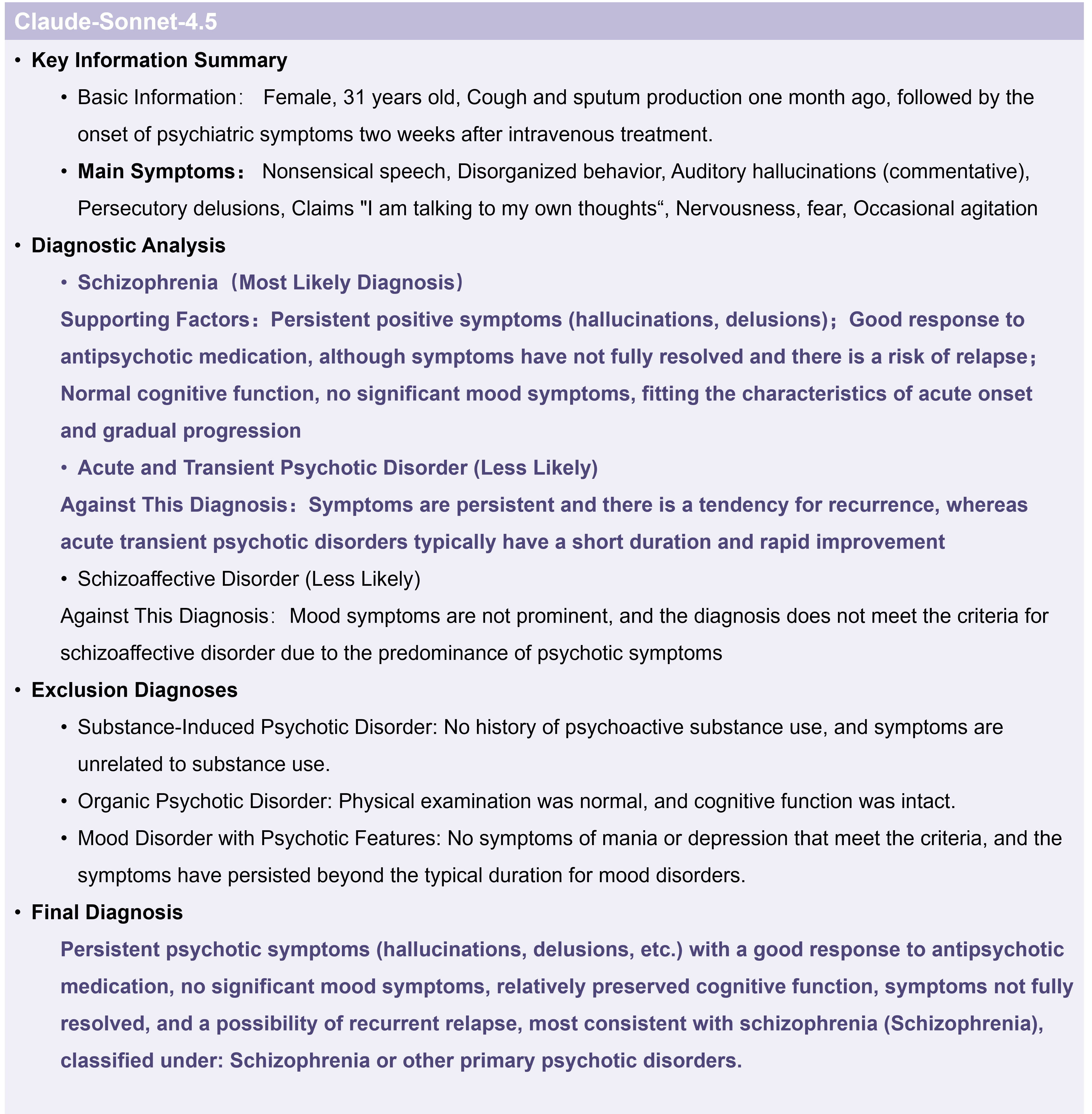}
\caption{Claude-Sonnet-4.5 prediction.}
\label{appendix_case_study_Claude}
\end{figure*}

We selected a representative case for a comparative analysis, as shown in Fig~\ref{appendix_case_information}, to illustrate the effects of different reasoning processes. Claude-Sonnet-4.5 employs a pattern-based association reasoning method, inferring specific disorders by identifying known patterns and retrospectively mapping them to diagnostic categories. However, this approach has significant flaws, manifesting as a paradigm misalignment. It relies on pre-set pattern associations while bypassing a structured psychiatric reasoning process, thereby neglecting a comprehensive evaluation of the patient’s condition. Specifically, Claude-Sonnet-4.5 directly links hallucinations to schizophrenia without considering alternative diagnoses or contextual factors. Its reasoning process fails to adequately explore differential diagnoses and does not effectively differentiate or validate various diagnostic possibilities. This reasoning model reflects a lack of a structured internal knowledge framework, ultimately leading to diagnostic errors and inaccurate predictions. The entire diagnostic process is shown in Fig~\ref{appendix_case_study_Claude}.

\begin{figure*}[t]
\centering
\includegraphics[width=\linewidth]{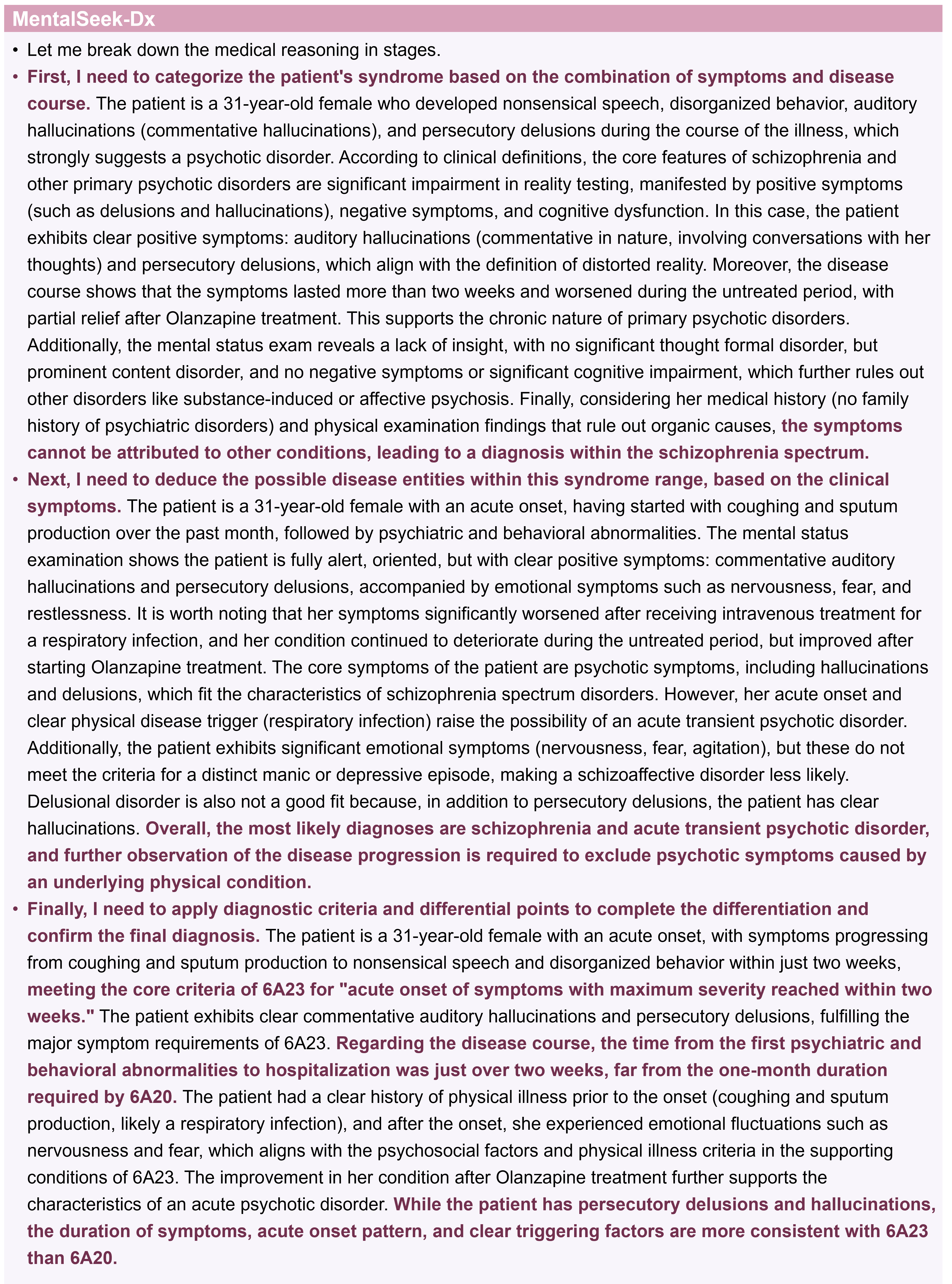}
\caption{MentalSeek-Dx (ours) prediction.}
\label{appendix_case_study_MentalSeekDx}
\end{figure*}

In contrast, MentalSeek-Dx adopts a more structured, clinically grounded reasoning paradigm. This system strictly constrains the categorization reasoning process, ensuring that each step is based on clinical evidence and diagnostic criteria. First, the system clearly defines the diagnostic domain and narrows the reasoning scope, then generates multiple hypotheses, which are further tested through scientific methods. Finally, MentalSeek-Dx applies standardized diagnostic criteria for differential diagnosis, ensuring that each hypothesis undergoes thorough clinical validation. This process not only follows a standard medical reasoning path but also makes the diagnostic process more transparent and accurate. Through this structured reasoning approach, MentalSeek-Dx provides more precise and interpretable diagnostic predictions, significantly reducing the risk of diagnostic errors and enhancing clinical decision-making reliability. The entire diagnostic process is shown in Fig~\ref{appendix_case_study_MentalSeekDx}.

\label{case_study_2}

\end{document}